\def\paperID{10110}
\def\confName{CVPR}
\def\confYear{2024}

\def\paperTitle{EMOPortraits: Emotion-enhanced Multimodal One-shot Head Avatars}

\def\authorBlock{
    Nikita Drobyshev \qquad
    Antoni Bigata Casademunt \qquad
    Konstantinos Vougioukas \qquad
    Zoe Landgraf \qquad
    \\
    Stavros Petridis \qquad
    Maja Pantic \qquad
    \\
    {\tt\small \{n.drobyshev, a.bigata-casademunt22, k.vougioukas,} \\
    {\tt\small zoe.landgraf15, stavros.petridis04, m.pantic\}@imperial.ac.uk} \\
    Imperial College London
}

\newif\ifreview 
\newif\ifarxiv \newcommand{\arxiv}{\arxivtrue}
\newif\ifcamera 
\newif\ifrebuttal 

\arxiv 

\pdfoutput=1
\documentclass[10pt,twocolumn,letterpaper]{article}
\usepackage{graphicx}
\usepackage[accsupp]{axessibility} 
\usepackage{tabularx}
\usepackage{color}
\usepackage{subcaption}
\usepackage{multirow}
\usepackage{balance}
\usepackage{etoolbox}
\usepackage{mwe}
\usepackage{amssymb}
\usepackage{pifont}
\newcommand{\cmark}{\ding{51}}%
\newcommand{\xmark}{\ding{55}}%
\usepackage{titlesec}

\ifreview \usepackage[review]{cvpr} \fi
\ifarxiv \usepackage[pagenumbers]{cvpr} \fi
\ifrebuttal \usepackage[rebuttal]{cvpr} \fi
\ifcamera \usepackage{cvpr} \fi

\usepackage{graphicx}	
\usepackage{amsmath}	
\usepackage{amssymb}	
\usepackage{booktabs}
\usepackage{times}
\usepackage{microtype}
\usepackage{epsfig}
\usepackage[table,xcdraw,dvipsnames]{xcolor}
\usepackage{caption}
\usepackage{float}
\usepackage{placeins}
\usepackage{color, colortbl}
\usepackage{stfloats}
\usepackage{enumitem}
\usepackage{tabularx}
\usepackage{xstring}
\usepackage{multirow}
\usepackage{xspace}
\usepackage{url}
\usepackage{subcaption}
\usepackage{xcolor}
\usepackage[hang,flushmargin]{footmisc}

\ifcamera \usepackage[accsupp]{axessibility} \fi

% ...

\ifarxiv  \fi

\newcommand{\R}[1]{{%
    \textbf{%
        \ifstrequal{#1}{1}{\textcolor{red}{R#1}}{%
        \ifstrequal{#1}{2}{\textcolor{blue}{R#1}}{%
        \ifstrequal{#1}{3}{\textcolor{magenta}{R#1}}{%
        \ifstrequal{#1}{4}{\textcolor{teal}{R#1}}{%
                           \textcolor{cyan}{R#1}%
        }}}}%
    }%
}}

\usepackage{mathtools}
\DeclarePairedDelimiter\abs{\lvert}{\rvert}%
\DeclarePairedDelimiter\norm{\lVert}{\rVert}

\usepackage{xr-hyper}

\makeatletter
\newcommand*{\addFileDependency}[1]{
  \typeout{(#1)}
  \@addtofilelist{#1}
  \IfFileExists{#1}{}{\typeout{No file #1.}}
}

\makeatother

\definecolor{cvprblue}{rgb}{0.21,0.49,0.74}
\usepackage[pagebackref,breaklinks,colorlinks,citecolor=cvprblue]{hyperref}
\usepackage[capitalize]{cleveref}
\crefname{section}{Sec.}{Secs.}
\crefname{table}{Table}{Tables}
\crefname{figure}{Fig.}{Figs.}

\begin{document}
\definecolor{brickred}{rgb}{0.8, 0.25, 0.33}
\title{\paperTitle}
\author{\authorBlock}

\twocolumn[{

\renewcommand\twocolumn[1][]{#1}
\maketitle
    
\begin{center}
    \includegraphics[width=0.9\linewidth]{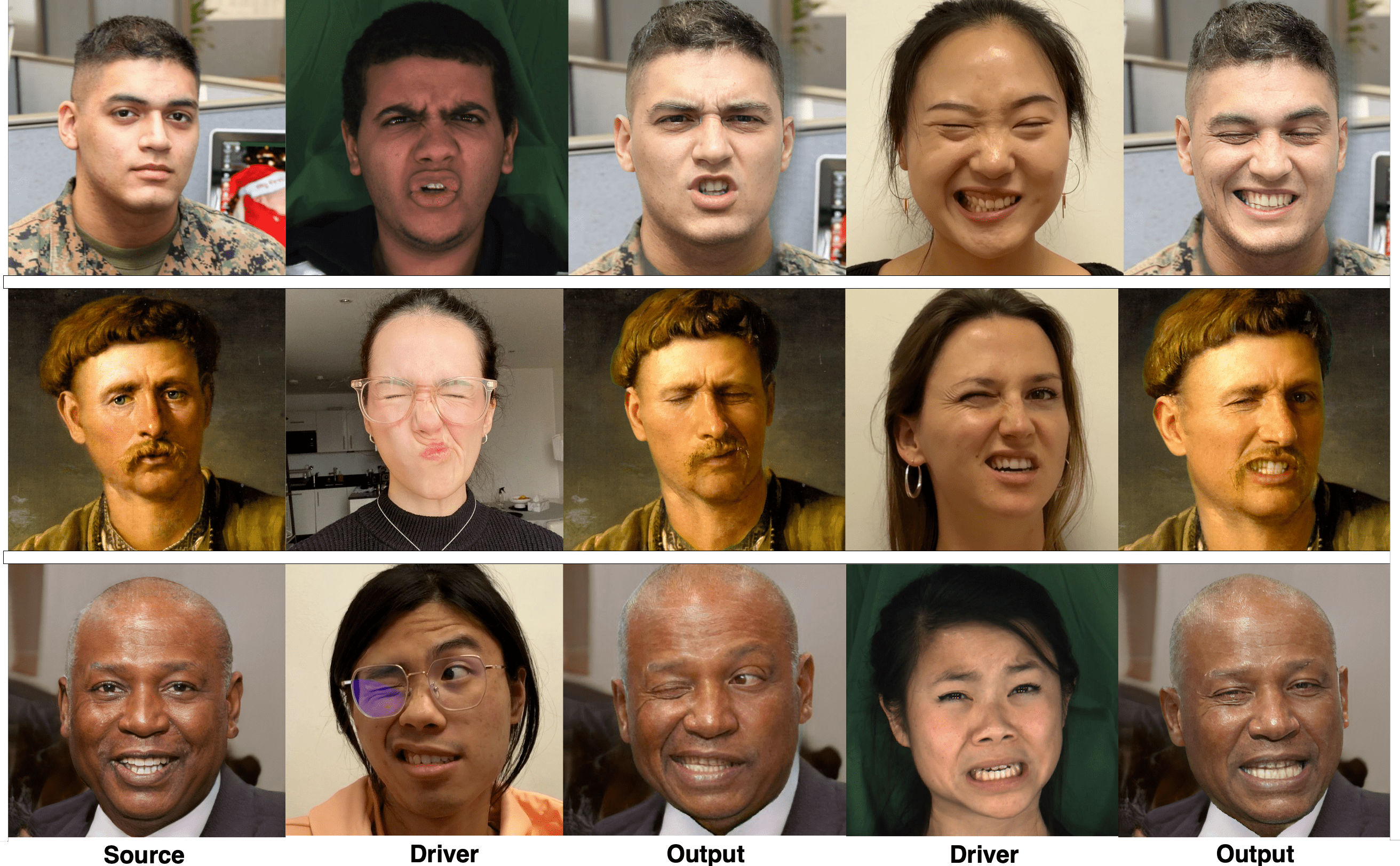}
    \captionof{figure}{
     Selected animation results for image-driven mode of our method.  
    } 
    \label{fig:teaser}
\end{center}%
}]

\def\X{\mathbf{X}}
\def\A{\mathbf{A}}
\def\x{\mathbf{x}}
\def\w{\mathbf{w}}
\def\v{\mathbf{v}}
\def\e{\mathbf{e}}
\def\E{\mathbf{E}}
\def\R{\mathbf{R}}
\def\z{\mathbf{z}}
\def\t{\mathbf{t}}
\def\G{\mathbf{G}}
\def\W{\mathbf{W}}

\begin{abstract}
\vspace{-10pt} 
Head avatars animated by visual signals have gained popularity, particularly in cross-driving synthesis where the driver differs from the animated character, a challenging but highly practical approach. The recently presented MegaPortraits model has demonstrated state-of-the-art results in this domain. We conduct a deep examination and evaluation of this model, with a particular focus on its latent space for facial expression descriptors, and uncover several limitations with its ability to express intense face motions. To address these limitations, we propose substantial changes in both training pipeline and model architecture, to introduce our \textbf{EMOPortraits} model, where we:\\

Enhance the model's capability to faithfully support intense, asymmetric face expressions, setting a new state-of-the-art result in the emotion transfer task, surpassing previous methods in both metrics and quality.

Incorporate speech-driven mode to our model, achieving top-tier performance in audio-driven facial animation, making it possible to drive source identity through diverse modalities, including visual signal, audio, or a blend of both.
 
We propose a novel multi-view video dataset featuring a wide range of intense and asymmetric facial expressions, filling the gap with absence of such data in existing datasets.
For dataset and video examples please visit \href{https://neeek2303.github.io/EMOPortraits/}{project page.}

\end{abstract}

\vspace{-14pt} 
\section{Introduction}
\label{sec:intro}

\subsection{Representation of facial expression}

Advancements in neural head technologies now enable the creation of realistic avatars from a few or even one image, with the latter being crucial when only one source image is available. Cross-driving synthesis, where avatars are animated with different identities is a key technique for virtual reality, filmmaking, photo animation, etc. However, accurate transfer of facial expressions, especially those that are intense and uneven, remains a substantial challenge in avatar animation, particularly for cross-driving synthesis. While previous research focused mainly on preserving identity and transferring moderate facial motions, our work also seeks to accurately drive high-intensity and asymmetric expressions.

Our research is built on and extends the MegaPortraits \cite{drobyshev2023megaportraits} model, notable for state-of-the-art results in cross-driving synthesis. An additional advantage using this method is that, unlike many avatar systems that depend on limited \cite{tellamekala20233d, Burkov_2020} predefined motion descriptors, MegaPortraits, learns expression representations from scratch, allowing for greater adaptability to a wider range of expressions. We investigate the latent expression spaces and training methods of MegaPortraits to enhance its ability to depict a broad spectrum of facial expressions. Our comprehensive analysis reveals that, while the original model shows limited effectiveness in intense motion representing, it has significant potential for improvement through targeted architectural modifications, adjustments in the training approach, and the integration of our novel dataset.

\subsection{Integrating speech driving}

We integrate speech in our model to be used either as a complementary or an alternative driver, a crucial aspect for applications like virtual assistance or mixed reality when a primary visual signal is absent. 

By improving disentanglement of facial expression latent space, we remove disturbing information from face motion descriptors and emphasize components that solely control lip movements. Based on this, we formulated a novel loss that helps to achieve desirable results. Furthermore, our method can generate plausible head rotations and blinks, which appear natural and enhance its applicability across various tasks. Thus, our final method can drive source identity through image, audio, or a mix of them.


\subsection{FEED: Facial Extreme Emotions Dataset}

Public data scarcity is a common obstacle in deep learning, the topic of human facial expressions and movements is not an exception. In particular, we note a lack of high quality video data capturing a wide range of facial expressions, with existing datasets not going beyond basic facial actions shown at \cref{fig:curr_dataset_vision}.
To address this, we collect a new dataset that includes basic expressions and also captures complex movements like blinks, winks, and head and tongue movements, along with varied extreme expressions which are difficult or impossible to categorize through basic facial actions. Given this variety in expressions in our dataset, we believe that it will be a valuable resource for research in human emotions and facial reconstruction fields. In summary, our main contributions are as follows:

\begin{itemize}


    \item We introduce our new \textbf{EMOPortraits} model for one-shot head avatars synthesis, that is capable of transferring intense facial expression, showing \textit{state-of-the-art} results in cross-driving synthesis. We achieve this through careful latent facial expression space development as well as novel losses and a minimal amount of domain-specific data \textit{($\sim$ 0.1 \% of the train set)}.


    \item We integrate a speech-driving mode in our model, that demonstrates cutting-edge results in speech-driven animations. It functions effectively alongside visual signals or independently, also generating realistic head rotations and eye blinks.
    
    \item We present a unique multi-view dataset that spans a broad spectrum of extreme facial expressions, filling the gap of absence of such data in existing datasets.

\end{itemize}

\section{Related Work}
\label{sec:related}

\subsection{Visual-driven head avatars}
In recent years, the domain of neural avatars has branched into two prominent sub-fields: person-specific and person-agnostic avatars. While person-specific methodologies ~\cite{zhou2021posecontrollable, zheng2022i, Xu_2023, gafni2020dynamic, grassal2022neural} excel in delivering stunning realism and motion fidelity for a particular individual, they encounter challenges representing an arbitrary person. Moreover, these approach demands multiple frames of training data, distinct training for each avatar, and can struggle to replicate motions not encountered during training.



Person-agnostic methods, on the other hand, don't need train or fine-tuning for each new person and present an alternative approach to talking-head synthesis. Earlier works in this domain generated avatars in a few-shot technique \cite{zakharov2019fewshot, Burkov_2020}, while subsequent studies introduced one-shot capabilities \cite{wang2021facevid2vid, doukas2021headgan, Siarohin2019FirstOM, zakharov2020fast, drobyshev2023megaportraits, yin2022styleheat, yu2023nofa, bounareli2023hyperreenact}. Many of these works employ predefined motion representations, such as 3DMM's blendshapes \cite{doukas2021headgan, yin2022styleheat, yu2023nofa, khakhulin2022realistic}. In contrast, some learned latent emotion representation from scratch \cite{zakharov2020fast, Burkov_2020, zhang2023metaportrait, drobyshev2023megaportraits, Siarohin2019FirstOM}. This latter approach holds potential for better motion representation as in such settings, expression descriptors become entirely trainable and free from inheriting the limitations typical for blendshapes as was shown in \cite{tellamekala20233d, Burkov_2020}. In our work, we adopt this strategy, remarkably improving upon the methodology used in MegaPortraits\cite{drobyshev2023megaportraits}.

\subsection{Face expression datasets}
Early image datasets\cite{lucey2010extended, barsoum2016training} predominantly offered annotated face expression data corresponded to up to 8 basic emotions. However, these datasets are not well-suited for training head avatars, which necessitate video or multi-view data.

The SAVEE \cite{jackson2014surrey} dataset captures facial expressions in speech, but involves just 4 actors. The MMI \cite{pantic2005web} dataset, more expansive in actor participation, offers spontaneous moderate-intensity expression but is restricted to single-view sequences. RAVDESS \cite{livingstone2018ryerson} distinguishes itself by capturing two motion intensities, but its limited recordings challenge broad applicability. CREMA-D \cite{cao2014crema} encompasses three expression intensities, and though MEAD \cite{wang2020mead} provides detailed multi-view data across three intensities, it's limited to the standard eight expression groups. Our novel FEED dataset aims to address the scientific community's demand for high-quality multi-view facial expression videos outside the standard categories: Joy, Fear, Sadness, Disgust, Anger Contempt and Surprise. 
\begin{table}
    \centering
    \resizebox{0.95\linewidth}{!}{
    \begin{tabular}{c|ccccccc}
        \multicolumn{8}{c}{\textbf{FEED dataset internal structure}}
        \\
        Method 
        &
        BFEs
        &
        Head rotations
                &
        Winks
                &
        Eyes move
                &
        Asym. em.
                &
        Tongue em.
                &
        Extreme em.

        \\
        \hline
        Mode & Mild/Strong & Axes/random & Mild/Strong & Random & Strong & Strong & Extreme 
        \\
        \# of part.   & 21/23 & 21 & 23 & 21 & 23 & 23 & 23  \\
        Avg. lenght        & 1:38/1:52 & 0:58 & 0:49 & 0:33 & 1:41 & 1:53 & 3:02 \\
        
    \end{tabular}
    }
    \caption{The tasks we asked our participants to perform for our FEED dataset. Here BFEs stands for basic facial expressions, shown in \cref{fig:curr_dataset_vision} }
    \label{tab:feed_tasks}
\end{table}

\begin{table}[t]
    \setlength{\tabcolsep}{3.0pt}
    \footnotesize
    \centering

    \resizebox{0.95\linewidth}{!}{
    \begin{tabular}{c|ccccccc}
        \multicolumn{8}{c}{\textbf{Datasets comparison}}
        \\
        Dataset 
        &
        \#Actors
        &
        \#Views
        &
        Basic exp.
        &
        Strong exp.
        &
        Extrime \& Assym. exp.
        &
        Tongue exp.
        &
        Resolution
        \\
        \hline
        SAVESS   & 4 & 1 & \textcolor{ForestGreen}{\cmark} & \textcolor{red}{\xmark} & \textcolor{red}{\xmark} & \textcolor{red}{\xmark} & 1280 × 1024 \\
        RAVGESS  & 24 & 1 & \textcolor{ForestGreen}{\cmark} & \textcolor{red}{\xmark} & \textcolor{red}{\xmark} & \textcolor{red}{\xmark}  & 1920 × 1080 \\
        CREMA-D  &  \textbf{91} & 1 & \textcolor{ForestGreen}{\cmark} & \textcolor{red}{\xmark} & \textcolor{red}{\xmark}  & \textcolor{red}{\xmark} & 1280 × 720 \\
        MMI      & 25 & 1 & \textcolor{ForestGreen}{\cmark} & \textcolor{ForestGreen}{\cmark} & \textcolor{red}{\xmark}  & \textcolor{red}{\xmark} & 1920 × 1080 \\
        MEAD     & \underline{48} &  \textbf{7} & \textcolor{ForestGreen}{\cmark} & \textcolor{ForestGreen}{\cmark} & \textcolor{red}{\xmark} & \textcolor{red}{\xmark} & 1920 × 1080 \\
        \hline
        Ours (FEED) & 23 & \underline{3} & \textcolor{ForestGreen}{\cmark} & \textcolor{ForestGreen}{\cmark} & \textcolor{ForestGreen}{\cmark} &  \textcolor{ForestGreen}{\cmark} &  \textbf{3840 x 2160} \\
    \end{tabular}
    }    
    \caption{Comparisons with modern, high-quality audio-visual datasets created in controlled settings.}
    \label{tab:feed_comparison}
\end{table}

\subsection{Speech-driven head avatars}

While numerous studies introduce speech-driven avatars \cite{prajwal2020lip, zhou2021posecontrollable, zhang2023sadtalker, stypułkowski2023diffused, yin2022styleheat, Zhou_2020, gururani2022space}, few excel in producing high-quality talking heads with authentic rotations, blinks, and the capability to use both visual and audio inputs. For instance, Wav2Lip \cite{prajwal2020lip} aims at re-dubbing videos with accurate lip motions, but often falls short in realism with using single images. Diffused heads \cite{stypułkowski2023diffused} struggles to generate long sequences and doesnt provide access to pose control. MakeItTalk \cite{Zhou_2020} animates facial landmarks in a speaker-specific manner, yet struggles with head pose controls due to its non-reliance on 3D. PC-AVS \cite{zhou2021posecontrollable} offers a solution, but demands a driving video for pose modulation. Newly developed models such as SadTalker \cite{zhang2023sadtalker} and StyleHEAT \cite{yin2022styleheat} have shown promising outcomes in generating high-quality talking heads. We conducted a quantitative evaluation of our audio-driven mode compared to the aforementioned models.

\begin{figure}[tp]
    \centering
    \includegraphics[width=0.9\linewidth]{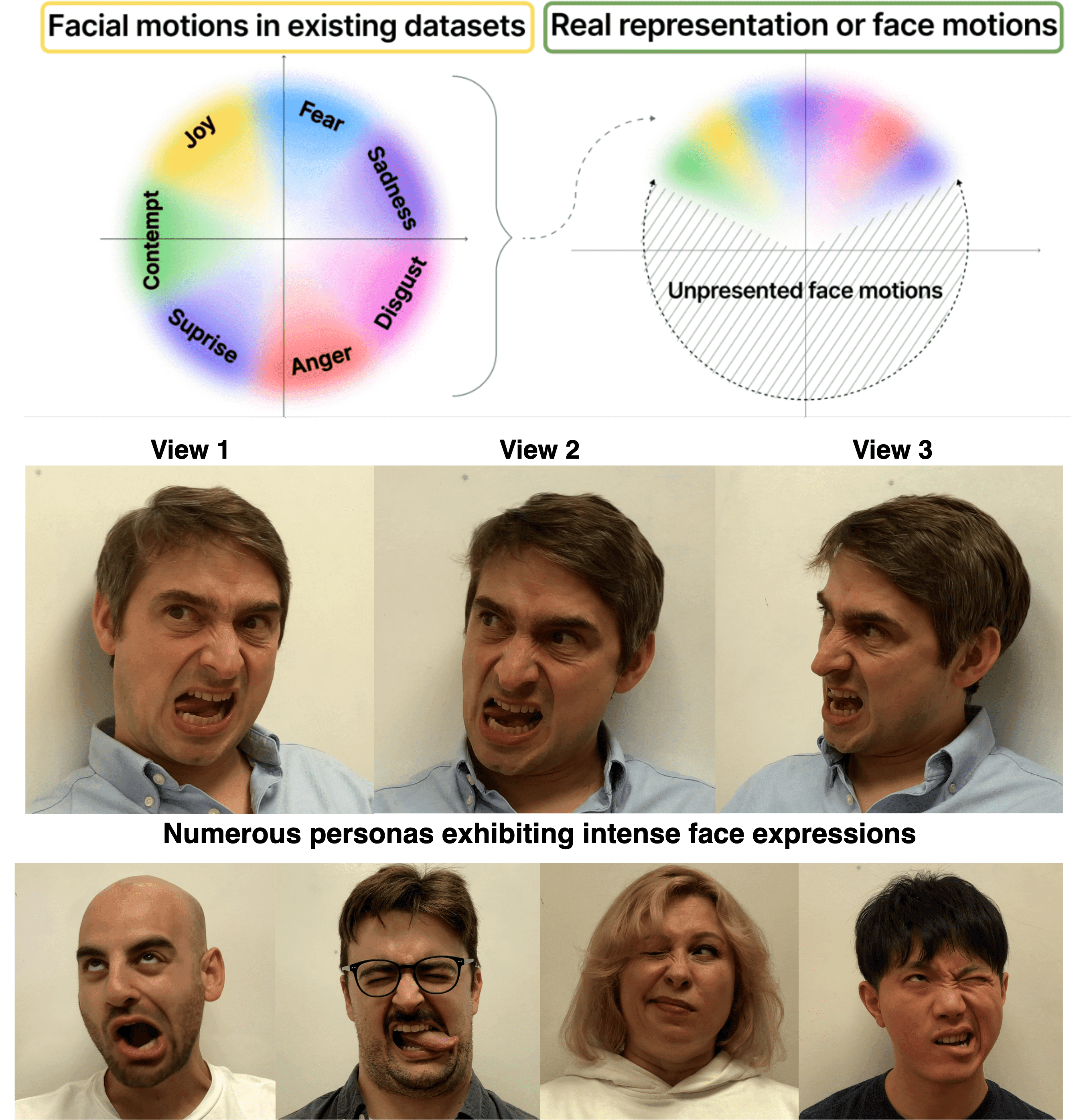}
    \caption{Illustration of the problem in publicly available face expression data and selected examples from our FEED dataset.}
    \label{fig:curr_dataset_vision}
\end{figure}

\section{FEED dataset}

\label{sec:feed_method}

As previously mentioned, the current publicly available datasets with human face expression videos fall short in capturing a broad range of facial manipulations (see illustration at \cref{fig:curr_dataset_vision}). In our view, this gap could stem from the complexity of designing clear tasks for participants, coupled with the reality that not everyone is ready for performing extreme expressions on camera. To bridge this gap, our pioneering FEED dataset is designed to meet the scientific community's need for high-quality, multi-view emotion videos that extend beyond standard emotions. It contains various expressions, including strong asymmetric ones, tongue and cheeks movements, winks, head rotations, eye movements and more nuanced gestures.
;

Our dataset consists of 520 multi-view videos
of 23 subjects, captured with 3 cameras. As extreme face motions are complex and their perception can be heavily influenced by subtle differences, we use a high resolution of 4k for all video, capturing the whole face up to the level of individual hair strands and wrinkles, as shown in \cref{fig:curr_dataset_vision}. For more examples, please refer to the supplementary materials. Our participants were asked to perform 7 tasks (see \cref{tab:feed_tasks}) to cover as many facial and head movements as possible.

In our comparison with other expression datasets detailed in \cref{tab:feed_comparison}, the MEAD dataset \cite{wang2020mead} emerges as the closest to ours regarding viewpoints and maximal expression intensity. However, despite MEAD's larger number of actors and viewpoints, it offers significantly less variety of facial expressions and have lower resolution than our dataset.

\label{sec:method_feed}
\section{Expression enhancement}
\label{sec:method}

We found that the MegaPortraits \cite{drobyshev2023megaportraits} model fails to transfer intense expressions correctly, as shown in \cref{fig:comparison_qual}. Using our FEED dataset of strong and uneven facial expression to fine-tune a pre-trained MegaPortraits model didn't enhance final results as shown at \cref{tab:ablation_img_main}. Training from scratch on FEED leads to fast overfitting due to the small number of identities presented in the dataset compared to VoxCeleb2 \cite{Chung2018VoxCeleb2DS} used in \cite{drobyshev2023megaportraits} (23 vs $\sim$5000). Our method effectively injects this small dataset into training, yielding desirable results. We list our key findings here, for all details, see supplementary materials. Our model's scheme, displayed in \cref{fig:main_scheme}, shares similarities with \cite{drobyshev2023megaportraits}, leading us to occasionally refer to certain elements of our scheme in the context of \cite{drobyshev2023megaportraits}'s pipeline.

\subsection{Latent expression space}
\label{ssec:lat_expr_spc}

\begin{figure}[H]
    \centering
    \includegraphics[width=0.8\linewidth]{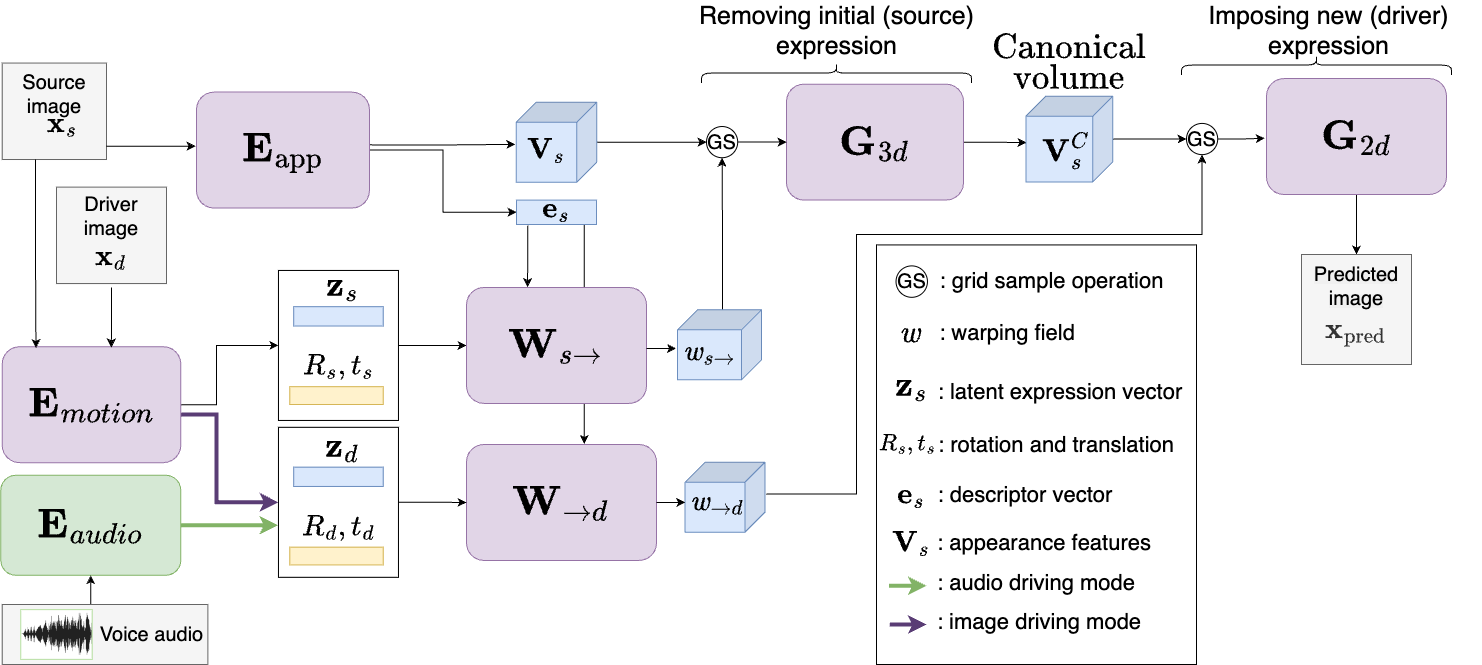}
    \caption{Method Overview. We use $\E_\text{app}$ to extract volume features $\mathbf{V}_s$ and a global descriptor $\e_s$ from the source image. Then $\E_\text{motion}$ or $\E_\text{audio}$  generates motion representations from source and driver, including head rotations $\R_{s/d}$, translations $\t_{s/d}$, and expression descriptors $\z_{s/d}$. Using them, we predict warpings $\w_{s \rightarrow}$ and $\w_{\rightarrow d}$. First warping and $\G_\text{3D}$ transform $\mathbf{V}_s$ into a canonical volume $\mathbf{V}_s^C$ by removing the source motions. Second warping and $\G_\text{2D}$ imposes the driver's motions and renders the final image.}
    \label{fig:main_scheme}
\end{figure}

We begin by exploring MegaPortraits' latent expression descriptors' space ($\z_{s/d}$ in \cref{fig:main_scheme}), which is crucial for expression transfer, through PCA analysis. Inspired by \cite{li2022understanding}, which demonstrated that the area under the cumulative explained variance curve of singular values (denoted here as \(\mathbf{AUC}_{\mathbf{z}}\) ) can serve as an effective metric for dimensionality collapse in some latent spaces, we applied this measure to our study.

This metric allows us to forecast model performance: greater collapse suggests reduced entropy in the latent space, which correlates with lower representational quality. High quality of the expression space, produced by \(\mathbf{E}_{\textit{motion}}\) (as shown in \cref{fig:main_scheme}), is crucial to capture and differentiate between the nuances of strong facial expressions. Our ablation study confirms (see \cref{tab:ablation_img_main}) that there is a notable correlation between how broad and isotropic model's latent space is (expressed by \(\mathbf{AUC}_{\mathbf{z}}\)) and its final performance. Visual comparison is available in supplementary materials. As shown in \cref{fig:lat_space}, our model's latent space outperforms MegaPortraits in expression representation ability. This is supported by both visual (\cref{fig:comparison_qual}) and quantitative analysis (\cref{tab:comparison_image}).

\begin{figure}[H]
    \centering
    \includegraphics[width=0.9\linewidth]{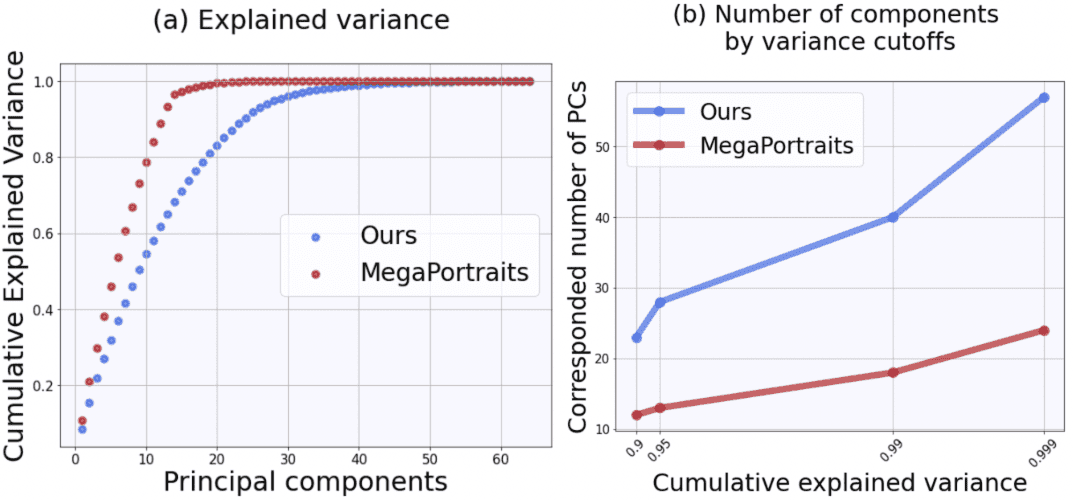}
    \caption{
    Comparison of latent spaces. Left plot shows that our model's latent space is wider and exhibits more even variance distribution. Also, as shown on the right plot, a greater number of principal components are involved in capturing variance across various thresholds. This implies a more robust representational capacity of expression space compared to \cite{drobyshev2023megaportraits}. The VoxCeleb2 test set was used for both plots.}
    \label{fig:lat_space}
\end{figure}

Both metrics shown in \cref{fig:lat_space} were vital in our model's development since they highly correlate with the final model's ability to transfer intense expressions. This was particularly useful for finding adjustments described in \cref{ssec:cv_method} and \cref{ssec:cd_loss}, as early visual results during training are not very telling, but changes in the latent space are noticeable after a few epochs. We also found that in MegaPortraits, just a few principal components significantly affect variance, with only 18 components making up 99\% (\cref{fig:lat_space}), suggesting their 512 $\mathbf{z}$'s dimension is oversized. For our model, we found a 128-dimensional $\mathbf{z}$ is optimal, offering solid performance and reducing overfitting in imbalanced data, as our ablation study reveals. We calculate the Explained Variance (\( \text{EV}_i \)) for the \(i^{th}\) component and (\( \mathbf{AUC}_{\mathbf{z}} \)) using equations shown in \cref{eq:pca}, which involves standardizing the vector set \(\mathcal{Z}\) for PCA and sorting the eigenvalues \(\boldsymbol{\lambda}\) in descending order.


{\small
\begin{equation}
\begin{aligned}
   \mathbf{Z}_{\text{std}} &= \frac{\mathcal{Z} - \overline{\mathcal{Z}}}{\sigma(\mathcal{Z})}, \quad
   \boldsymbol{\lambda} = \text{sorted}_{\text{desc}}\left( \text{eig} \left(\frac{1}{n-1} \mathbf{Z}_{\text{std}}^\top \mathbf{Z}_{\text{std}} \right) \right), \\
   \text{EV}_i &= \frac{\lambda_i}{\sum_{j=1}^{p} \lambda_j}, \quad
   \mathbf{AUC}_{\mathbf{z}} = \frac{1}{d}  \frac{\sum_{i=1}^d \sum_{k=1}^i \lambda_k}{\sum_{j=1}^d \lambda_j},
\end{aligned}
\label{eq:pca}
\end{equation}}

\subsection{Canonical volume}
\label{ssec:cv_method}
\begin{figure}[H]
    \centering
    \includegraphics[width=\linewidth]{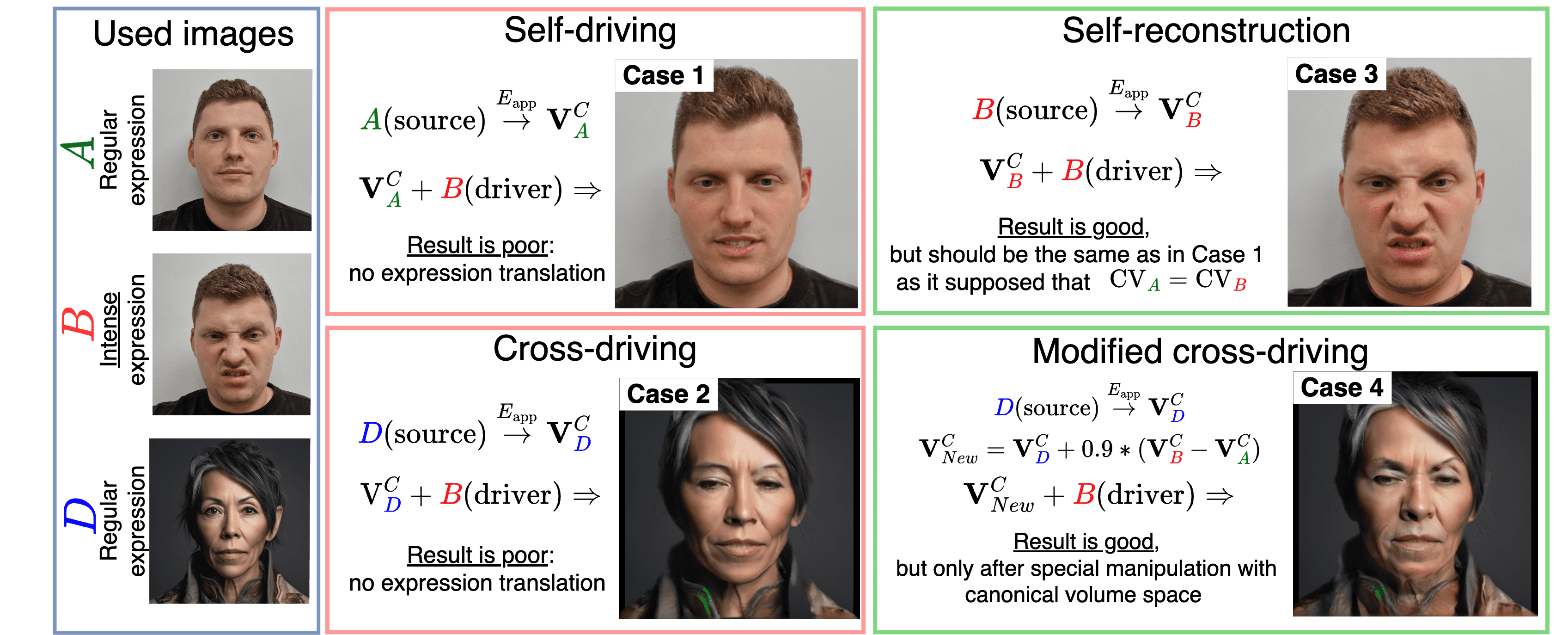}
    \caption{Canonical volumes ($\textbf{V}^C$) in  MegaPortraits \cite{drobyshev2023megaportraits} are not expression-neutral. To show it we use three portraits: A (regular expression), B (intense expression), D (regular expression, new identity). \textit{Case 1} visualizes poor results in transferring B's intense expression to A using a self-driving mode, contrasting with the effective reconstruction of B when used as both driver and source in \textit{case 3}. This discrepancy, indicative of expression leakage into the canonical volume, is further quantified by a 43\% relative difference in $\textbf{V}^C$s between B and C, contrary to expectations of their similarity for same identity. In cross-driving generation \textit{case 2}, using B as the driver and D as the source yielded poor results. However, in \textit{case 4}, after using additive operations on source canonical volumes, the output expression is much closer to driver than in \textit{case 2}. This manipulation again confirms the significant retention of expression information in canonical volumes.}
    \label{fig:cv_problem}
\end{figure}

Despite being designed to exclude face expression details, we found that  canonical volume ($\textbf{V}^C$ in \cref{fig:main_scheme}) in MegaPortraits  actually retains significant expression information from the source image, contributing to poor translation of intense expressions, as shown in our experiment detailed in \cref{fig:cv_problem}. 
In this experiment, using portraits with varied expressions, we assessed the expression translation accuracy and how expression intensity affects the $\textbf{V}^C$s. Our findings suggest that the canonical volumes in MegaPortraits are not truly neutral. As confirmed by our ablation study (\cref{tab:ablation_img_main}), creating an expression-free $\textbf{V}^C$ is essential for precise translation of intense motions. We believe that a more neutral canonical volume improves tractability and effectiveness in expression translation tasks.

To overcome this issue, we propose to match canonical volumes from the different images of the same person ($\textbf{V}^C_{s^n}, \textbf{V}^C_{d^n}$) during training as following:

\begin{equation}
    \mathcal{L}^\text{n}_\text{CV} = \mathcal{L}_\text{MAE} \big(\textbf{V}^C_{s^{n}},  \textbf{V}^C_{d^{n} } \big),
    \label{eq:cv_loss}
\end{equation}

This loss ensures $\textbf{V}^C$ remains stable and expression-independent, crucial for translating intense expressions, as shown in our ablation study (\cref{tab:ablation_img_main}).

\subsection{Source-driver mismatch loss}
\label{ssec:cd_loss}


In addition to maintaining expression-free $\textbf{V}^C$, it's vital to remove all identity information from the expression vector $\z$. While contrastive losses address this in \cite{drobyshev2023megaportraits}, our experiments indicate they are insufficient to prevent overfitting in our imbalanced data scenario, where emotionally intense images represent only 0.1\% of our training set but are sampled 25\% of the time. We introduce a novel self-supervised loss which mitigates identity information in latent expression vectors:

\begin{equation}
\begin{aligned}
\mathcal{L}_{sdm}(\z_{s}, \z_{d}) &=  w*\max(0, \text{cos}(\z_{s}, \z_{d}) - \textit{margin}), \\
w; \textit{margin} &= 
\begin{cases} 
1; 0.5 & \text{if $\z_{s}, \z_{d}$ are from VC2} \\
10; 0.25 & \text{if $\z_{s}, \z_{d}$ are from FEED} 
\end{cases}
\end{aligned}
\label{eq:sdm}
\end{equation}

\begin{figure}[tp]
    \centering
    \includegraphics[width=0.8\linewidth]{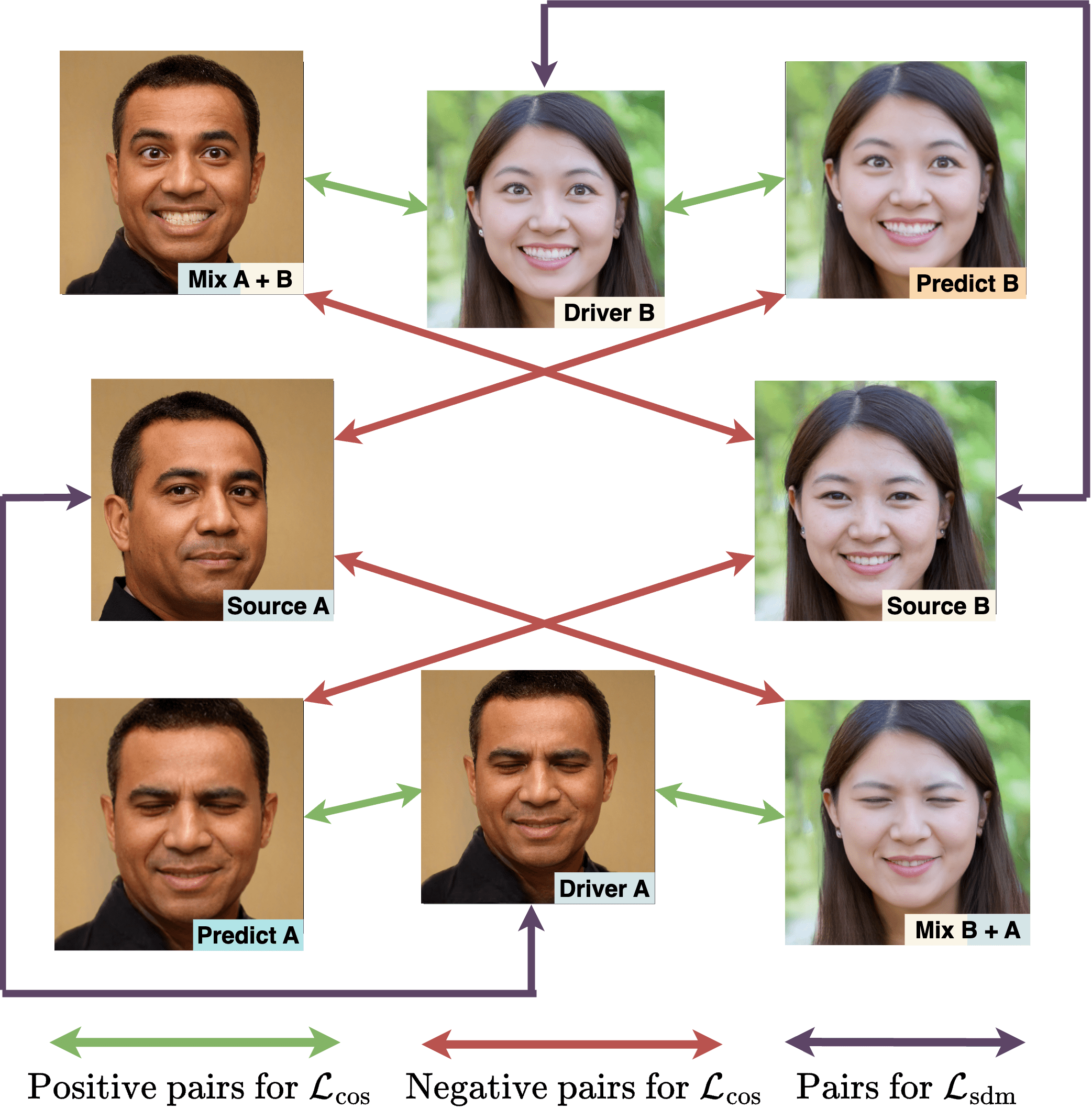}
    \caption{The visualisation of our self-supervised losses. $\mathcal{L}_\text{cos}$ is a contrastive loss, similar to one used in \cite{drobyshev2023megaportraits} but uses different pairs as described in \cref{ssec:full_dis}. Our novel $\mathcal{L}_{sdm}$ is designed to prevent overfitting in an imbalanced dataset.}
    \label{fig:all_cont}
\end{figure}

Here, $\z_{s}, \z_{d}$ are emotion vectors from source and driver images, $w, \textit{margin}$ adjust the loss's intensity and strictness respectively. We increase $w$ for FEED due to its higher sampling rate and decrease \textit{margin} for more assured expression variance in $s, d$. Refer to \cref{fig:all_cont} for a visual summary of all self-supervised losses enhancing our latent emotion space.



\section{Incorporating speech-driving mode}
\label{sec:method_audio}

\subsection{Latent space disentanglement}
\label{ssec:full_dis}

\begin{figure}[tp]
    \centering
    \includegraphics[width=\linewidth]{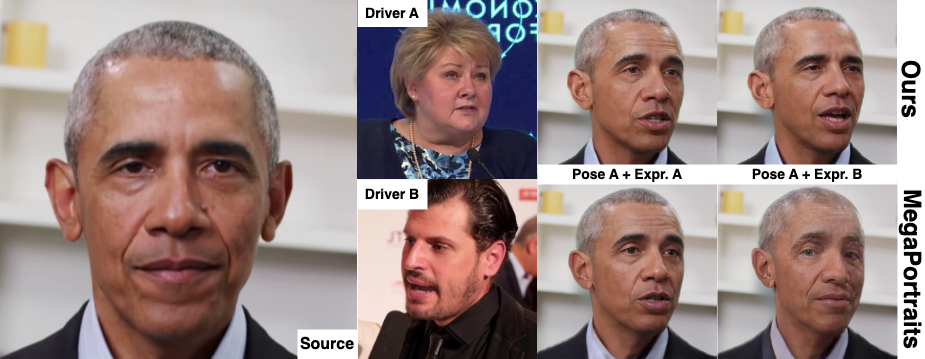}
    \caption{The illustration of disentanglement problem. Images in the right row indicate the expression descriptor $\z$ entangled with head pose for MegaPortraits, whereas we avoid it using changes proposed in \cref{ssec:full_dis}.}
    \label{fig:pose_fix}
\end{figure}

During our latent space analysis described in \cref{sec:method}, we found that combining expression from one driver with the head pose from another resulted in poor performance for MegaPortraits model, as illustrated in \cref{fig:pose_fix} (bottom row). Indeed, the base model lacks a mechanism to intentionally prevent head pose leakage. Disentangled expression latent space, besides expanding the model's use cases, plays the crucial role for the speech-driving mode. When the latent vector $\z$ is entangled with head pose data, predicting it from speech becomes challenging because speech lacks head rotation information, unlike images. Moreover, we would like to have a full control over head rotations during speech mode. We were able to make our latent space disentangled by changing the way we sample images for $\mathcal{L}_\text{cos}$ \cref{eq:cosloss}. 


The MegaPortraits model uses $\mathcal{L}_\text{cos}$ - a modified  \textit{Large Margin Cosine Loss} \cite{wang2018cosface} to mitigate the transfer of appearance features to expression descriptor (latent vector $\mathbf{z}$ on \cref{fig:main_scheme}), which is crucial for cross-driving mode. Without this, differences in appearance between the source and driver, like hairstyle or skin tone, leak from driver to output image.

For computing this loss, the authors utilize a supplementary source-driving pair ($\x_{s^*}$ and $\x_{d^*}$) from a different video with another identity, ensuring a distinct appearance from the current $\x_s$, $\x_d$ pair. The base model is then employed to produce \textit{cross-reenacted} image ($\hat\x_{s^* \rightarrow d} = \G_\text{base}(\x_{s^*}, \x_d)$).  Concurrently, they determine a separate motion descriptor, $\z_{d^*} = \E_\text{motion}(\x_{d^*})$. Descriptors $\z_{* \rightarrow d}$ from the respective forward passes are also used for positive and negative pairs.


Our innovation here is a new sampling strategy: for each $\x_s$, $\x_d$ pair, we sample one more random additional pair $\x_{s^m}$ and $\x_{d^m}$ apart from $\x_{s^*}$ and $\x_{d^*}$. We then apply the model to produce the following  \textit{cross-reenacted moved} image using source identity from $\x_{s^*}$, desired head pose from $\x_{s^m}$ and emotions from $\x_d$, represented as: $\hat\x_{s^{*m} \rightarrow d} = \G_\text{base}(\x_{s^{*m}}, \x_d)$. This creates a positive pair with the same desired emotion but varying head poses.

Motion descriptors are organized into \textit{positive pairs} $\mathcal{P}$ for alignment, and \textit{negative pairs} $\mathcal{N}$ for non-alignment:  $\mathcal{P} = \big\{ ( \z_{s \rightarrow d}, \z_d ), ( \z_{s^* \rightarrow d}, \z_d ), ( \z_{s^{*m} \rightarrow d}, \z_d ) \big\} $, and  $ \mathcal{N} = \big\{ ( \z_{s \rightarrow d}, \z_{d^*} ), ( \z_{s^* \rightarrow d}, \z_{d^*} ), ( \z_{s^{*m} \rightarrow d}, \z_{d^*} ) \big\}$ (see \cref{fig:all_cont}). These pairs are used to calculate the following cosine distance:
\begin{equation}
    d(\z_i, \z_j) = s \cdot \big( \langle \z_i, \z_j \rangle - m \big),
\end{equation}
where both $s$ and $m$ are hyperparameters. This distance is then used to calculate a large margin cosine loss (CosFace)~\cite{Wang2018CosFaceLM}:

\begin{equation}
    \resizebox{.9\hsize}{!}{$
    \mathcal{L}_\text{cos} =
    - \hspace{-0.4cm} \sum\limits_{(\z_k, \z_l) \in \mathcal{P}} \hspace{-0.4cm} \log \dfrac{ \exp \big\{ d(\z_k, \z_l) \big\} }{ \exp \big\{ d(\z_k, \z_l) \big\} + \sum\limits_{(\z_i, \z_j) \in \mathcal{N}} \exp \big\{ d(\z_i, \z_j) \big\} }$}
    \label{eq:cosloss}
\end{equation}

This loss prevents the head pose from leaking into the embeddings as shown in \cref{fig:pose_fix} (upper row)

By doing this, we achieved the capability to interpret $\z_i$ as a latent vector encapsulating the emotional content from an image $\x_i$. This advancement opened the door for an intriguing possibility during inference: the prediction of this vector could potentially be derived from an audio signal instead of a facial image using the motion encoder $\E_\text{motion}$, effectively transforming the model into a speech-driven system. The critical challenge here is aligning the audio input with the pretrained latent space. We addressed this by introducing an additional audio encoder, $\E_\text{aud}$, and employing $\E_\text{motion}$ as a teacher model for this new encoder. A crucial factor in achieving remarkable accuracy in lip synchronization was our ability to isolate specific components within the expression latent space, that are solely responsible for mouth movements. We describe it in the next subsection \cref{ssec:mouth_move}.


\begin{figure}[tp]
    \centering
    \includegraphics[width=\linewidth]{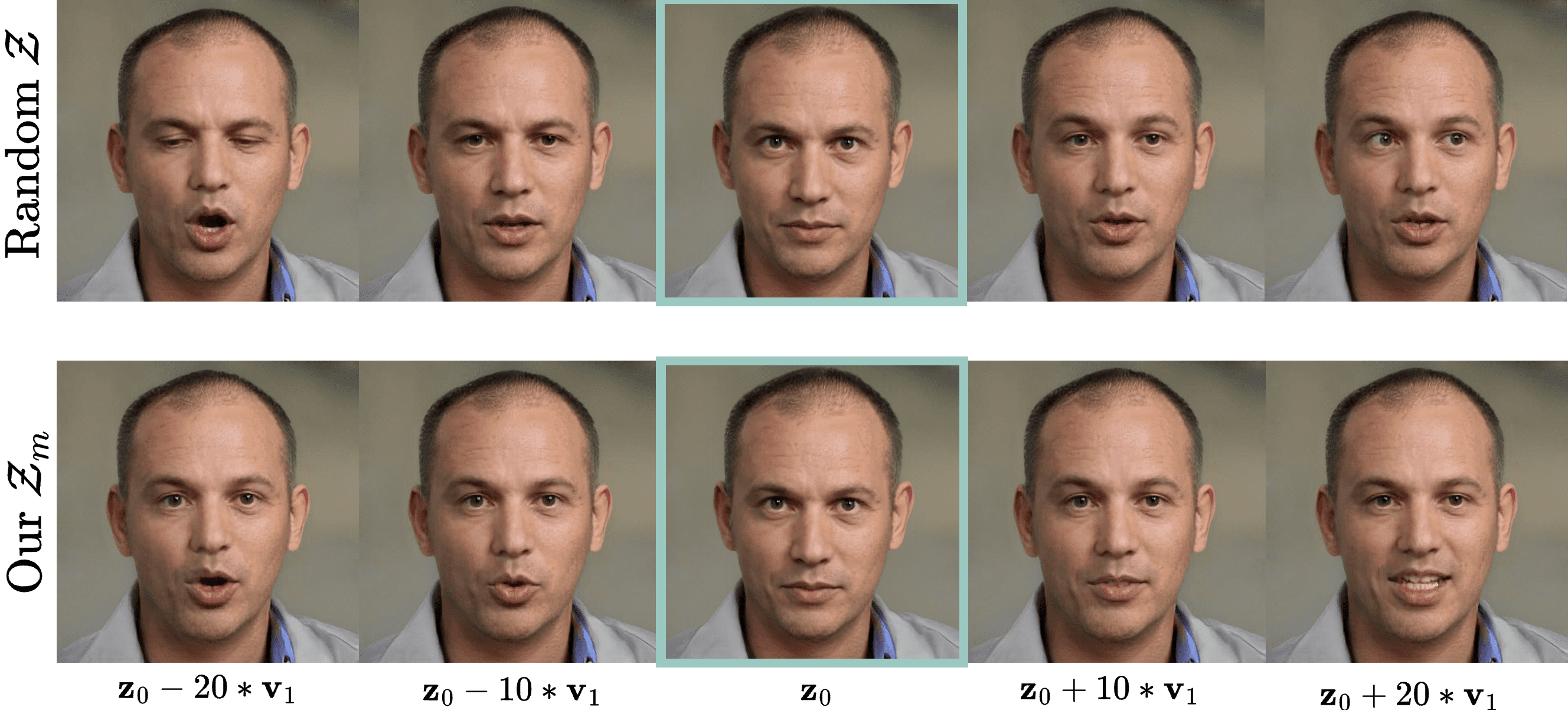}
    \caption{In the upper row, the first principal component from a random subset of expression vectors $\mathcal{Z}$ affects not just the mouth but also blinks and gaze, making it difficult to interpret and isolate mouth movement components, needed for training $\E_\text{aud}$. However, as shown in the bottom row, using $\mathcal{Z}_m$ from \cref{ssec:mouth_move}, the first principal component, $\mathbf{v}_1$, focused on mouth movements, is effectively isolated, $\z_0$ - latent vector, corresponded to the central image.}
    \label{fig:pca_mouth}
\end{figure}

\subsection{Mouth movements}
\label{ssec:mouth_move}

To train the audio encoder $\E_\text{aud}$ for predicting $\z$ from speech, we chose to use the motion encoder $\E_\text{motion}$ as a teacher model. However, direct matching vectors $\z_i^\text{aud}$ from $\E_\text{aud}$ and pseudo-ground truth $\z_i$ from $\E_\text{motion}$ gives poor results (see \cref{tab:ablation_audio_main}). We believe that the reason behind this is that speech-derived $\z_i^\text{aud}$, while able to capture lip movements, struggles with other facial motions, especially in the upper face. Thus, we focused on isolating mouth movement components in the expression latent space using PCA analysis.
 

Employing PCA on a broad array of expression vectors $\mathcal{Z}$, generated from images with varied facial emotions, yields principal components with their explained variances ($\text{EV}_i$). Altering these components, particularly those with high $\text{EV}_i$, significantly changes facial expressions in an image. However, they typically affect a combination of facial features\cref{fig:pca_mouth} (upper row), not isolated areas like the mouth or eyes.

To focus on mouth movements, we created a unique set of $\mathcal{Z}_m$ from just one video of a person performing mouth manipulations and manually edited it to minimize upper face movements. This involved using a still upper face from the first frame in all subsequent frames, ensuring that the principal components mainly represented mouth movements. 
Applying PCA of $\mathcal{Z}_m$ reveals that most principal components are responsible for solely mouth movements. For illustration of the first component, see \cref{fig:pca_mouth} (bottom row). Based on distilled components, we introduce mouth PCA mouth loss:
\begin{equation}
\mathcal{L}_\text{PCA}(v_i, v_j, n) = \frac{1}{n} \sum_{k=1}^{n} \left| \text{PC}_{\mathbf{v}_i}(k) - \text{PC}_{\mathbf{v}_j}(k) \right|
\end{equation}

Where \( n \) is the number of principal components to be considered, \( \text{PC}_{\mathbf{z}_i}(k) \) and \( \text{PC}_{\mathbf{z}_j}(k) \) are the \( k \)-th principal components of vectors \( \mathbf{z}_i \) and \( \mathbf{z}_j \) respectively. Implementing this loss has demonstrably enhanced the quality and accuracy of our results, as our ablation study confirms (see \cref{tab:ablation_audio_main}).


\section{Experiments}
\label{sec:experiments}
In this section, we describe experiments we set for comparison for both our image-driven and speech-driven mode. Please see implementation details such as all used losses, component's sizes and architecture details, training procedure, data preprocess in supplementary materials.

\subsection{Image-driven comparison}
\label{ssec:image_comp}

\textbf{Methods}. We evaluated our image-driven model alongside various one-shot models. For this comparison, we included models like NOFA \cite{yu2023nofa} and StyleHEAT \cite{yin2022styleheat}, which use 3DMM expression blendshapes without disentanglement training, and models with trainable latent face motion representations like FOMM \cite{Siarohin2019FirstOM}, UVA \cite{siarohin2023unsupervised}, MetaPortrait \cite{zhang2023metaportrait} and MegaPortraits \cite{drobyshev2023megaportraits}. It's noteworthy that NOFA and UVA require per-source optimization, unlike the others, including our model.

\textbf{Evaluation metrics and data}. For our evaluation, we use 100 random, non-child, single-person images from the FFHQ dataset as sources. From the MEAD dataset, we chose 100 images covering all high-intensity emotions except neutral, and another 100 from the FEED dataset's extreme emotion task, featuring 5 identities not seen during training of our model. Each source image was paired with 20 driving images from these sets, totaling 2000 pairs.

\begin{figure*}[]
    \centering    
    \newlength{\wid}
    \newlength{\mrg}
    \setlength{\wid}{0.18\textwidth}
    \setlength{\mrg}{-0.3cm}
    \resizebox{0.93\linewidth}{!}{
    \begin{tabular}{cccccccccc}
        \includegraphics[width=\wid]{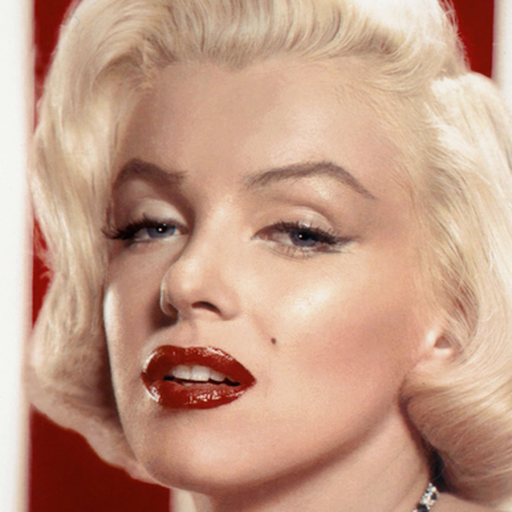} & 
        \hspace{\mrg}
        \includegraphics[width=\wid]{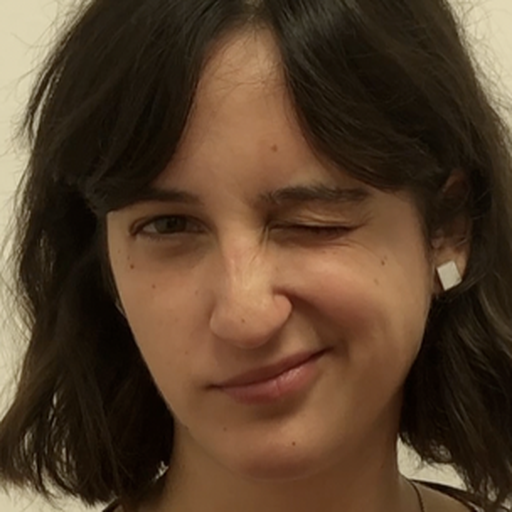} & 
        \hspace{\mrg}
        \includegraphics[width=\wid]{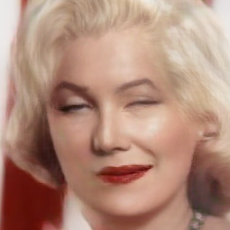} & 
        \hspace{\mrg}
        \includegraphics[width=\wid]{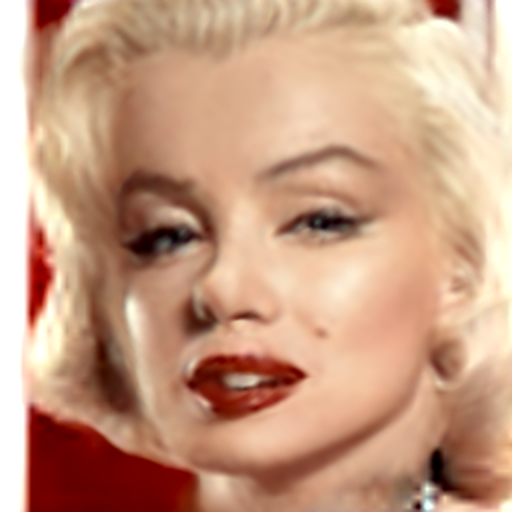} & 
        \hspace{\mrg}
        \includegraphics[width=\wid]{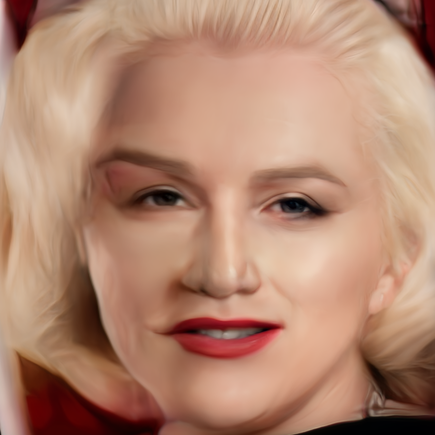} & 
        \hspace{\mrg}
        \includegraphics[width=\wid]{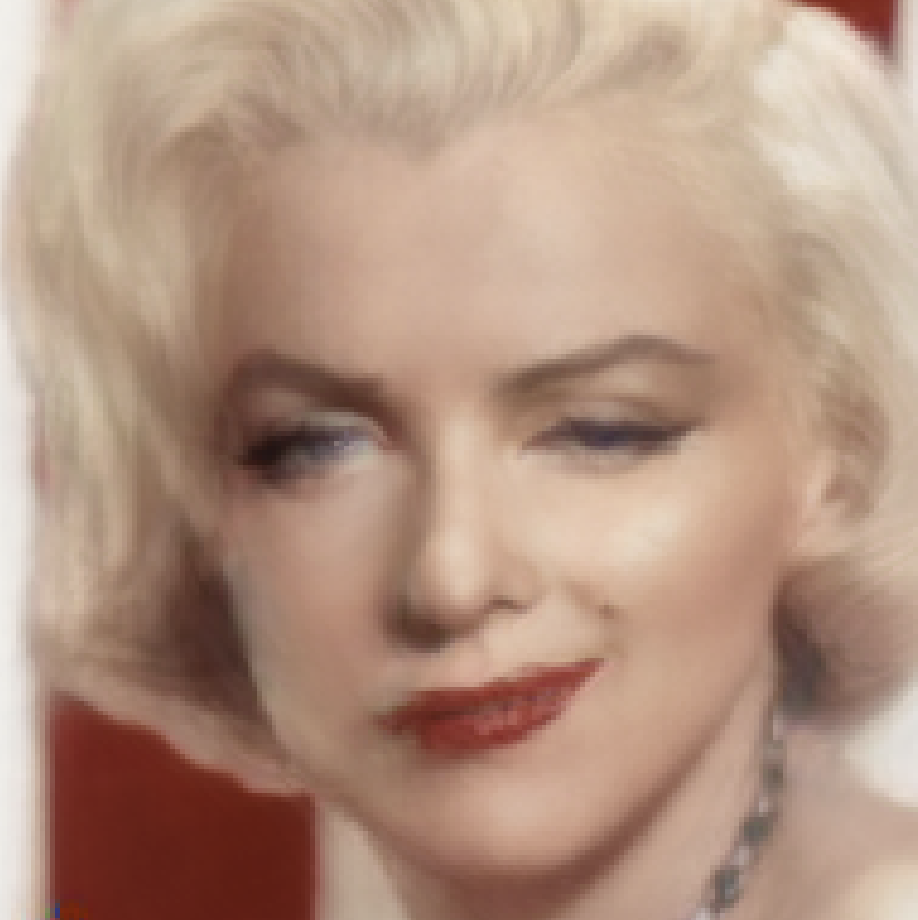} & 
        \hspace{\mrg}
        \includegraphics[width=\wid]{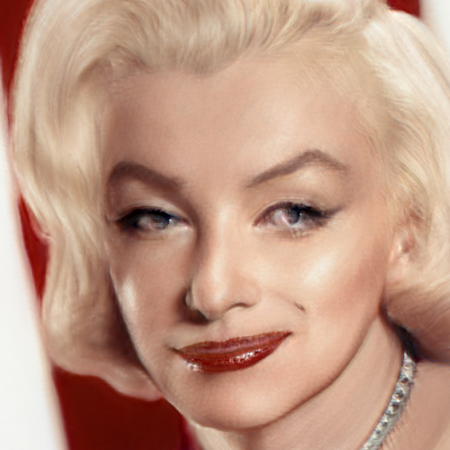} & 
        \hspace{\mrg}
        \includegraphics[width=\wid]{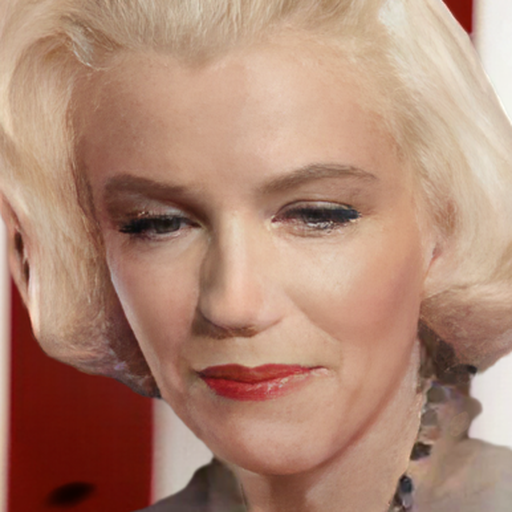} &
        \hspace{\mrg}
        \includegraphics[width=\wid]{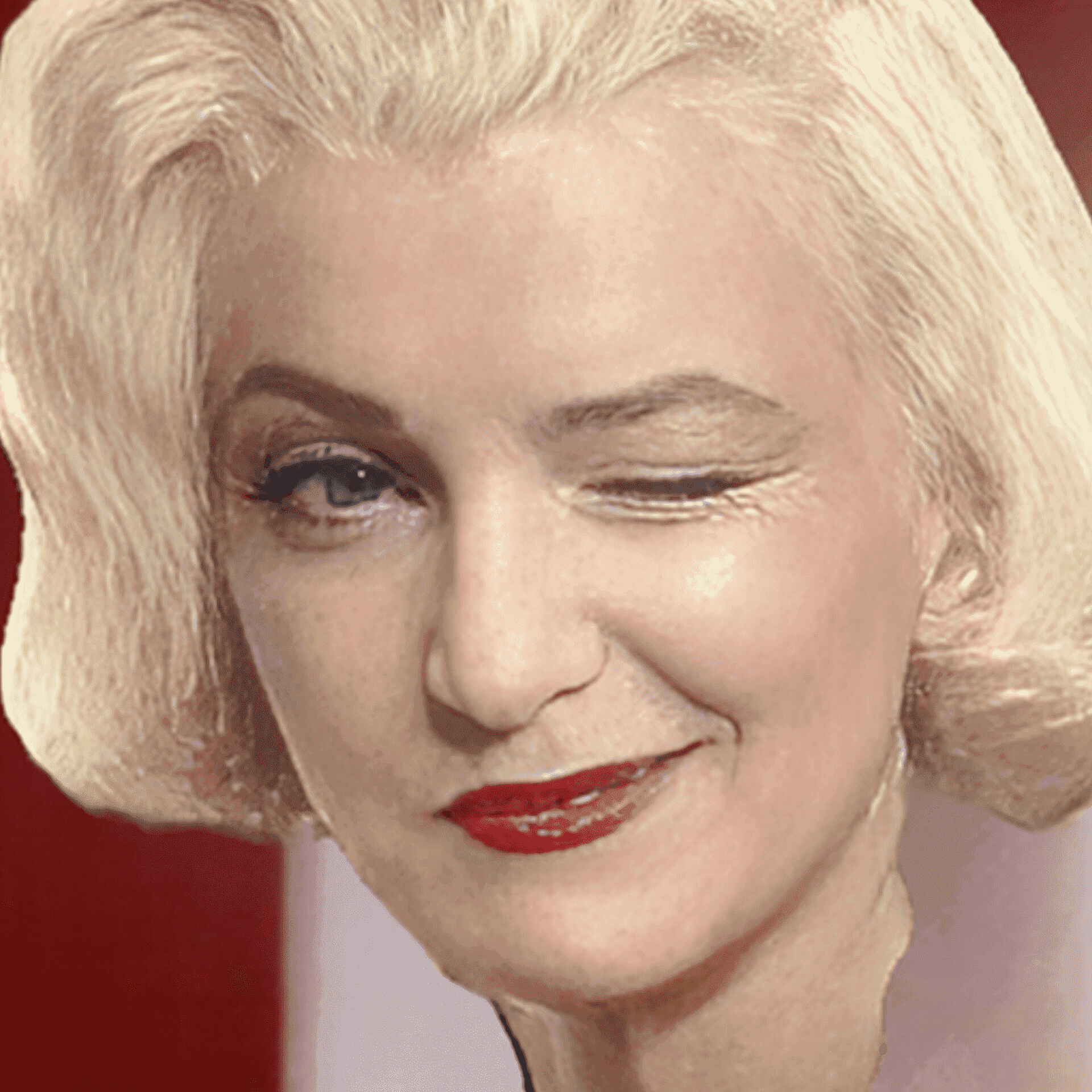} & 
        \\ %
        \includegraphics[width=\wid]{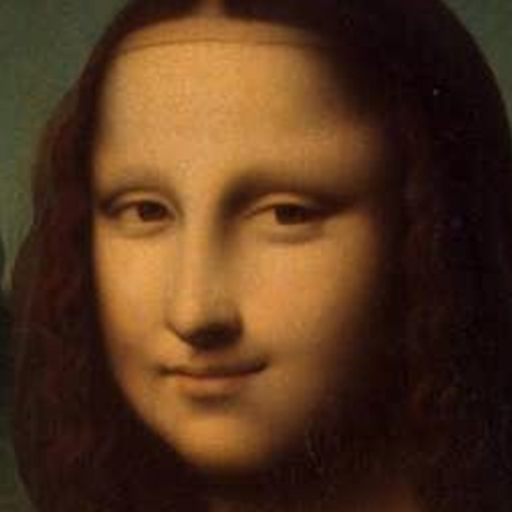} & 
        \hspace{\mrg}
        \includegraphics[width=\wid]{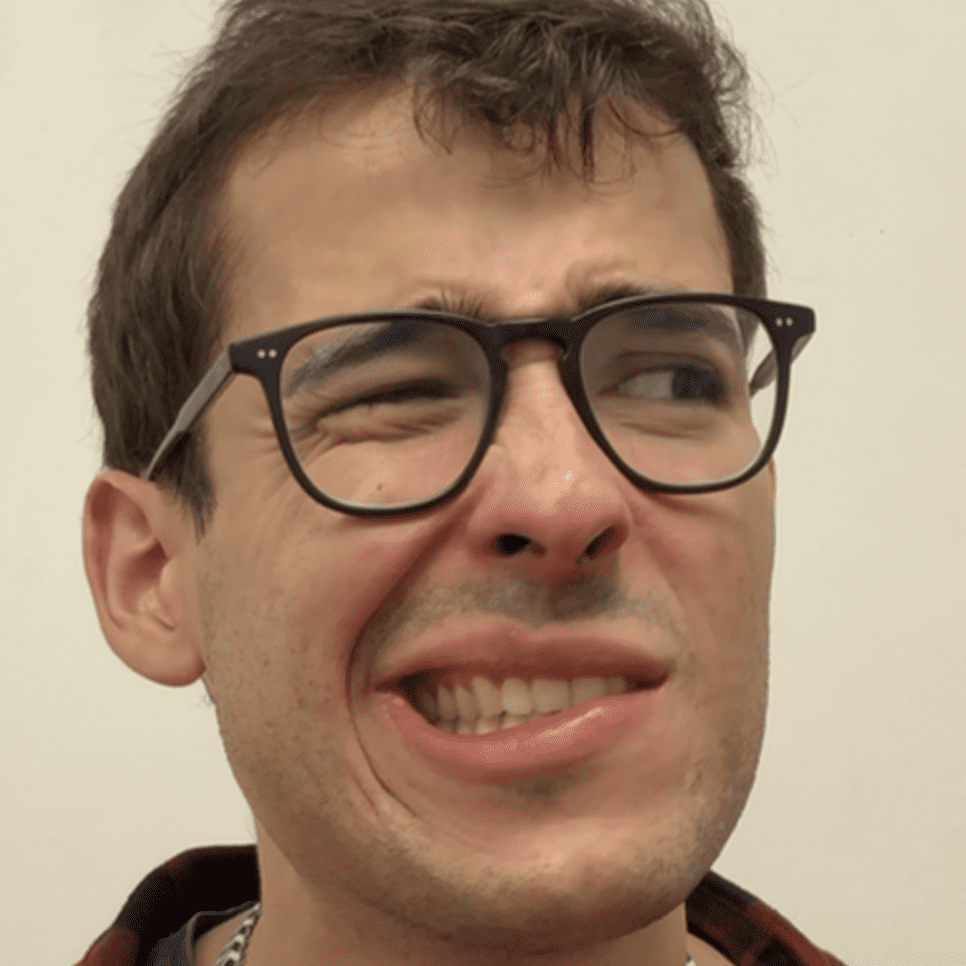} & 
        \hspace{\mrg}
        \includegraphics[width=\wid]{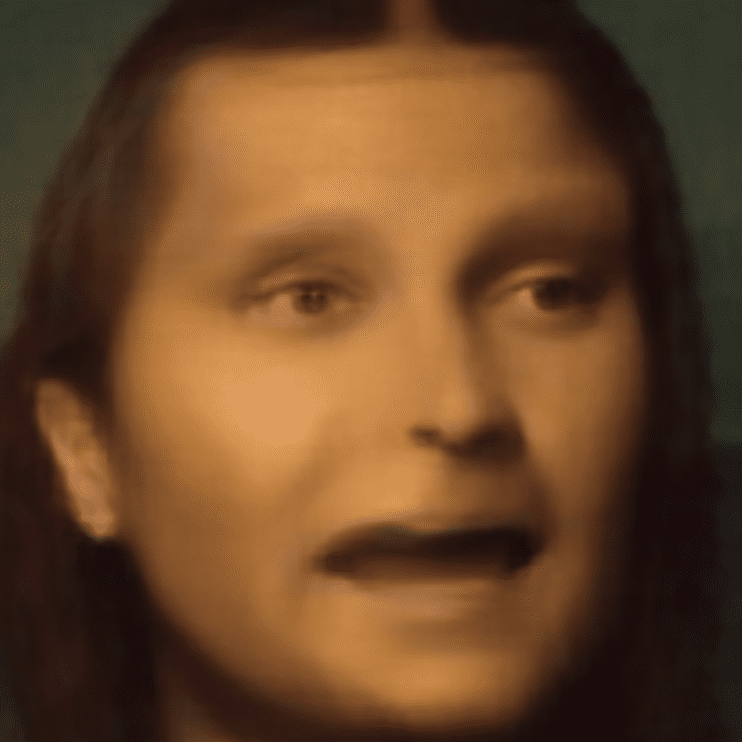} & 
        \hspace{\mrg}
        \includegraphics[width=\wid]{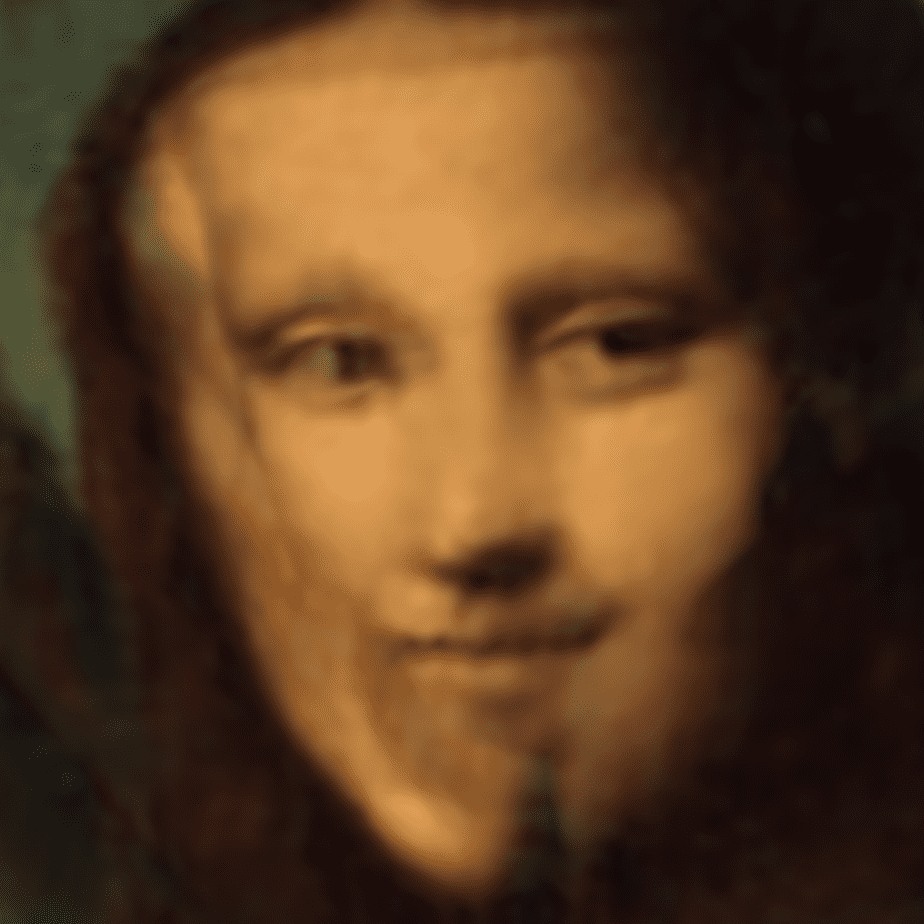} & 
        \hspace{\mrg}
        \includegraphics[width=\wid]{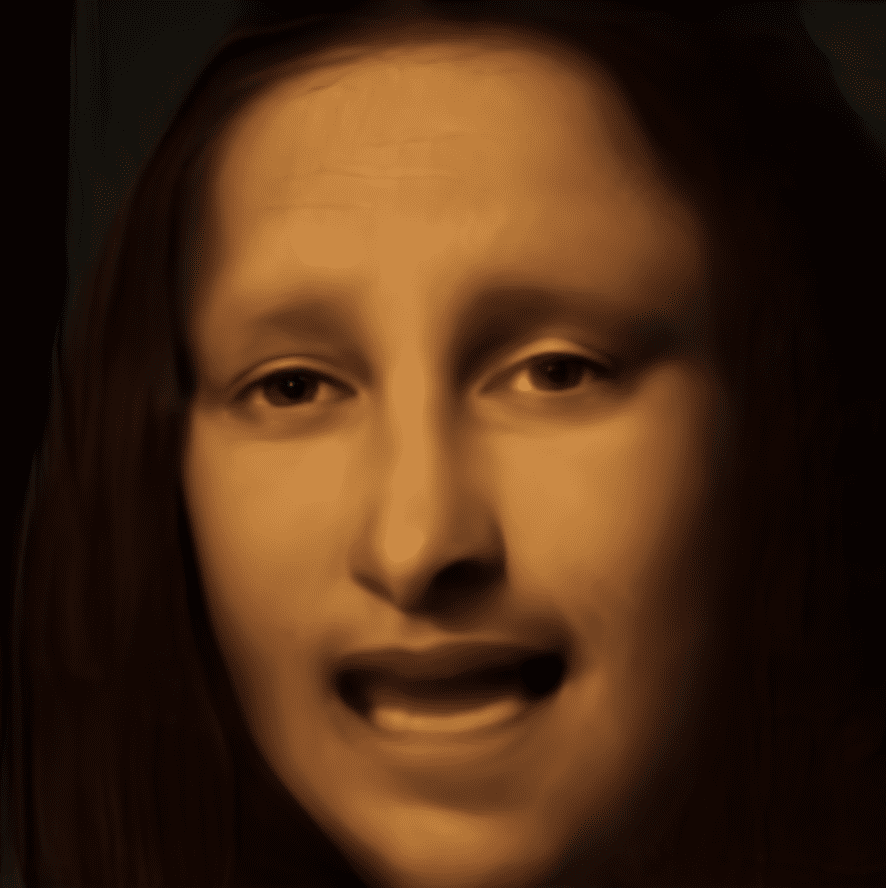} & 
        \hspace{\mrg}
        \includegraphics[width=\wid]{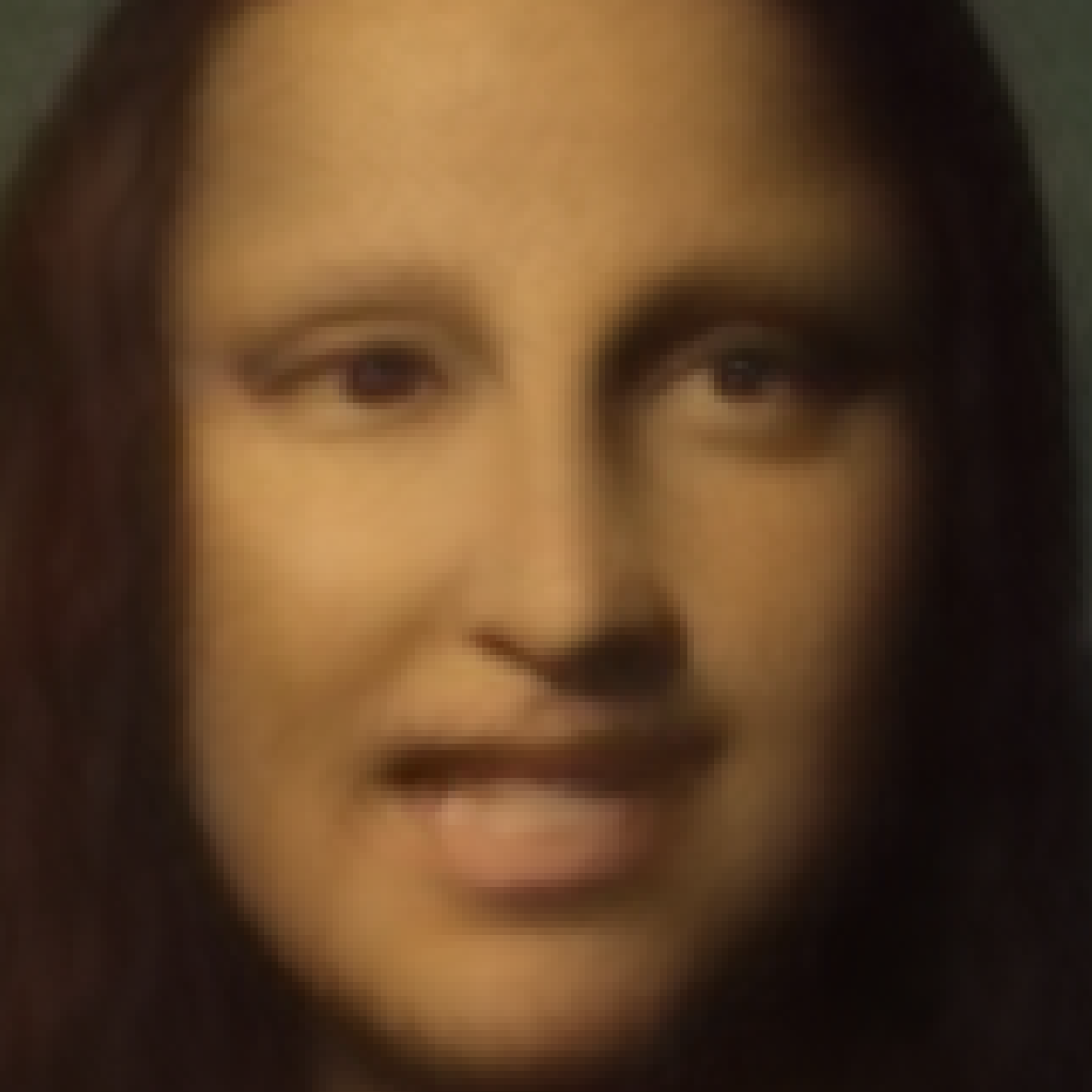} & 
        \hspace{\mrg}
        \includegraphics[width=\wid]{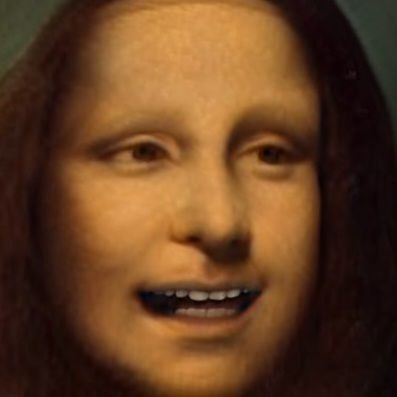} & 
        \hspace{\mrg}
        \includegraphics[width=\wid]{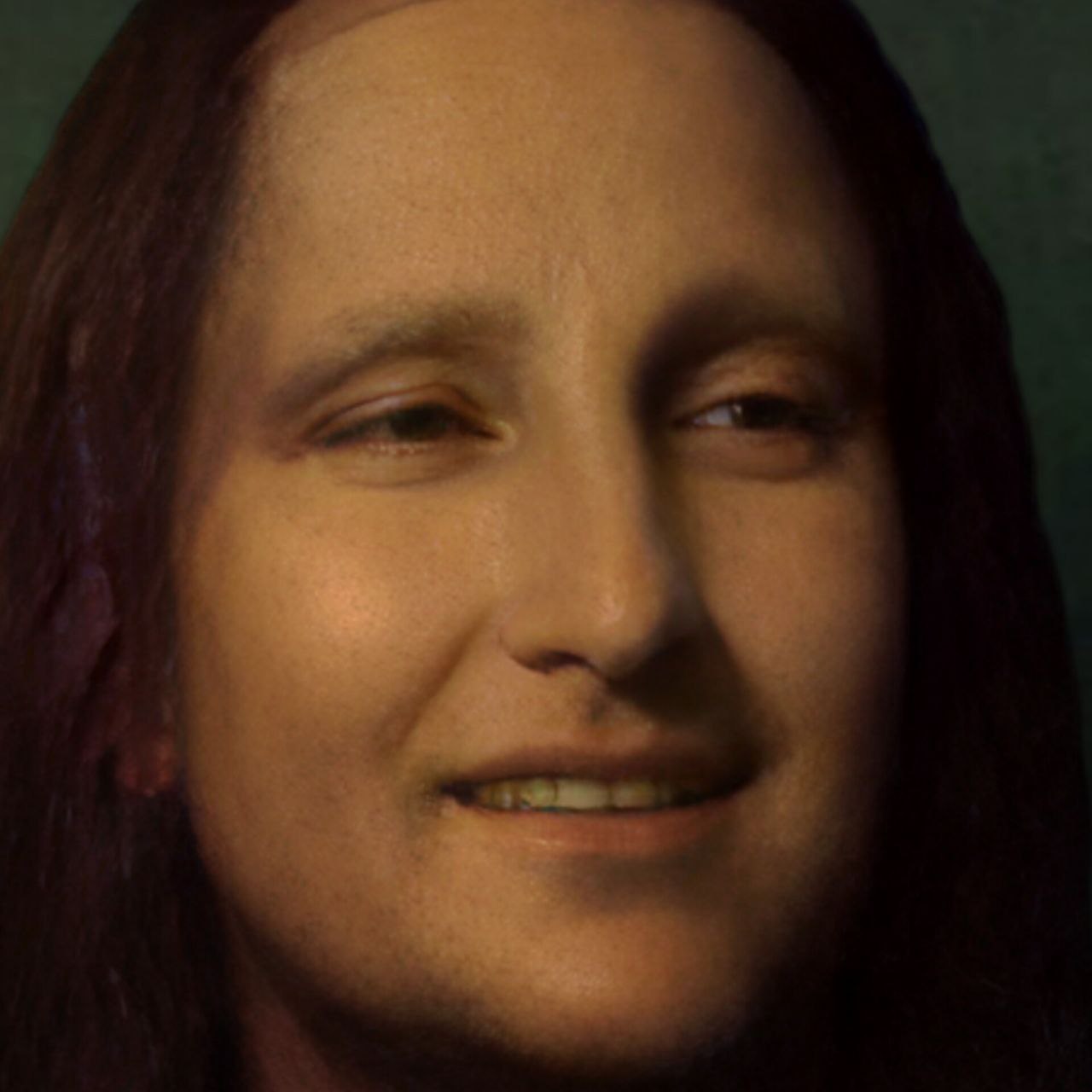} &
        \hspace{\mrg}
        \includegraphics[width=\wid]{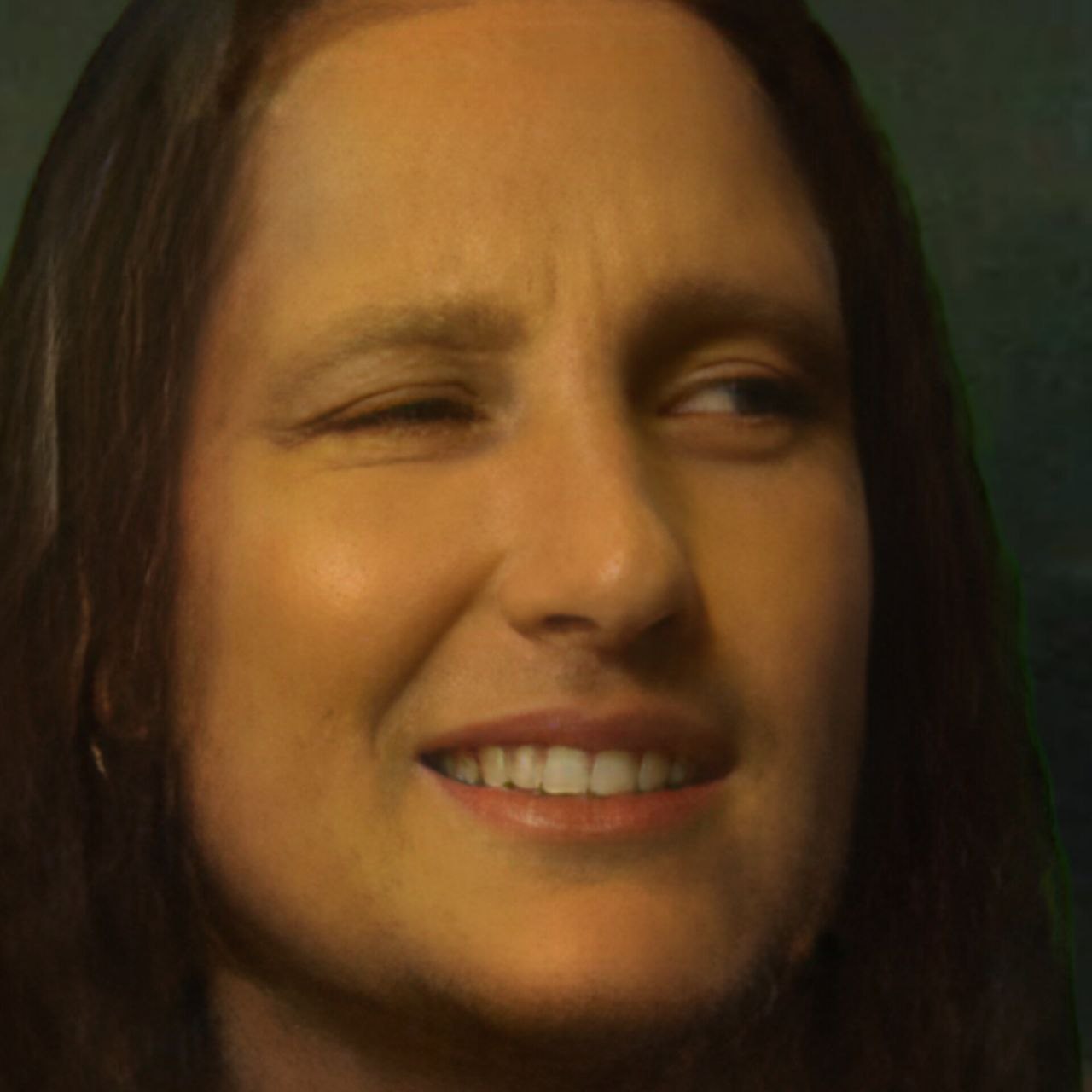} & 
        \\
        \includegraphics[width=\wid]{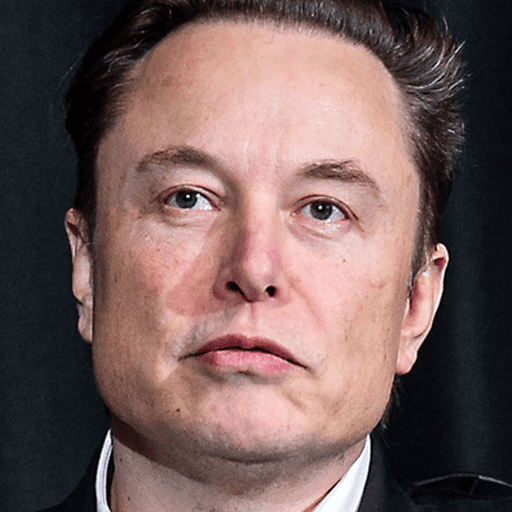} & 
        \hspace{\mrg}
        \includegraphics[width=\wid]{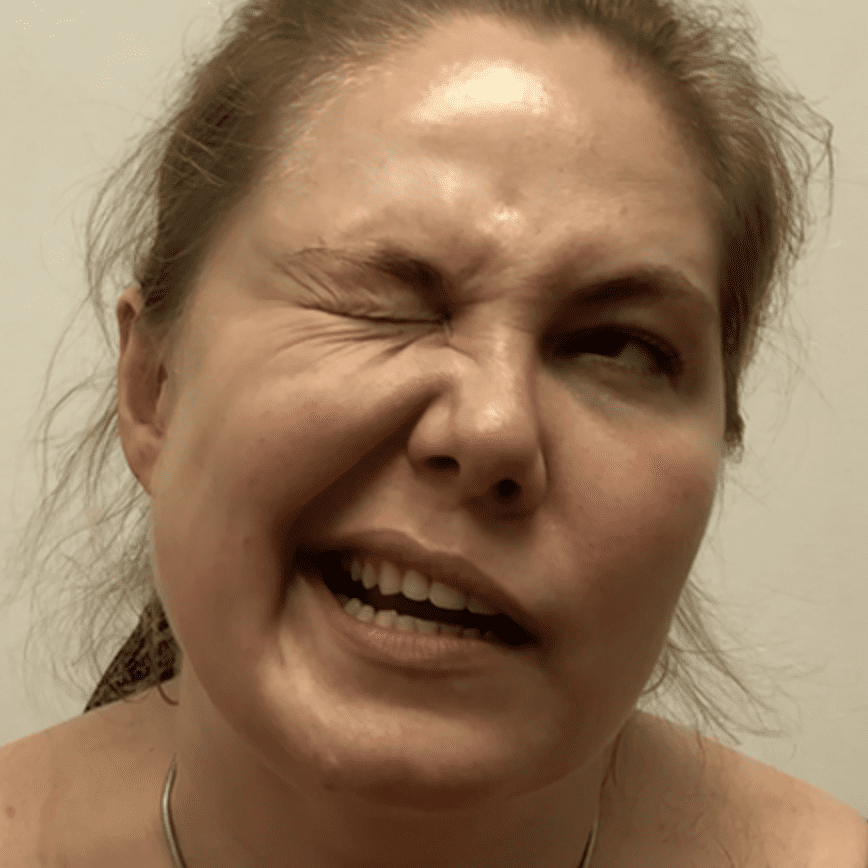} & 
        \hspace{\mrg}
        \includegraphics[width=\wid]{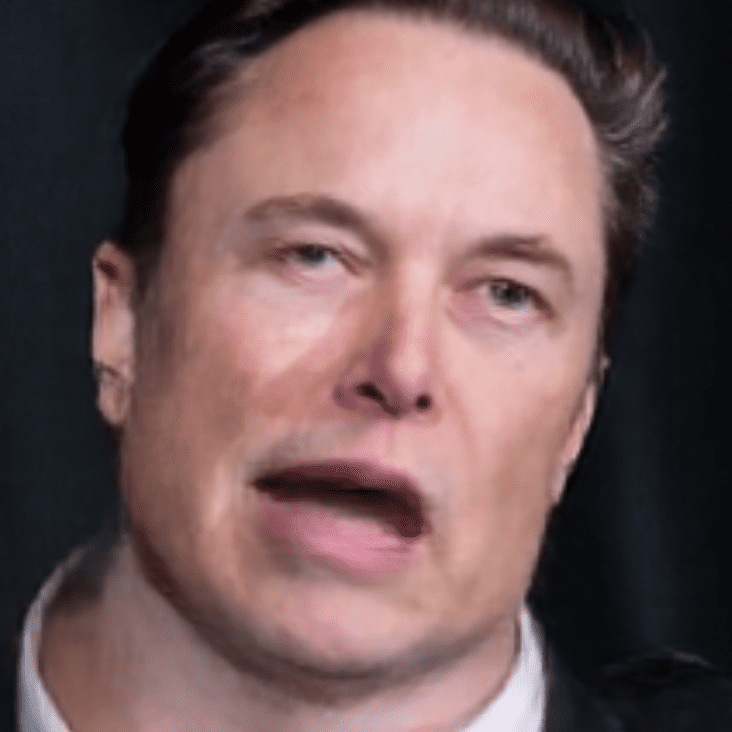} & 
        \hspace{\mrg}
        \includegraphics[width=\wid]{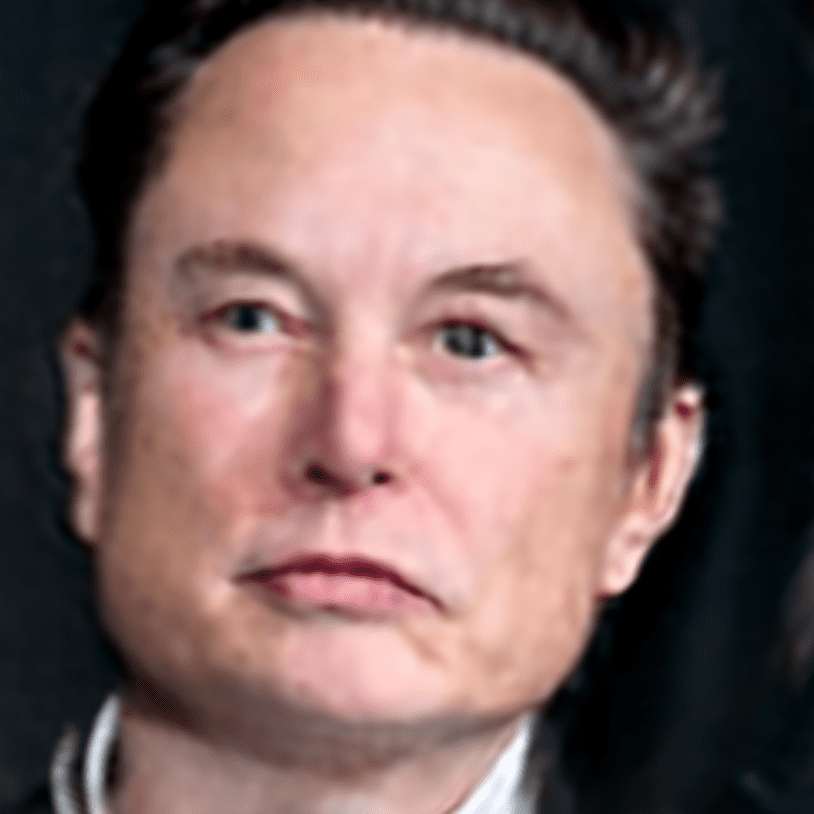} & 
        \hspace{\mrg}
        \includegraphics[width=\wid]{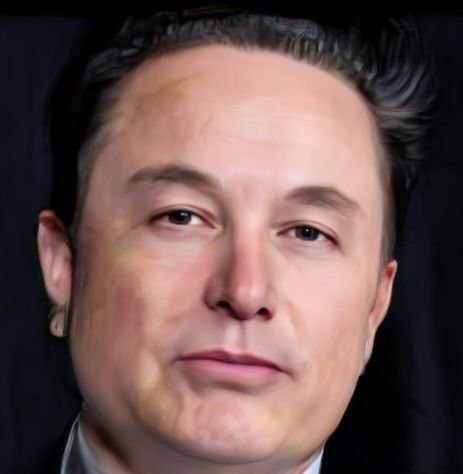} & 
        \hspace{\mrg}
        \includegraphics[width=\wid]{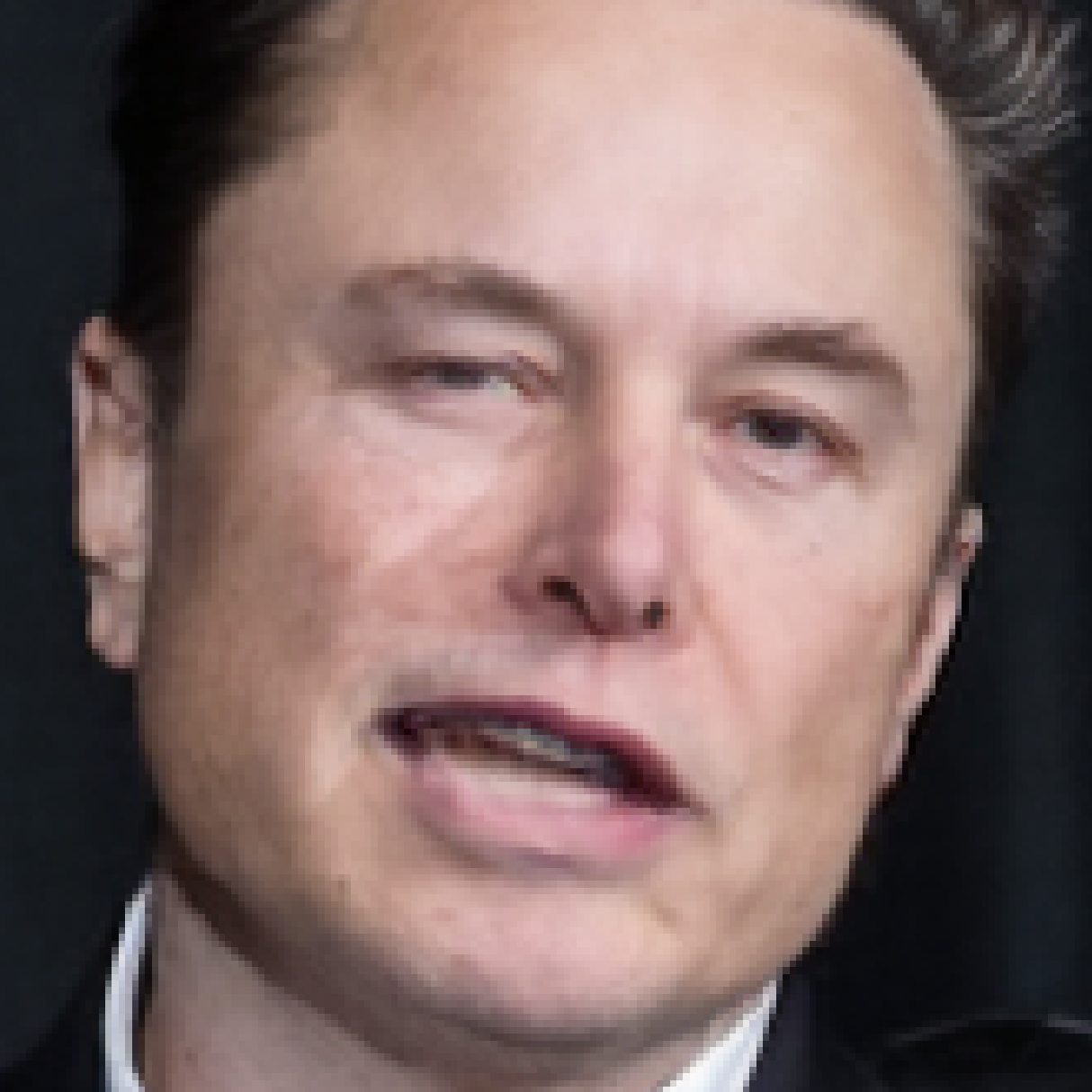} & 
        \hspace{\mrg}
        \includegraphics[width=\wid]{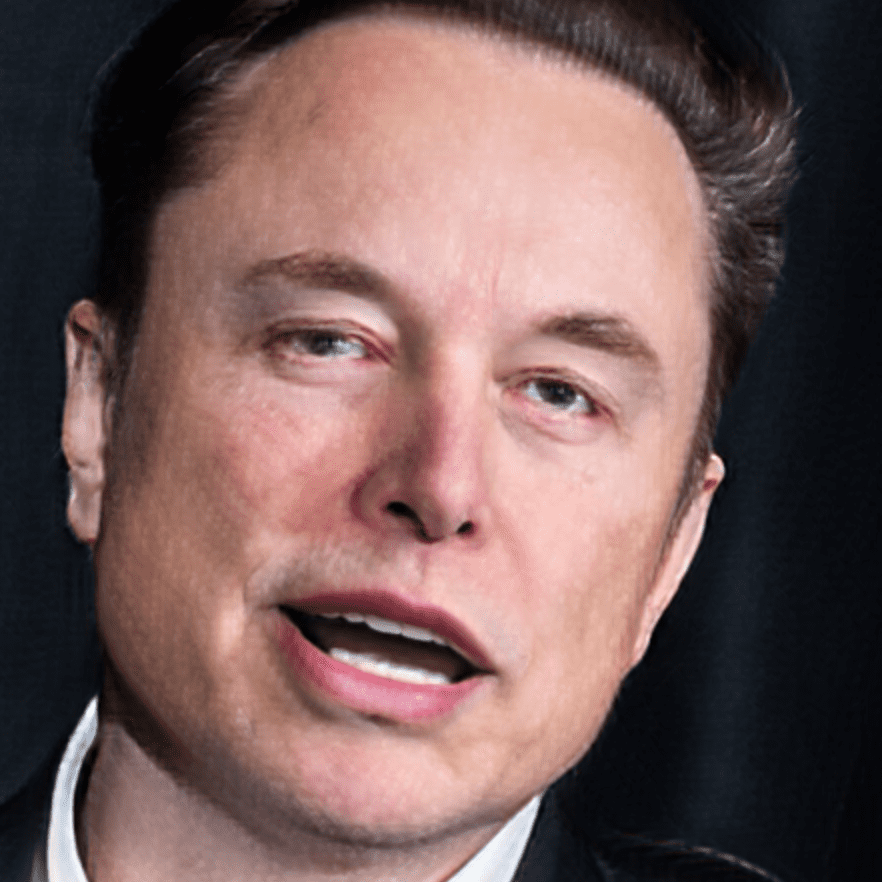} & 
        \hspace{\mrg}
        \includegraphics[width=\wid]{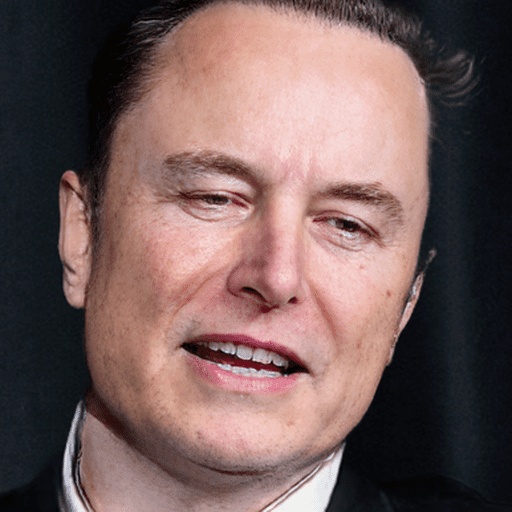} &
        \hspace{\mrg}
        \includegraphics[width=\wid]{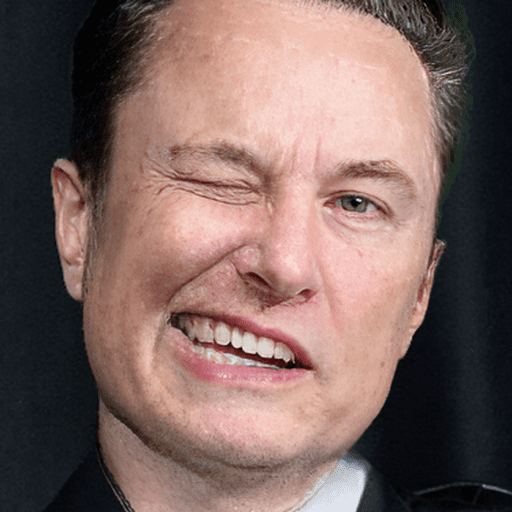} & 
        \\
        \includegraphics[width=\wid]{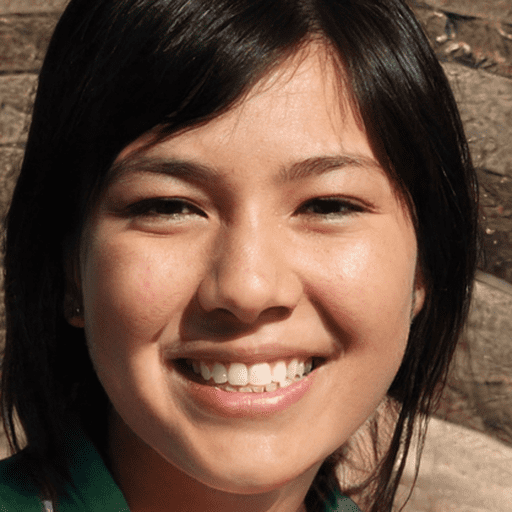} & 
        \hspace{\mrg}
        \includegraphics[width=\wid]{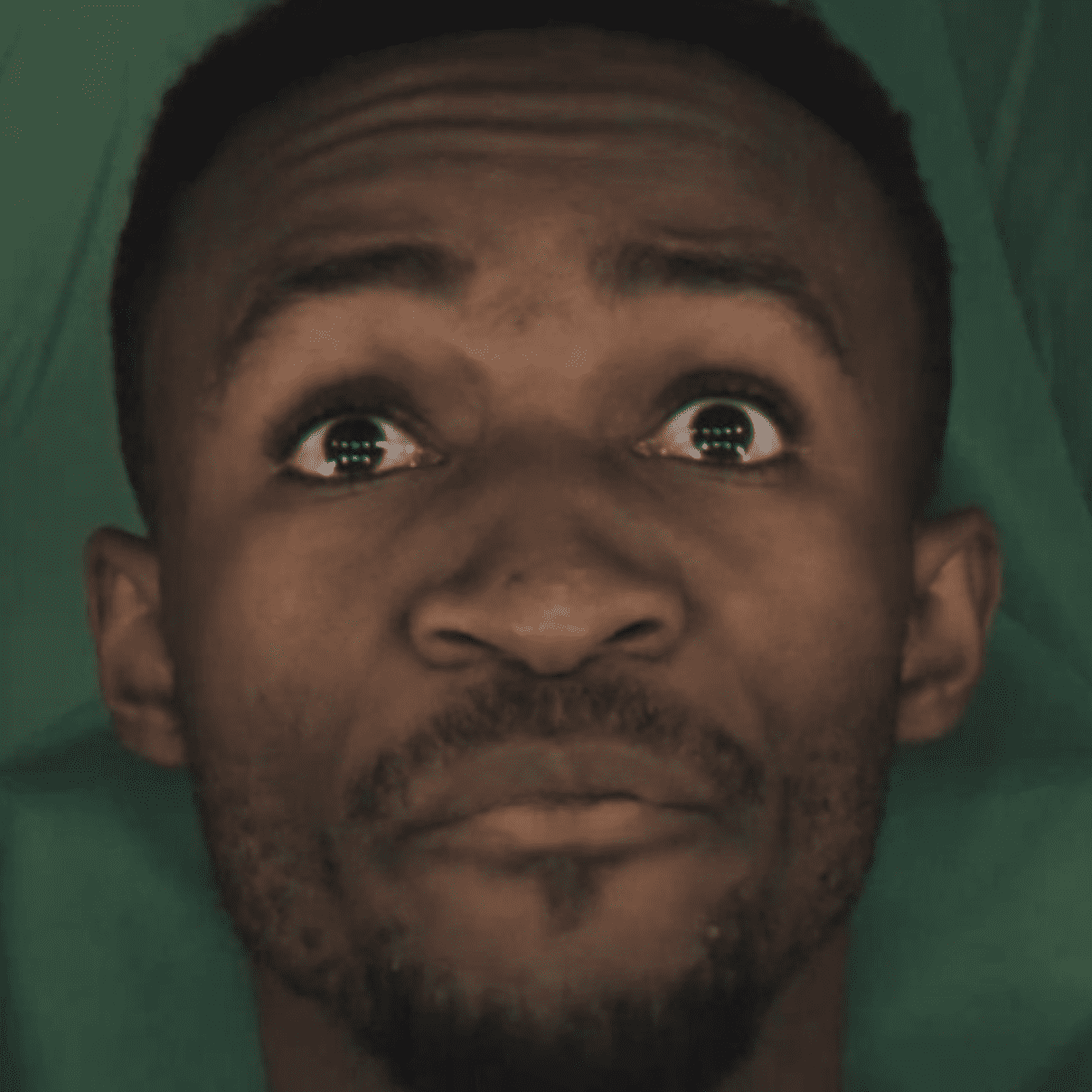} & 
        \hspace{\mrg}
        \includegraphics[width=\wid]{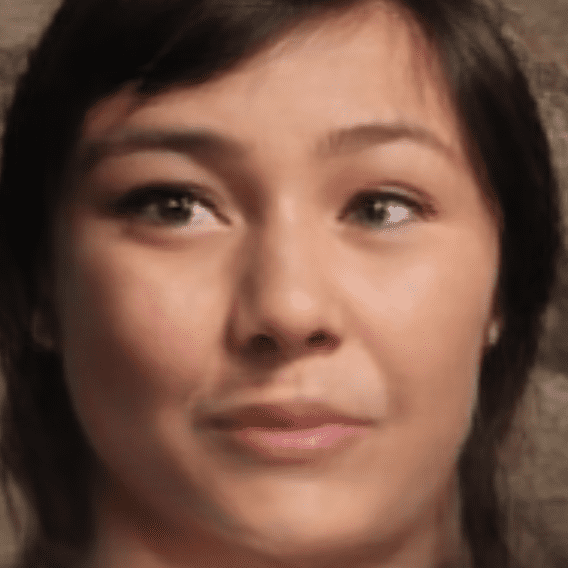} & 
        \hspace{\mrg}
        \includegraphics[width=\wid]{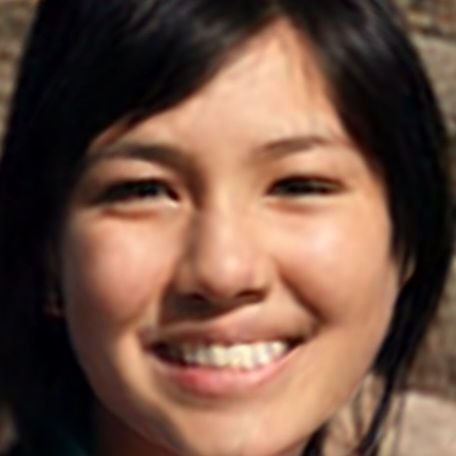} & 
        \hspace{\mrg}
        \includegraphics[width=\wid]{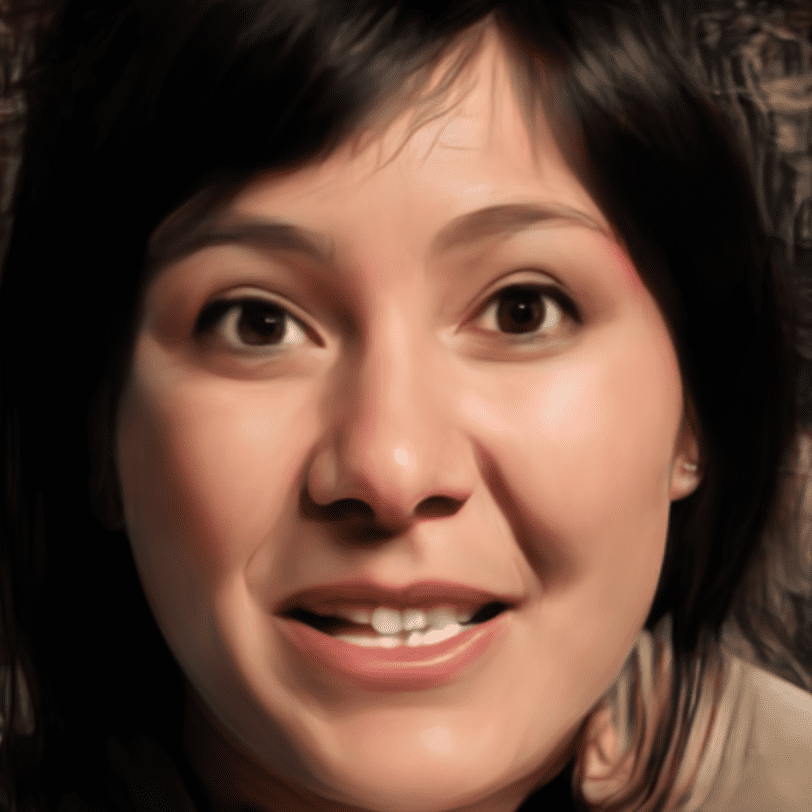} & 
        \hspace{\mrg}
        \includegraphics[width=\wid]{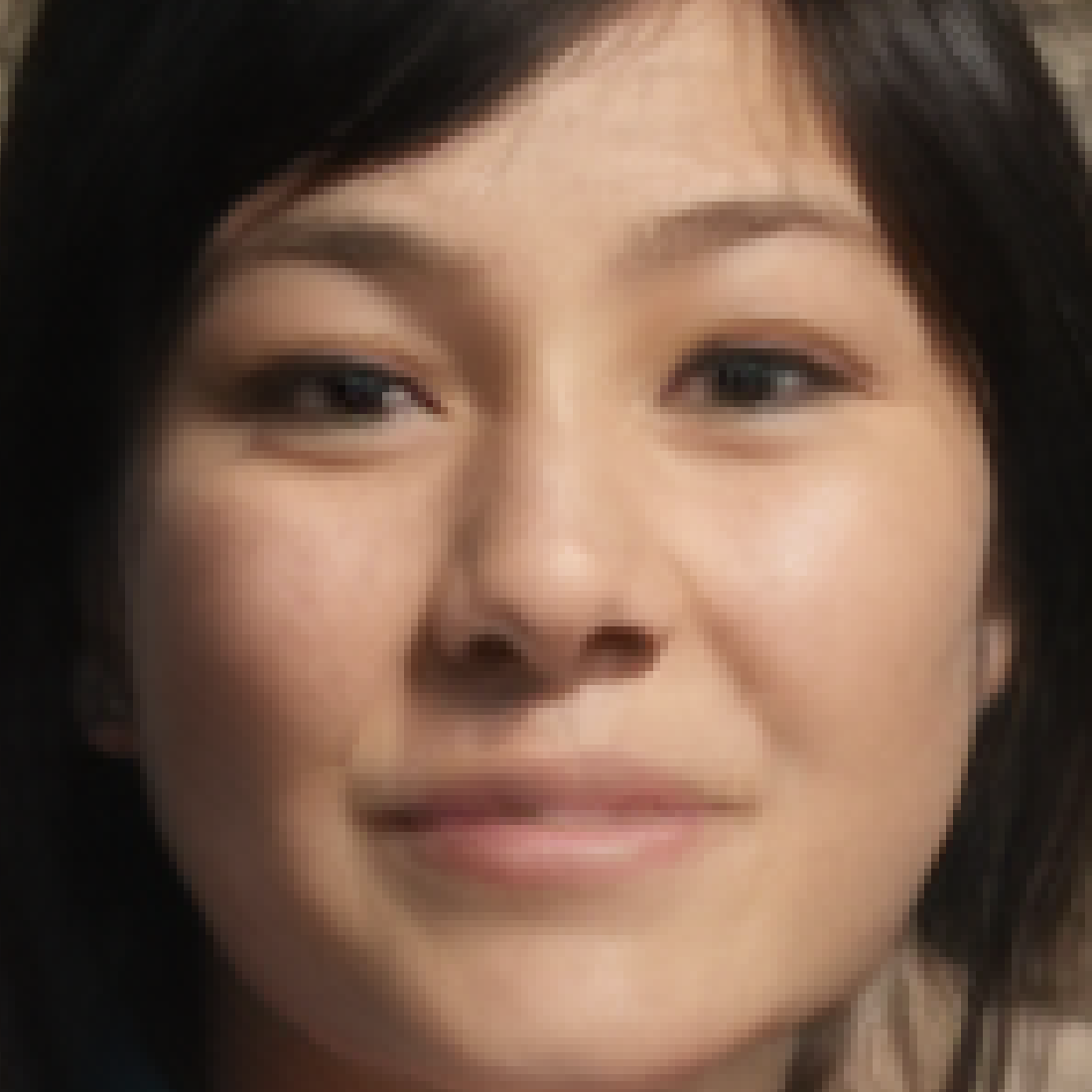} & 
        \hspace{\mrg}
        \includegraphics[width=\wid]{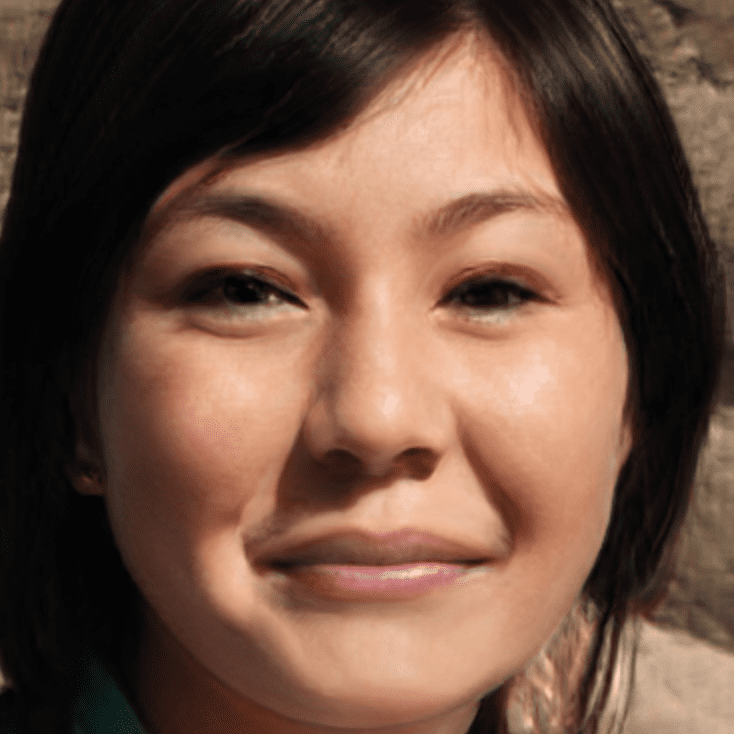} & 
        \hspace{\mrg}
        \includegraphics[width=\wid]{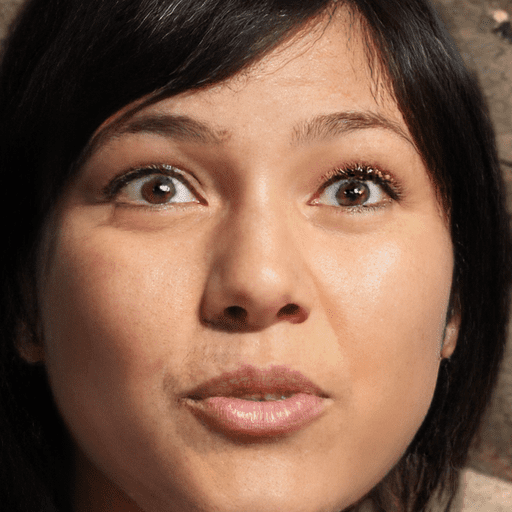} &
        \hspace{\mrg}
        \includegraphics[width=\wid]{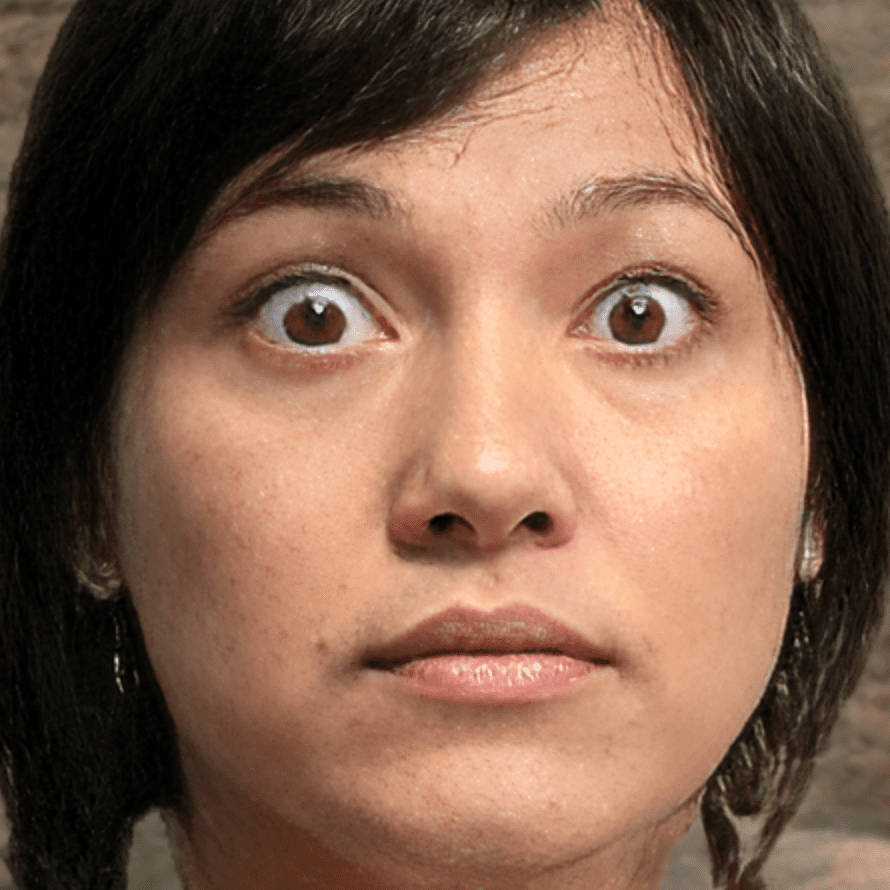} &  
        \\

        \textbf{Source} & 
        \hspace{\mrg} 
        \textbf{Driver} & 
        \hspace{\mrg} 
        \textbf{FOMM}~\cite{Siarohin2019FirstOM} & 
        \hspace{\mrg} 
        \textbf{UVA}~\cite{li2023generalizable} & 
        \hspace{\mrg} 
        \textbf{StyleHEAT}~\cite{yin2022styleheat} &
        \hspace{\mrg} 
        \textbf{MetaPortrait}~\cite{zhang2023metaportrait} &
        \hspace{\mrg} 
        \textbf{NOFA}~\cite{yu2023nofa} &
        \hspace{\mrg} \textbf{MegaPortraits}~\cite{drobyshev2023megaportraits} &
        \hspace{\mrg}
        \textbf{Ours}
    \end{tabular}
    }
    \vspace{-0.2cm}
    \caption{A qualitative comparison of head avatar systems in cross-reenactment scenario.}
    \label{fig:comparison_qual}
\end{figure*}

For assessing reenactment quality, we employ various metrics including the Frechet Inception Distance (FID) \cite{fid} to measure the distributional discrepancy between synthetic and real images. Cosine similarity (CSIM) from a face recognition network \cite{cao2018vggface2} quantifies the identity preservation in generated images. Additionally, we conducted user studies to determine preferences regarding motion (UMTN) and appearance (UAPP) preservation, presenting participants with triplets—either a driving or source image and two method outputs—and asking them to choose the one with better motion or appearance preservation. Our user study includes 34 participants and around 300 unique questions. Both metrics numerically reflect preference percentages. As our visual results \cref{fig:comparison_qual}, and quantitative metrics  \cref{tab:comparison_image} show, our method demonstrates exceptional ability in translating strong and extreme expressions, notably surpassing other image-driven avatar techniques in user preference for facial expression translation (UMTN) and FID scores.

\subsection{Speech-driven comparison}

\textbf{Methods}. To assess our model's speech-driven capabilities, we compared it with StyleHEAT \cite{yin2022styleheat}, which also operates using both image and audio inputs, and audio-exclusive models including SadTalker \cite{zhang2023sadtalker}, MakeItTalk \cite{Zhou_2020}, and PC-AVS \cite{zhou2021posecontrollable}.

\textbf{Evaluation metrics and data}. We employ the evaluation procedure described in \cite{gururani2022space}. We measure video realism using Frechet Inception Distance (FID) \cite{fid}. We also extract facial landmarks from outputs of all methods, frontalized, and normalized to position the mouth edges at (-1, 0) and (1, 0). We compute mean absolute error (MAE) for both predicted mouth landmark positions ($M_P$) and velocities ($M_V$), along with errors in facial geometry ($F_P$) and velocity ($F_V$). For our evaluation, we use 100 randomly selected video sequences from the HDTF dataset \cite{zhang2021flow}, each up to 30 seconds. As the results show in \cref{tab:comparison_audio}, our speech-driven model performs on par with leading methods, excelling in realism and facial dynamics. SyncNet \cite{chung2017out} is used to assess lip sync quality, providing offset (temporal misalignment between the audio and video) and confidence (certainty of audio-visual alignment) scores. 

\begin{table}[t]
    \setlength{\tabcolsep}{6.0pt}
    \footnotesize
    \centering
    \resizebox{\linewidth}{!}{
    \begin{tabular}{c|cccc|cccc}
        \toprule
        Driver: &\multicolumn{4}{c}{MEAD} & \multicolumn{4}{c}{FEED} \\
        \midrule
        Method & FID$\downarrow$ & CSIM$\uparrow$ & UMTN$\uparrow$ & UAPP$\uparrow$ & FID$\downarrow$ & CSIM$\uparrow$ & UMTN$\uparrow$ & UAPP$\uparrow$ \\
        \midrule
         MetaPortrait \cite{zhang2023metaportrait}  & 77.5 & 0.64 & * & *  & 78.8 & 0.66 & * & * \\
         NOFA \cite{yu2023nofa}  & 69.2 & 0.65 & * & *  & 69.8 & 0.66 & 3.8 & 8.8 \\
         FOMM \cite{Siarohin2019FirstOM} & 84.4 & 0.56 & 2.4 & 0.3 & 85.9 & 0.55 & 1.8 & 0.5    \\
         UVA \cite{li2023generalizable}  & 78.7 & 0.68 & 0.9 & 6.4 & 79.5 & 0.68 & 0.6 & 9.4 \\
         StyleHEAT \cite{yin2022styleheat}  & 71.9 & 0.63 & 3.7 & 3.3 & 72.4 & 0.61 & 2.6 & 3.8 \\

         MegaPortraits \cite{drobyshev2023megaportraits} & \underline{61.1} & \underline{0.73} & \underline{22.7} & \textbf{46.6}  & \underline{62.8} & \textbf{0.73} & \underline{16.7} & \textbf{42.8}  \\
         \midrule
         Our (EMOPortraits) & \textbf{59.6} & \textbf{0.74}  & \textbf{70.3} & \underline{43.4} & \textbf{60.2} & \underline{0.70} & \textbf{74.6} & \underline{34.7}  \\
         \bottomrule
    \end{tabular}
    }
    \vspace{-0.3cm}
    \caption{ Our method notably outperforms others in the FID score and strongly leads in the user preference metric for face expression translation (UMTN). It excels in reliably translating strong and extreme expressions from the driving image, distinguishing it as the most capable among the compared methods. Additionally, our approach maintains the source image's shape and appearance (measured by CSIM and UAPP metrics) on par with MegaPortraits \cite{drobyshev2023megaportraits} when using strong and regular expressions from the MEAD dataset as the driving signal. However, we observe a minor deviation in identity preservation metrics for extreme emotions (using the FEED dataset). This difference may arise because our method attempts to transfer asymmetric facial expressions, while MegaPortraits might modify the source image less, or not at all, with a challenging driver (as depicted in \cref{fig:cv_problem}, case 1), thereby achieving better identity preservation. UVA was excluded from the MEAD-driven user study due to the authors were able to provide only part of the test data. Also, MetaPortrait was excluded from both user studies because of late data provision by authors, offering only the basic model results, without their super-resolution module outputs. 
}
    \label{tab:comparison_image}
    \vspace{-0.3cm}
\end{table}

\begin{table}[t]
    \centering
    \resizebox{0.97\linewidth}{!}{
    \begin{tabular}{l ccccc}
        \\
        Method 
        &
        FID$\downarrow$
        &
        CSIM$\uparrow$
        &
        UMTN$\downarrow$
        &
        UAPP$\downarrow$
        &
        $\mathbf{AUC}_{\mathbf{z}}$$\downarrow$
        \\
        \hline
        MegaPortraits fine-tuned   & 62.3 & 0.63 & 8.5 & 12.4 & 0.89  \\
        \hline
        Ours w/o $\mathcal{L}_{CV}^n$        & \underline{61.4} & \underline{0.67} & 9.2 & \underline{17.9} & 0.88  \\
        Ours w/o $\mathcal{L}_{sdm}$       & 61.8 & 0.59 & 11.3 & 6.7 & 0.85 \\
        Ours $\textit{dim}(\z) = 512$       & 63.4 & 0.44 & \underline{25.6} & 1.1 & \textbf{0.77} \\
        \hline
        Ours             & \textbf{59.6} & \textbf{0.74} & \textbf{45.4} & \textbf{61.9} & \underline{0.80}  \\ 
    \end{tabular}
    }
    \caption{Our ablation study examines key aspects of the image-driven mode. After fine-tuning MegaPortraits on the FEED dataset, we observed a decline in identity preservation due to overfitting on FEED identities, without any improvement in emotion translation. The lack of $\mathcal{L}_{CV}^n$ primarily impacts the ability to translate extreme expression, as indicated by UMTN. Additionally, not using $\mathcal{L}_{sdm}$ or setting $\textit{dim}(\z) = 512$ results in identity leakage, particularly evident in the latter scenario. For this ablation, we use the part of our driven by MEAD dataset and described in \cref{ssec:image_comp}}
    \label{tab:ablation_img_main}
\end{table}

\begin{table}[H]
    \centering
    \resizebox{\linewidth}{!}{
    \begin{tabular}{l ccccccc}
        \\
        Method 
        &
        FID$\downarrow$
        &
        $F_P$$\downarrow$
        &
        $F_V$$\downarrow$
        &
        $M_P$$\downarrow$
        &
        $M_V$$\downarrow$
        &
        Sync. offset $\downarrow$
        &
        Sync. conf. $\uparrow$
        \\
        \hline
        StyleHEAT \cite{yin2022styleheat}       & 62.4 & 1.34 & 0.41 & 2.98 & 1.67 & 0.41 & 7.22 \\
        PC-AVS \cite{zhou2021posecontrollable}  & 71.5 & 2.78 & 0.51 & 2.56 & \textbf{1.04} & 3.79 & \textbf{8.42} \\
        SadTalker \cite{zhang2023sadtalker}     & \underline{55.6} & \underline{0.91} & \underline{0.35} & \textbf{2.12} & \underline{1.17} & \underline{0.27} & \underline{7.58} \\
        MakeIrTalk \cite{Zhou_2020}             & 63.3 & 2.15 & 0.38 & 3.11 & 1.59 & 2.98 & 5.17 \\
        \hline
        Ours (EMOPortraits)  & \textbf{28.5} & \textbf{0.82} & \textbf{0.33} & \underline{2.53} & 1.38 & \textbf{0.07} & 5.77 \\
    \end{tabular}
    }
    \caption{Our model matches other leading speech-driven models, excelling in realism (FID), facial geometry and velocity, but is slightly less accurate in mouth landmark positions and velocities. We also score best in audio-video sync offset, but lower in confidence. We believe that while SyncNet metrics are still helpful, they may not fully represent true synchronization quality, as generated videos can sometimes score higher than ground truth without necessarily being better synchronized. In our experimentation, ground truth videos scored an offset of 1.13 (which is worse than our method, SadTalker, and StyleHEAT) and a confidence score of 8.35 (less than PC-AVS). For qualitative analysis, please refer to our supplementary materials.}
    \label{tab:comparison_audio}
\end{table}

\begin{table}[H]
    \centering
    \resizebox{\linewidth}{!}{
    \begin{tabular}{l ccccc}
        \\
        Method 
        &
        FID$\downarrow$
        &
        $M_P$$\downarrow$
        &
        $M_V$$\downarrow$
        &
        $F_P$$\downarrow$
        &
        $F_V$$\downarrow$

        \\
        \hline
        Ours w/o disentanglement        & 44.7 & 0.98 & 0.45 & 5.78 & 2.55  \\
        Ours w/o $\mathcal{L}_\text{PCA}$       & 29.3 & 0.84 & 0.35 & 3.13 & 1.88\\
        Ours with $\text{MAE}(\z_i, \hat\z_i^\text{speech})$       & \textbf{28.2} & \underline{0.84} & \underline{0.34} &\underline{2.61} & \underline{1.45}\\
        \hline
        Ours             & \underline{28.5} & \textbf{0.82} & \textbf{0.33} & \textbf{2.53} & \textbf{1.38} \\
    \end{tabular}
    }
    \caption{Ablation study of our speech-driven mode, where in the first line, we show that it notably underperforms without head pose - face expression disentanglement of the latent space (see \cref{ssec:full_dis}). Removing  $\mathcal{L}_\text{PCA}$ leads to a significant reduction in generating realistic mouth movements. Replacing $\mathcal{L}_\text{PCA}$ with $\text{MAE}_(\z_i, \hat\z_i^\text{speech})$ causes a noticeable, decline in mouth movement generation. This suggests that identifying mouth-related PCA components was beneficial in enhancing the outcome.}
    \label{tab:ablation_audio_main}
\end{table}

\subsection{Ablation study}
\label{ssec:ablation}

We conducted an extensive ablation study to evaluate the contributions of individual components within our method. For our main model, we present the evaluation of the importance of the proposed source-driver unmatch loss, canonical volume loss, and change of $\textit{dim}(\z)$ from 512 in original model to 128 \cref{tab:ablation_img_main}. For our speech-driven mode, we show how vital the disentanglement of the latent space for speech-drive ability, how $\mathcal{L}_\text{PCA}$ is better than naive matching of PCA components from random $\mathcal{Z}$ and how it helps match speech and latent expression space \cref{tab:ablation_audio_main}. For more ablation analysis, plots and visual comparison, please refer to our supplementary.

\section{Conclusion}
\label{sec:conclusion}


In this paper, we introduce EmoPortraits, a novel method for creating neural avatars with superior performance in image-driven, cross-identity emotion translation. Our speech-driven mode makes it possible to drive the facial animation through multiple conditions (video, audio, head motion). We collected FEED dataset which, we believe, will be a valuable asset for researchers in diverse human-centered studies.

However, our method has some limitations. It doesn't generate the avatar's body or shoulders, limiting some use cases. We currently integrate our output with a source image body. Additionally, the model sometimes struggles with accurate expression translation and underperforms with extensive head rotation. These challenges are crucial for future enhancements and remain central to our ongoing research efforts.



{\small
\bibliographystyle{ieeenat_fullname}
\bibliography{_main}
}

\clearpage

\section{Main model implementation details}

\subsection{Full List of Findings}
Beyond the core findings highlighted in the main text, this section outlines additional key distinctions that set our work apart from the MegaPortraits model \cite{drobyshev2023megaportraits}.

\begin{figure}[t]
    \centering
    \includegraphics[width=\linewidth]{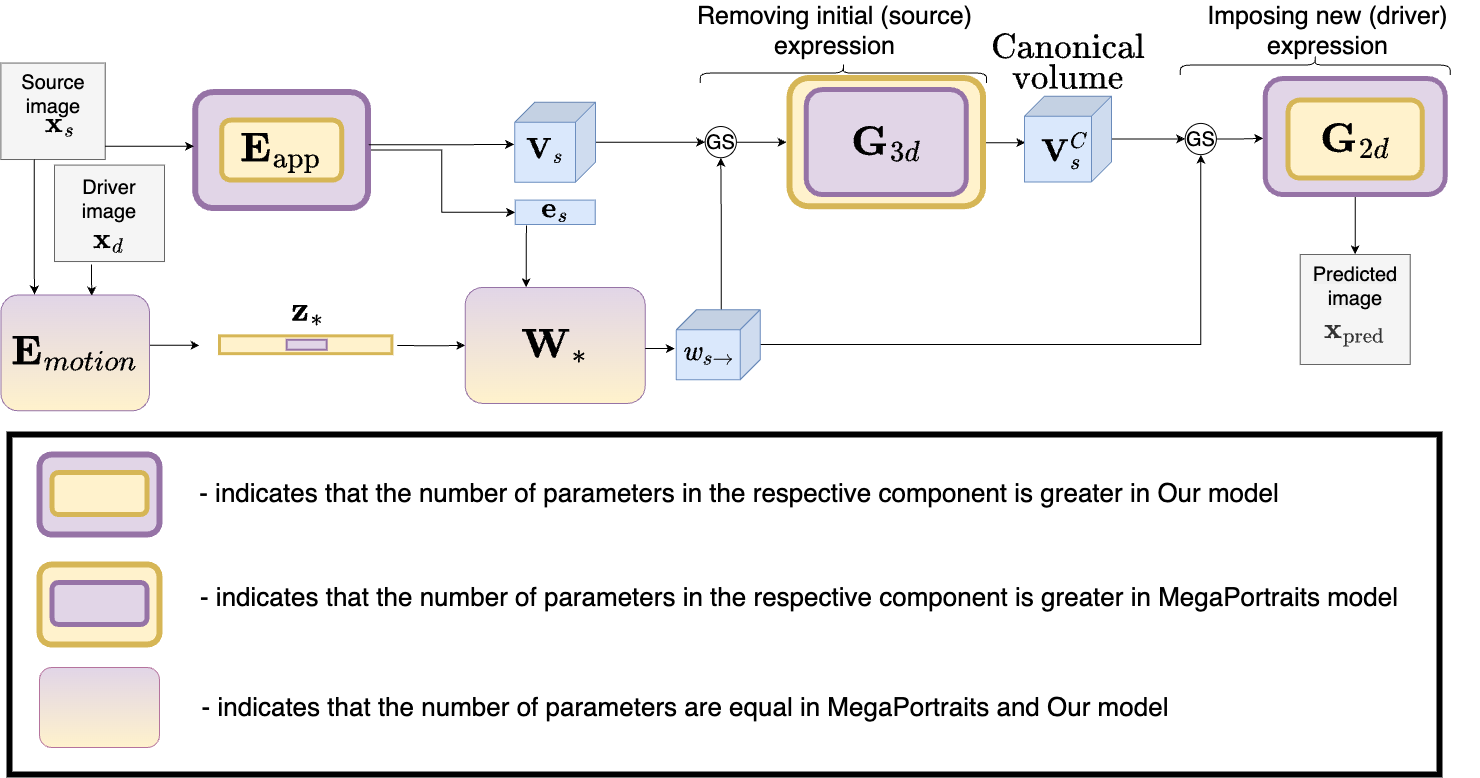}
    \caption{Comparison of our model's scheme with the MegaPortraits scheme, showing the relative sizes of individual components.}
    \label{fig:mp_vs_ours}
\end{figure}

\textit{Model Architecture.} Our main model's structure, depicted in Figure~\ref{fig:main_scheme} of the main text, excluding the speech-driven component, shares a conceptual resemblance with that of MegaPortraits~\cite{drobyshev2023megaportraits}. However, we have implemented several architecture alterations. First is the reduction in the dimensionality of the latent expression descriptors from 512 to 128, a change detailed in Section~\ref{ssec:lat_expr_spc}. This reduction enhances the efficiency of the model without compromising the quality of expression representation. Additionally, we have made comprehensive modifications to the architecture and size of the model's main components to optimize model's performance. These modifications are visually represented in \cref{fig:mp_vs_ours}.

\textit{Dropout.} We have incorporated dropout as a last layer of $\E_{\text{motion}}$ that predicts the expression descriptors (\(\z\)) in our model. This implementation serves to improve \(\E_{motion}\)'s capability to construct a more nuanced and robust latent representation of facial expressions, and also aids in preventing overfitting. By ensuring that the model does not become overly reliant on any element of the latent vector, we achieve a more generalized and versatile expression representation capability, essential for dealing with a wide range of facial motions.

\textit{Enhanced Loss Functions.} Beyond new loss functions introduced in the main text, such as the canonical volume loss (outlined in Section~\ref{ssec:cv_method}) and the source-driver mismatch loss (described in Section~\ref{ssec:cd_loss}), we have also integrated the \(\mathcal{L}_\text{head}\) loss, as mentioned in Section~\ref{ssec:losses}. This specific loss function plays a noticeable role in the precise predicting of facial regions critical for emotional expression, particularly the eyes and mouth. Additionally, it addresses the previously noted challenges in accurately generating ears. The integration of this loss underscores our model's attention to detail and commitment to achieving a high degree of realism in facial expression synthesis.

\subsection{Implementation Details}

\textbf{Data Preparation.} Our data preparation approach for the VoxCeleb2 (VC2) dataset~\cite{Chung2018VoxCeleb2DS} follows the protocol established in the original model~\cite{drobyshev2023megaportraits}. For our novel FEED dataset, we cropped frames around the face region and resized them. The dataset subset used in our experiments, detailed in Table~\ref{tab:feed_tasks}, includes "Winks," "Tongue Emotion," and "Extreme \& Asymmetric Emotion." Notably, during training, we did not exploit the multi-view nature of the FEED dataset. Specifically, in each iteration, both the source and driver images were selected from the same single-camera video. Both datasets were employed for training and evaluating our model at a resolution of \(512 \times 512\).

\textbf{Training Details.} Our framework was trained on 8 Nvidia Tesla A100 GPUs for 250,000 iterations, using a batch size of 2 per GPU (16 in total). The training data consisted of a mixture of 75\% VC2 examples and 25\% FEED examples. Every second iteration involved a batch comprising one image pair from VC2 and one from the FEED dataset. This sampling strategy, integrating image pairs from both datasets, proved more effective than using separate batches from each dataset. By employing contrastive losses where positive and negative pairs spanned both datasets, we mitigated overfitting risks associated with the limited identity variety in the FEED dataset. This approach also facilitated the asymmetric emotion translation to unseen identities.

\subsection{Used losses}
\label{ssec:losses}

\textbf{Photometrical losses}. These are key to aligning the motion and appearance of the predicted image (\(\hat\x_{s \rightarrow d}\)) with the ground truth image (\(\x_d\)). To achieve this, we use three distinct pre-trained networks:

\begin{itemize}
\item \textit{VGG19} ~\cite{Simonyan2015VeryDC}  (ILSVRC/ImageNet \cite{Deng2009ImageNetAL} Trained): This helps in matching the overall content of the images.
\item \textit{VGGFace} \cite{Parkhi2015DeepFR} (Face Recognition Focused): Essential for aligning facial features accurately.
\item \textit{Gaze Direction Based on VGG16} \cite{Fischer2018RTGENERE}: Specifically trained to emulate a top-notch gaze detection system, ensuring precise gaze direction matching.
\end{itemize}

We measure the similarity by calculating the L1 distance between the feature maps of both the predicted and ground truth images, utilizing all these networks. Additionally, we employ face masks for the eyes, mouth, and ears (sourced from FaceParsing network \cite{yu2018bisenet}), focusing our model on these critical head areas. Then we use these masks to match mentioned head regions on the predicted and the ground truth images using L1 loss between pixel corresponded to a specific region. The final photometric loss combines these individual perceptual losses, formulated as:

\begin{equation}
\mathcal{L}_\text{pho} = w_\text{IN} \mathcal{L}_\text{IN} + w_\text{face} \mathcal{L}_\text{face} + w_\text{gaze} \mathcal{L}_\text{gaze} + \mathcal{L}_\text{head}.
\end{equation}

Here, \(\mathcal{L}_\text{head}\) further breaks down into:

\begin{equation}
\mathcal{L}_\text{head} = w_\text{eyes} \mathcal{L}_\text{eyes} + w_\text{mouth} \mathcal{L}_\text{mouth} + w_\text{ears} \mathcal{L}_\text{ears}.
\end{equation}

\textbf{Self-supervised losses}. As detailed in Section~\ref{sec:method_audio}, we trained our expression descriptors using a modified large margin cosine loss (CosFace)~\cite{Wang2018CosFaceLM}, denoted as \(\mathcal{L}_\text{cos}\) and presented in Equation~\ref{eq:cosloss}. This approach is similar to the one employed by the authors of MegaPortraits~\cite{drobyshev2023megaportraits}. Additionally, we introduced two more losses. The source-driver mismatch loss \(\mathcal{L}_\text{sdm}\)  Equation~\ref{eq:sdm} (described in Section~\ref{ssec:cd_loss}), which directly influences the expression's latent space. This loss is pivotal in eliminating identity information from the expression descriptor \(\z_{i}\) and in preventing overfitting, especially in the context of our extremely imbalanced dataset. The combination of these two losses forms our latent space loss:

\begin{equation}
\mathcal{L}_\text{lat} = w_\text{cos} \mathcal{L}_\text{cos} + w_\text{sdm} \mathcal{L}_\text{sdm}.
\end{equation}

The second additional self-supervised loss that enhances the disentanglement of identity and expression is the canonical volume loss \(\mathcal{L}_\text{CV}\) (described in Section~\ref{ssec:cv_method} and Equation~\ref{eq:cv_loss}). This loss functions to extract expression information from the canonical volume, thereby reducing the overlap of information contained in \(\textbf{V}^C_i\) and \(\z_{i}\).

\textbf{Adversarial losses}. To ensure the predicted images look realistic, adversarial losses are computed using the same predicted and reference images. We follow \cite{drobyshev2023megaportraits} and train a multi-scale patch discriminator  \cite{Zhu2017UnpairedIT} with a hinge adversarial loss. To boost training stability, a standard feature-matching loss is also included (\cite{wang2018pix2pixHD}). The GAN loss for the generator is expressed as:

\begin{equation}
\mathcal{L}_\text{GAN} = w_\text{adv} \mathcal{L}_\text{adv} + w_\text{FM} \mathcal{L}_\text{FM}.
\end{equation}

To conclude, the total loss which is used to train our model is the sum of individual losses:
\begin{equation}
    \mathcal{L}_\text{main} = \mathcal{L}_\text{pho} + \mathcal{L}_\text{lat}  + w_\text{CV} \mathcal{L}_\text{CV} + \mathcal{L}_\text{GAN}.
\end{equation}

We utilized the AdamW optimizer~\cite{Loshchilov2019DecoupledWD} with a cosine learning rate schedule. The initial learning rate was gradually reduced from \(2 \times 10^{-4}\) to \(1 \times 10^{-6}\) over the training iterations. The hyperparameters for the losses were set as follows: \(w_\text{IN} = 20\), \(w_\text{face} = 10\), \(w_\text{gaze} = 10\), \(w_\text{adv} = 1\), \(w_\text{FM} = 40\), \(w_\text{cos} = 2\), \(w_\text{std} = 1 \) (increased to 10 for pairs from the FEED dataset), and \(w_\text{CV} = 1\). Additionally, we set \(s = 5\) and \(m = 0.2\) in the cosine loss.






\begin{figure}[tp]
    \centering
    \includegraphics[width=\linewidth]{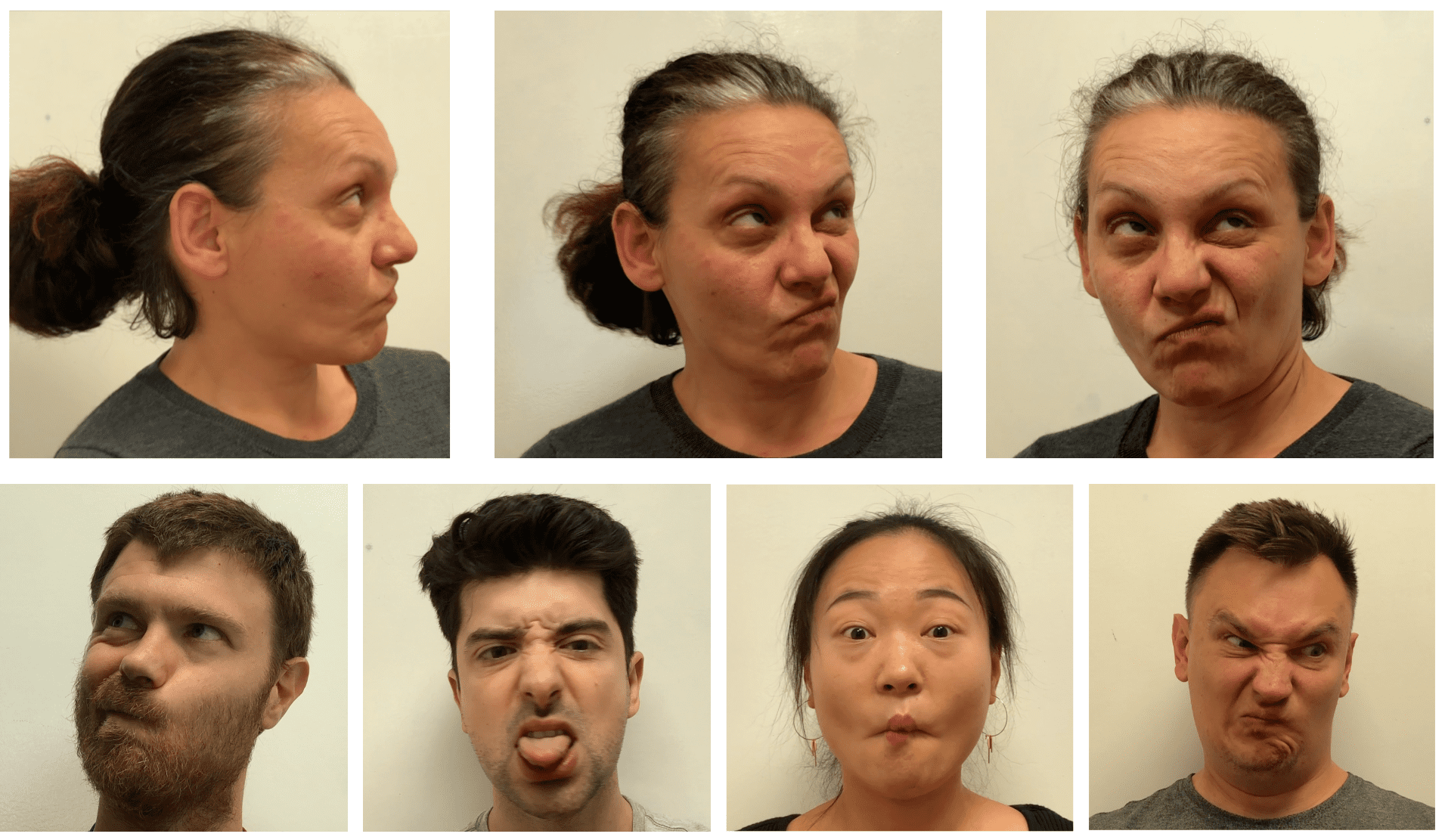}
    \caption{More selected examples from our FEED dataset.}
    \label{fig:more_feed}
\end{figure}

\subsection{Visual Comparison}
Our choice of baseline methods, as outlined in our experiment section in the main text, was driven by two key factors. Firstly, these methods are prominent in the field of talking-head video generation using arbitrary identities, making them relevant benchmarks for our study. Secondly, the accessibility of their source code and pretrained model weights, either through public availability or provided by the authors for use with our test set, was a crucial consideration.

The setup for our visual comparison, as described in Section 6.1, was chosen based on specific criteria. We choose FFHQ images as source images due to the consistent clarity of facial features across the dataset, which is essential for accurate comparison. Additionally, it was critical to select identities that were not part of the training datasets for any of the methods used in comparison. This ensures that our comparisons are based on novel identities, providing a fair assessment of each method's generalization capabilities. This criterion was also applied in selecting driving identities from the MEAD and FEED datasets, which were not used in training by any of the compared methods.

\begin{figure*}[]
    \centering    
    \setlength{\wid}{0.18\textwidth}
    \setlength{\mrg}{-0.3cm}
    \resizebox{\linewidth}{!}{
    \begin{tabular}{cccccccccc}
        \includegraphics[width=\wid]{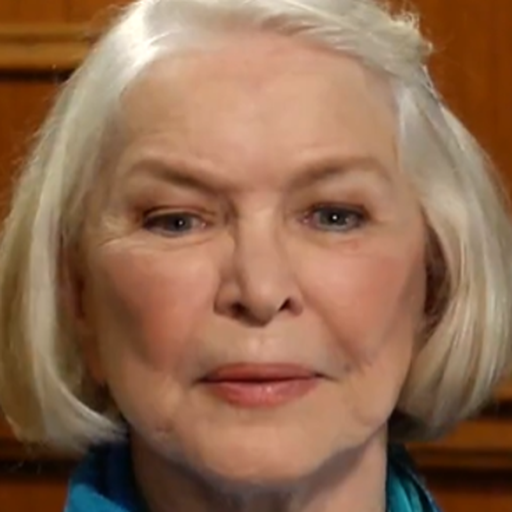} & 
        \hspace{\mrg}
        \includegraphics[width=\wid]{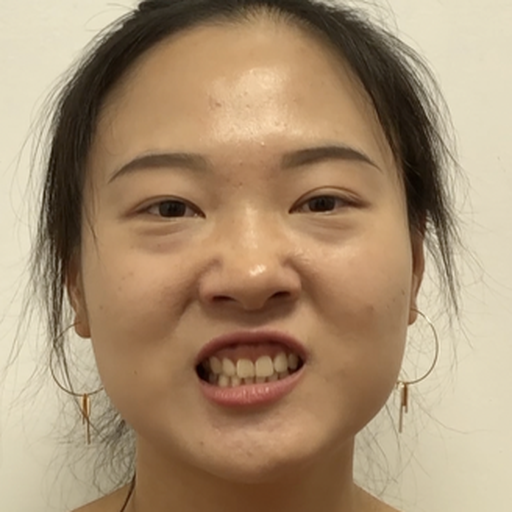} & 
        \hspace{\mrg}
        \includegraphics[width=\wid]{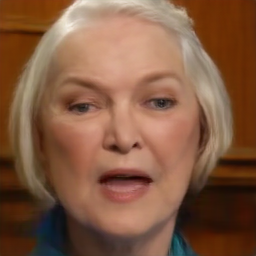} & 
        \hspace{\mrg}
        \includegraphics[width=\wid]{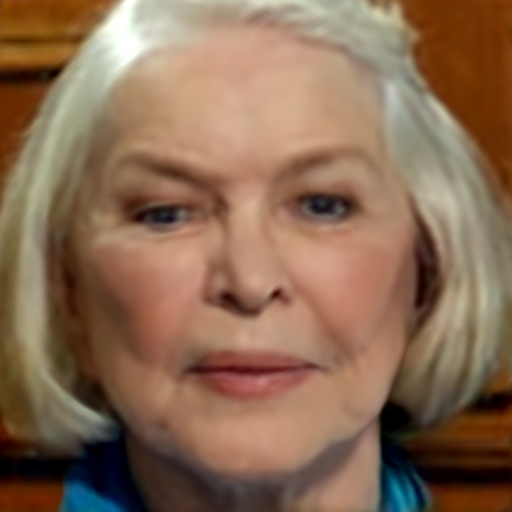} & 
        \hspace{\mrg}
        \includegraphics[width=\wid]{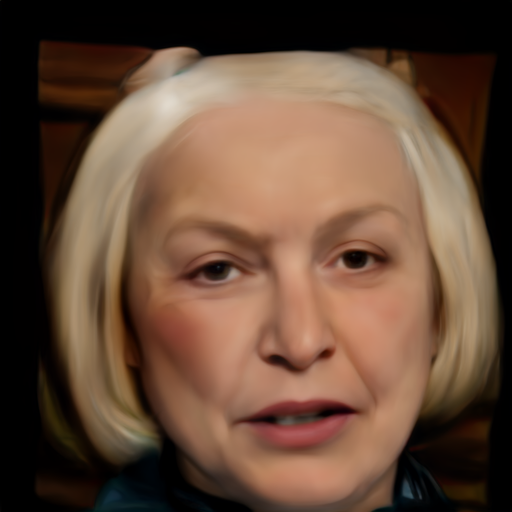} & 
        \hspace{\mrg}
        \includegraphics[width=\wid]{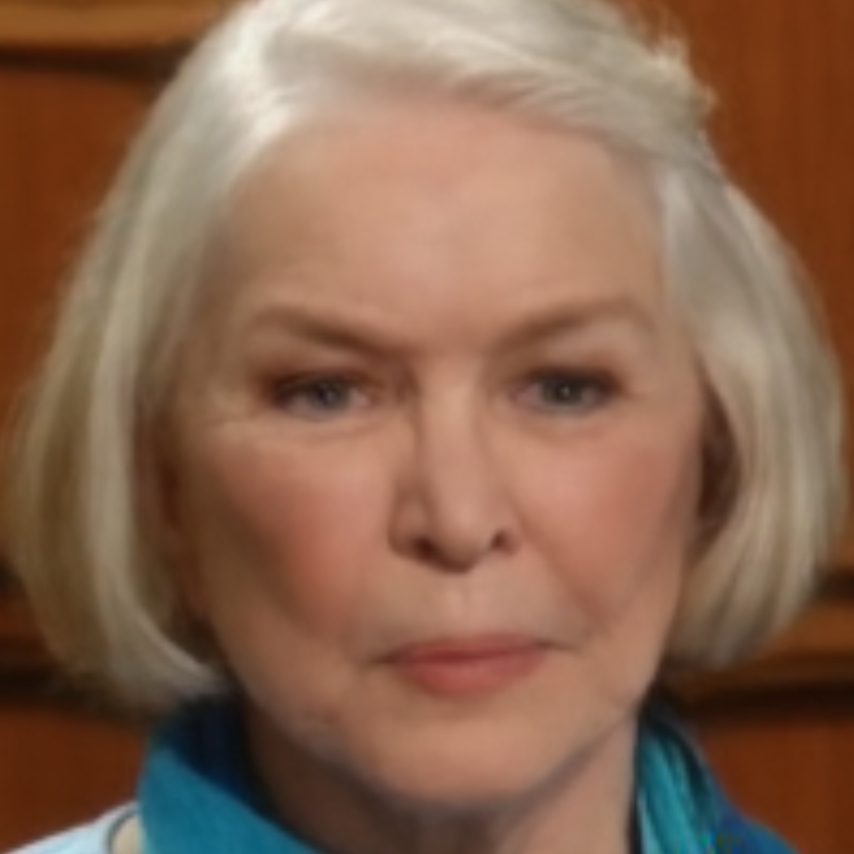} & 
        \hspace{\mrg}
        \includegraphics[width=\wid]{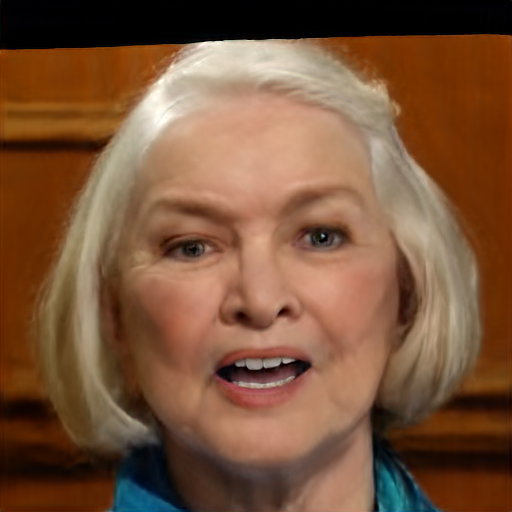} & 
        \hspace{\mrg}
        \includegraphics[width=\wid]{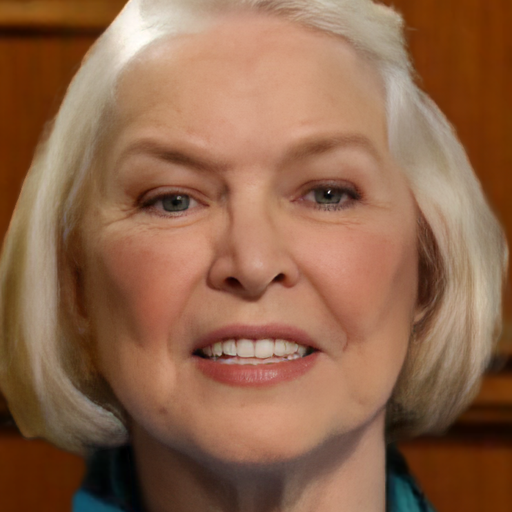} &
        \hspace{\mrg}
        \includegraphics[width=\wid]{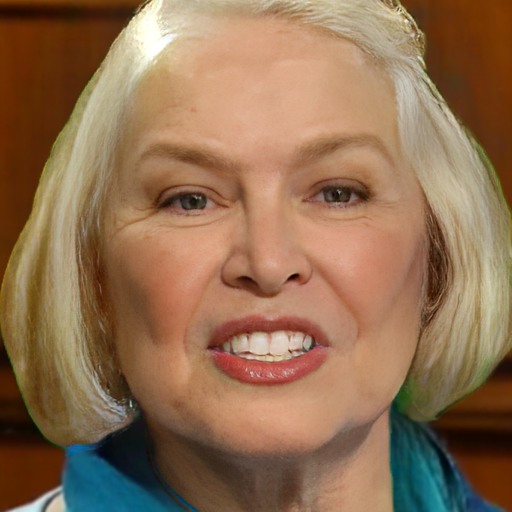} & 
        \\ %
        \includegraphics[width=\wid]{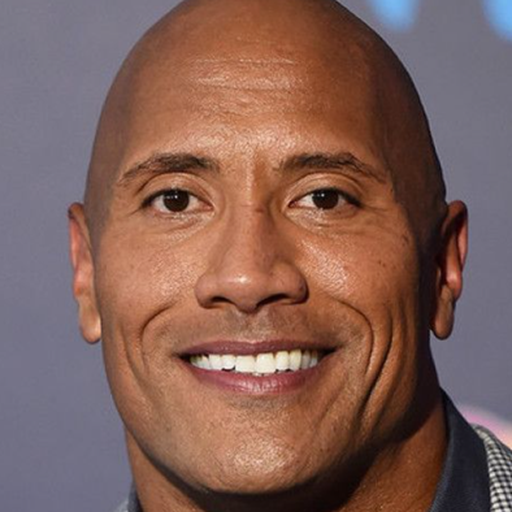} & 
        \hspace{\mrg}
        \includegraphics[width=\wid]{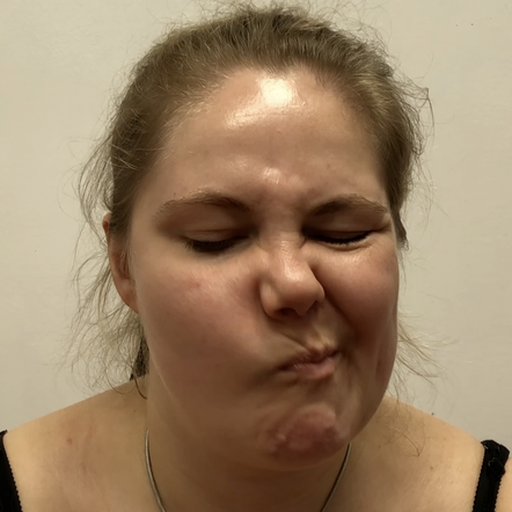} & 
        \hspace{\mrg}
        \includegraphics[width=\wid]{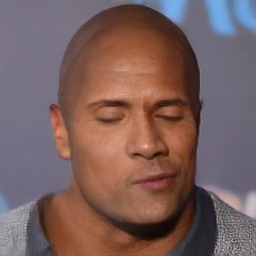} & 
        \hspace{\mrg}
        \includegraphics[width=\wid]{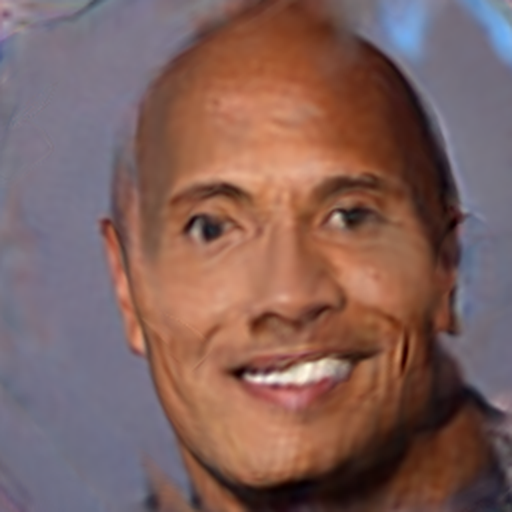} & 
        \hspace{\mrg}
        \includegraphics[width=\wid]{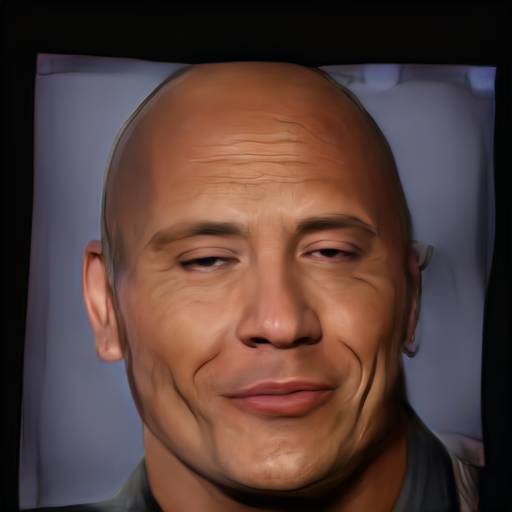} & 
        \hspace{\mrg}
        \includegraphics[width=\wid]{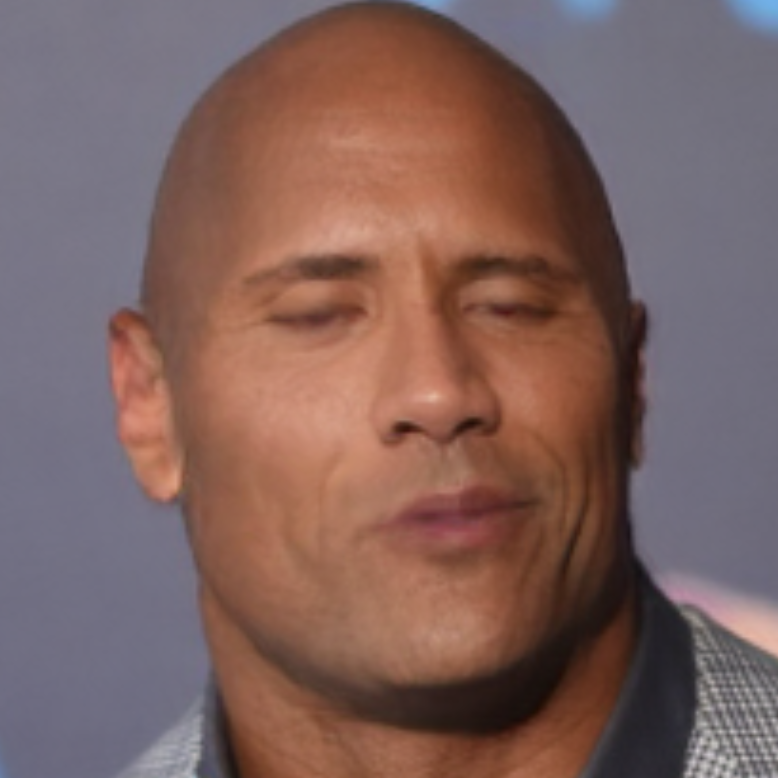} & 
        \hspace{\mrg}
        \includegraphics[width=\wid]{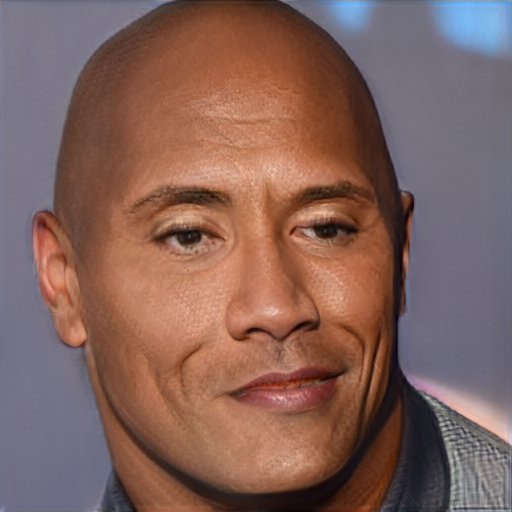} & 
        \hspace{\mrg}
        \includegraphics[width=\wid]{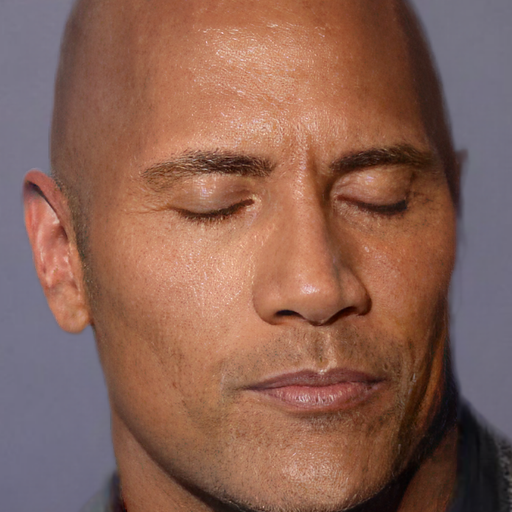} &
        \hspace{\mrg}
        \includegraphics[width=\wid]{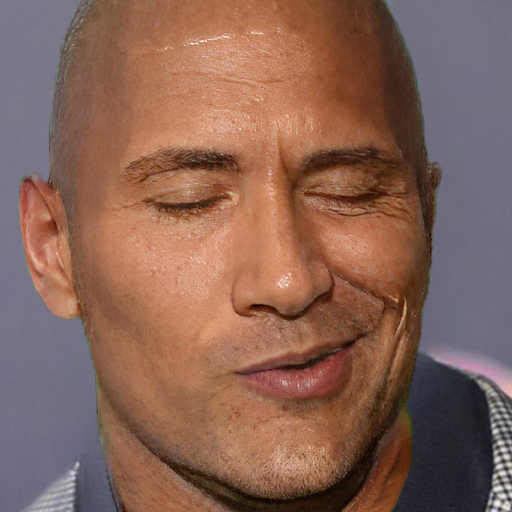} & 
        \\
        \includegraphics[width=\wid]{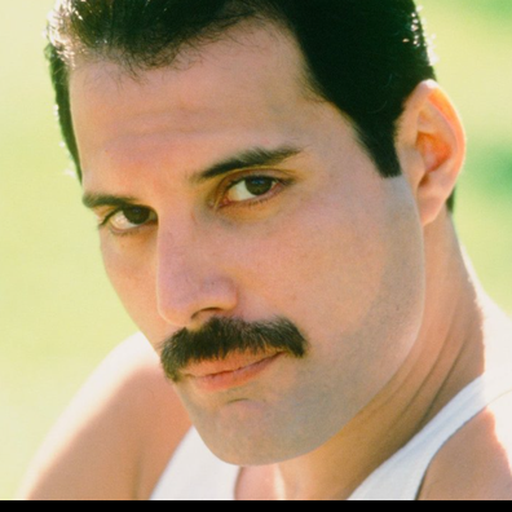} & 
        \hspace{\mrg}
        \includegraphics[width=\wid]{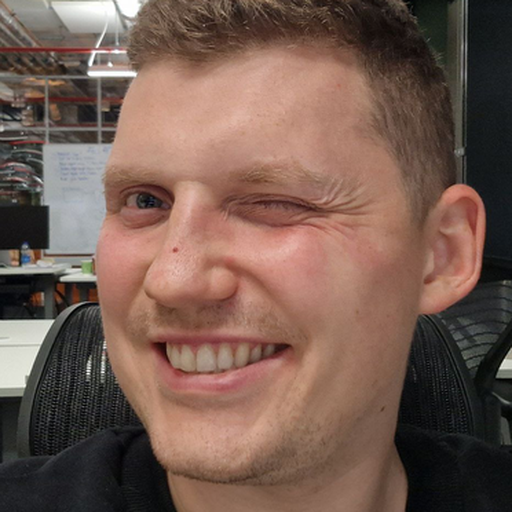} & 
        \hspace{\mrg}
        \includegraphics[width=\wid]{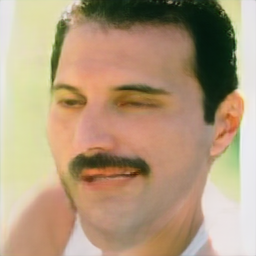} & 
        \hspace{\mrg}
        \includegraphics[width=\wid]{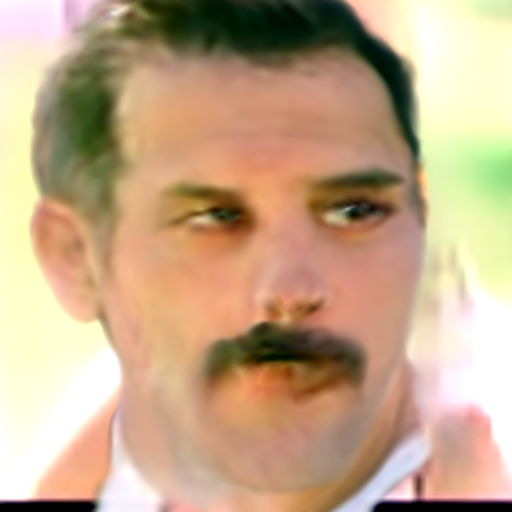} & 
        \hspace{\mrg}
        \includegraphics[width=\wid]{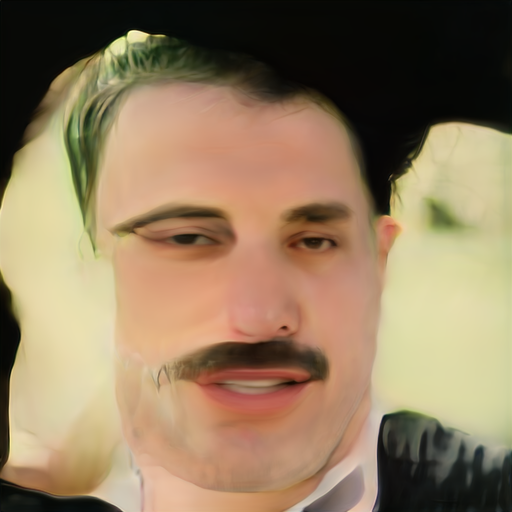} & 
        \hspace{\mrg}
        \includegraphics[width=\wid]{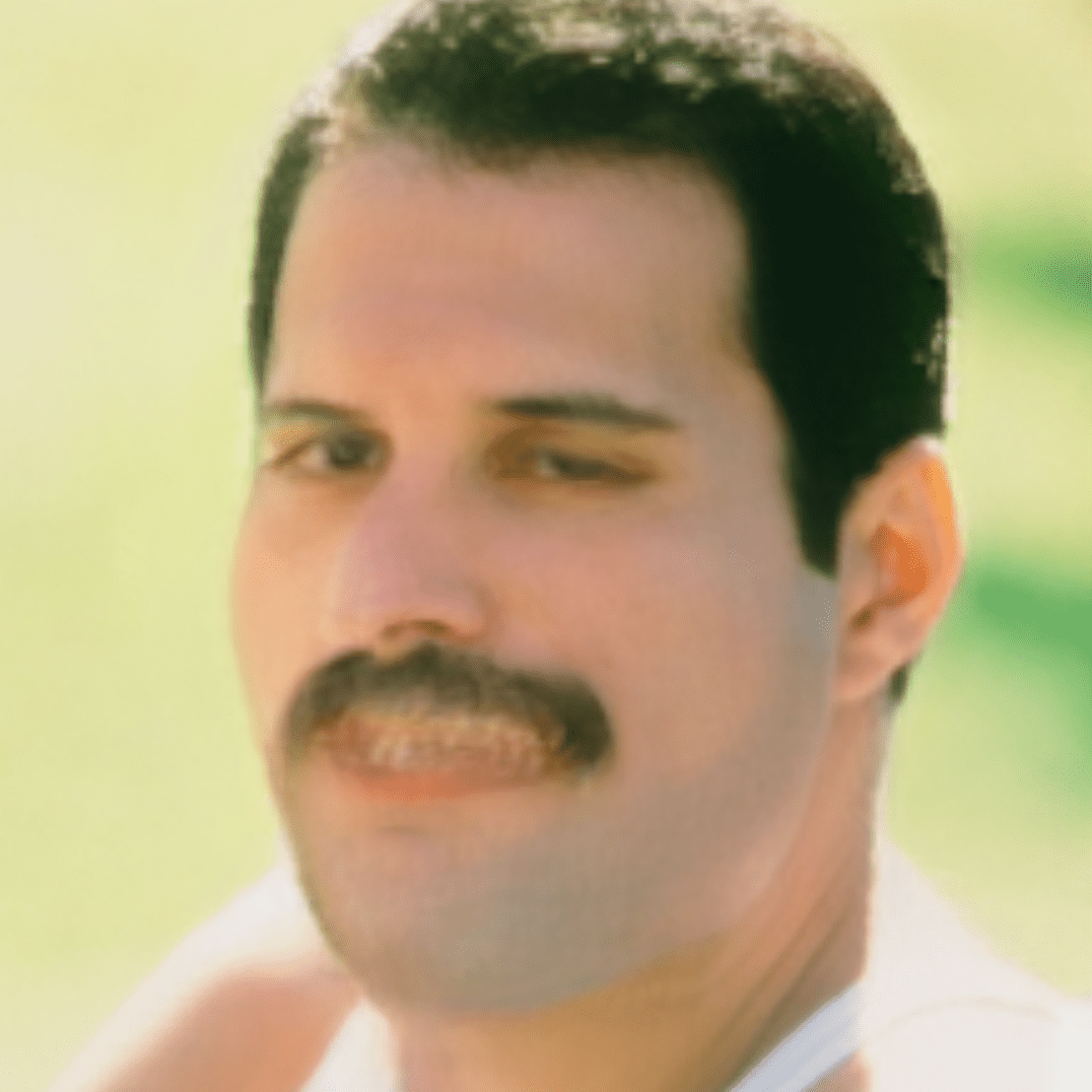} & 
        \hspace{\mrg}
        \includegraphics[width=\wid]{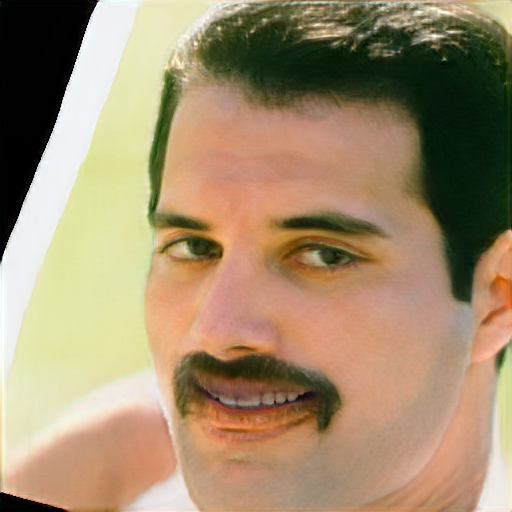} & 
        \hspace{\mrg}
        \includegraphics[width=\wid]{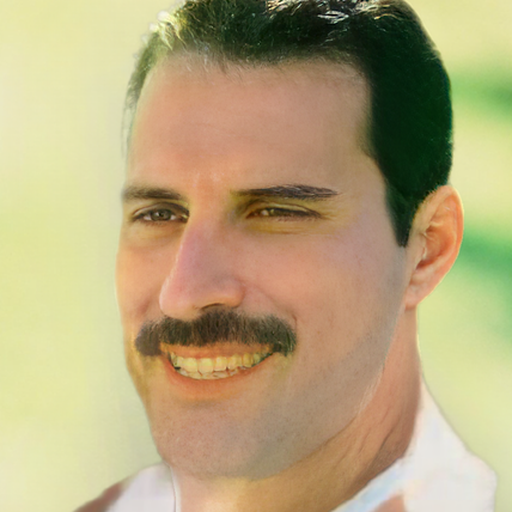} &
        \hspace{\mrg}
        \includegraphics[width=\wid]{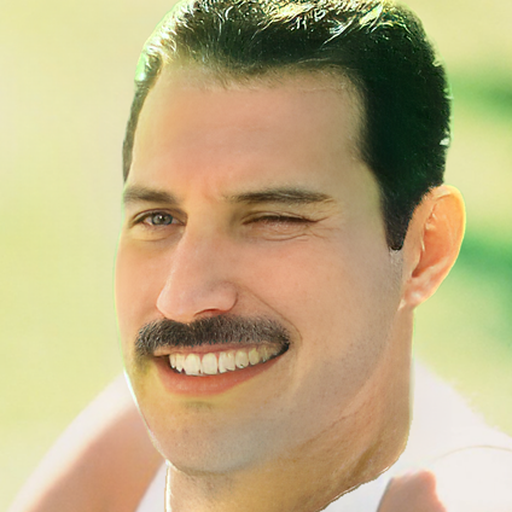} & 
        \\

        \textbf{Source} & 
        \hspace{\mrg} 
        \textbf{Driver} & 
        \hspace{\mrg} 
        \textbf{FOMM}~\cite{Siarohin2019FirstOM} & 
        \hspace{\mrg} 
        \textbf{UVA}~\cite{li2023generalizable} & 
        \hspace{\mrg} 
        \textbf{StyleHEAT}~\cite{yin2022styleheat} &
        \hspace{\mrg} 
        \textbf{MetaPortrait}~\cite{zhang2023metaportrait} &
        \hspace{\mrg} 
        \textbf{NOFA}~\cite{yu2023nofa} &
        \hspace{\mrg} \textbf{MegaPortraits}~\cite{drobyshev2023megaportraits} &
        \hspace{\mrg}
        \textbf{Ours}
    \end{tabular}
    }
    \vspace{-0.2cm}
    \caption{An additional qualitative comparison of head avatar systems in cross-reenactment scenario.}
    \label{fig:comparison_imgs_supp_1}
\end{figure*}

\begin{figure*}[]
    \centering    
    \setlength{\wid}{0.18\textwidth}
    \setlength{\mrg}{-0.3cm}
    \resizebox{\linewidth}{!}{
    \begin{tabular}{ccccccccc}
        \includegraphics[width=\wid]{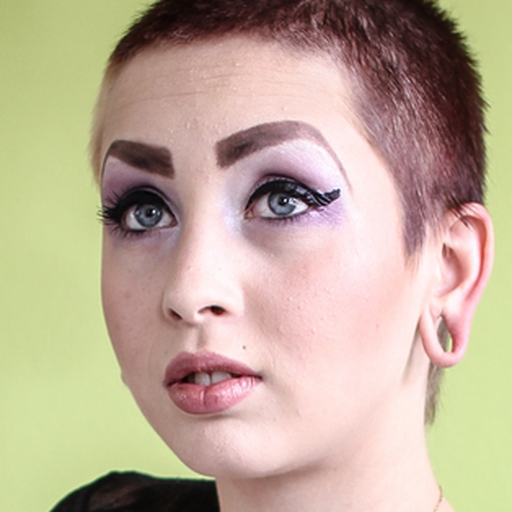} & 
        \hspace{\mrg}
        \includegraphics[width=\wid]{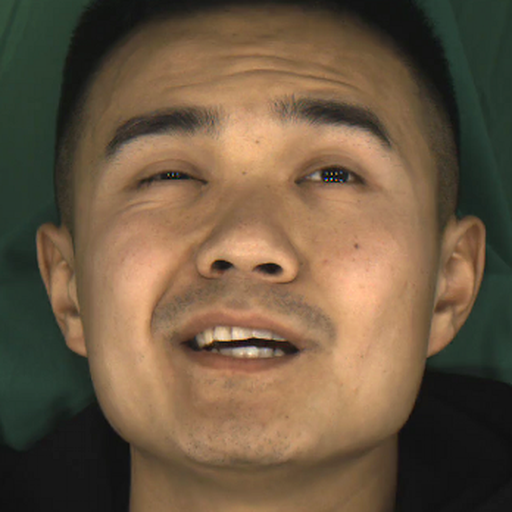} & 
        \hspace{\mrg}
        \includegraphics[width=\wid]{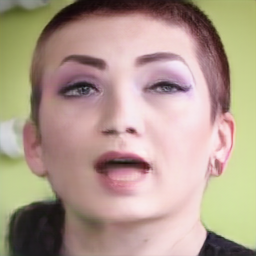} & 
        \hspace{\mrg}
        \includegraphics[width=\wid]{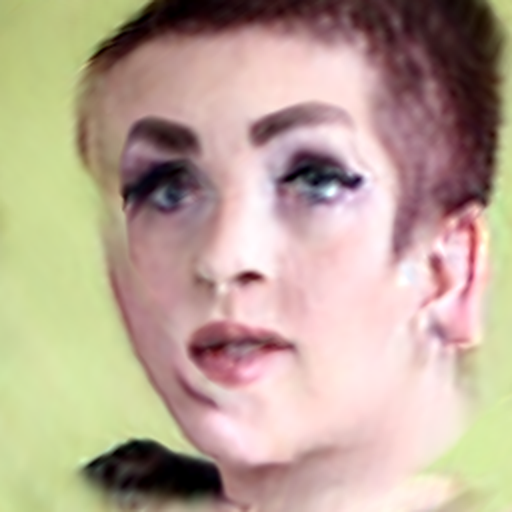} & 
        \hspace{\mrg}
        \includegraphics[width=\wid]{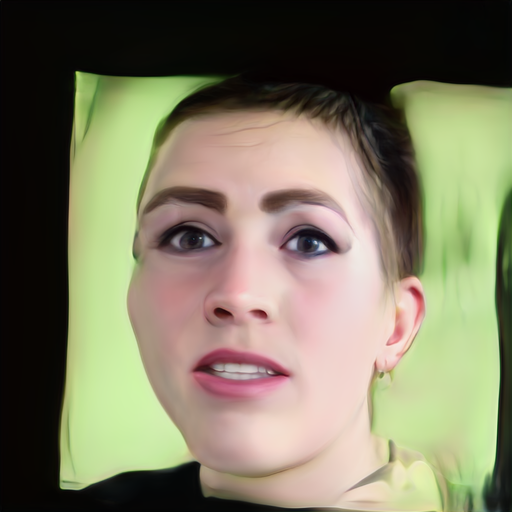} & 
        \hspace{\mrg}
        \includegraphics[width=\wid]{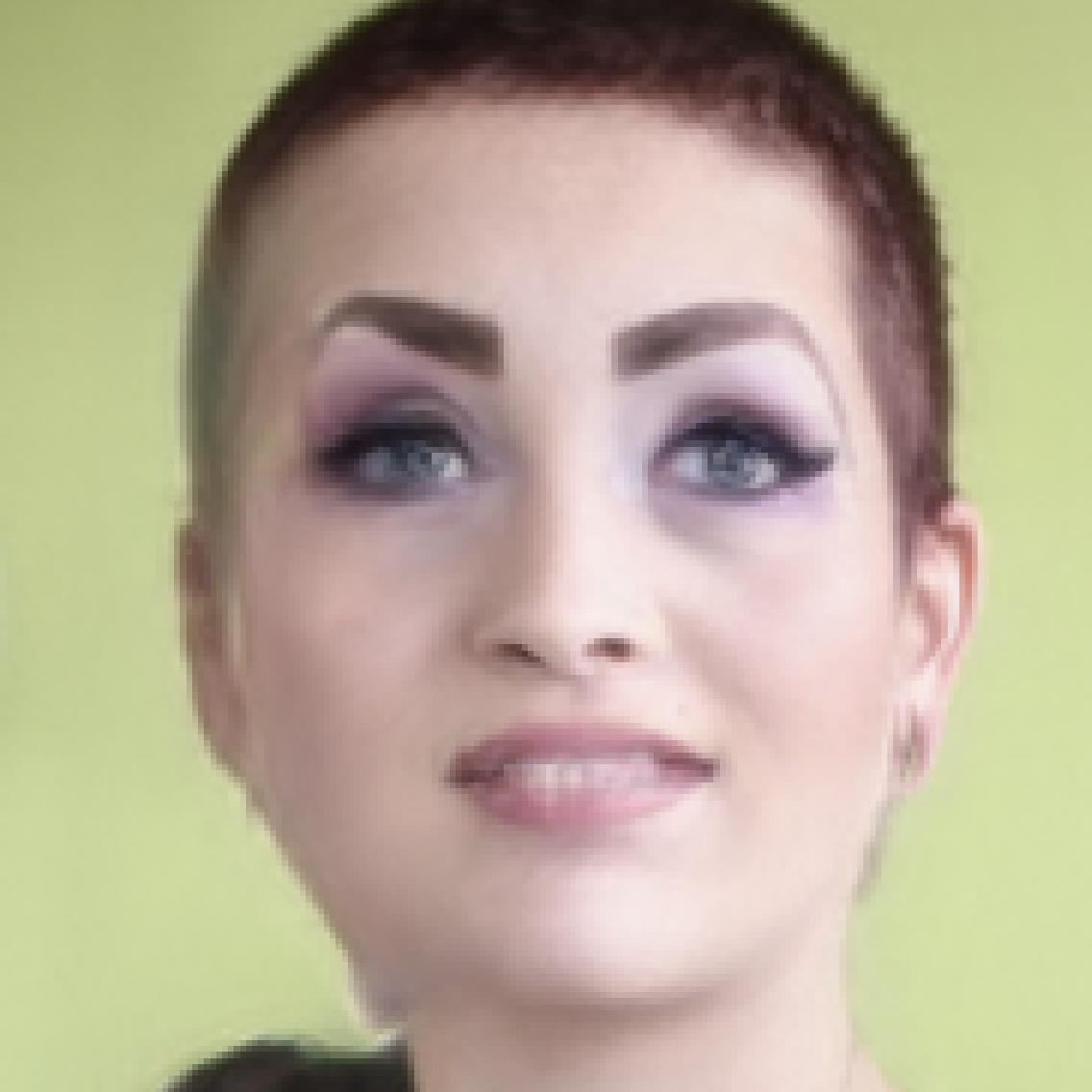} & 
        \hspace{\mrg}
        \includegraphics[width=\wid]{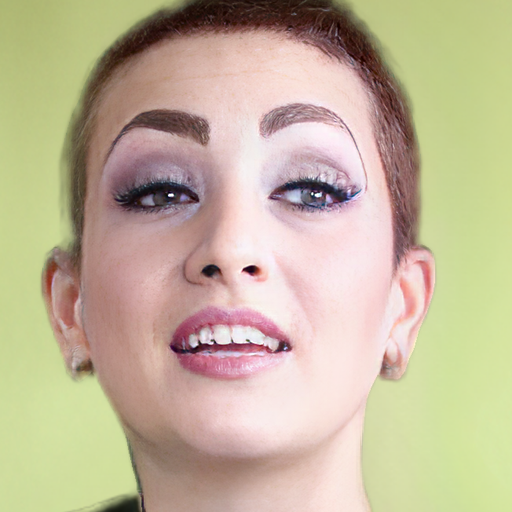} &
        \hspace{\mrg}
        \includegraphics[width=\wid]{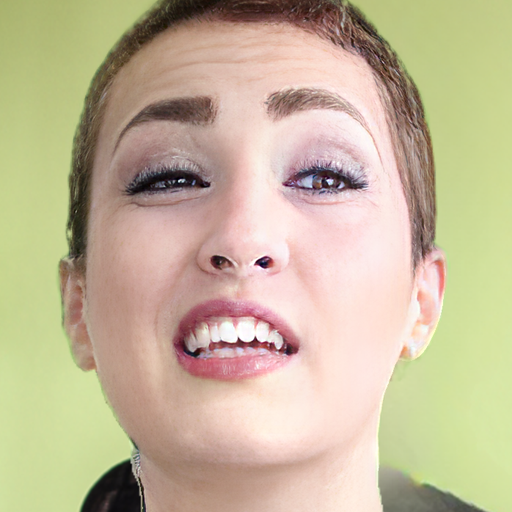} & 
        \\ %
        \includegraphics[width=\wid]{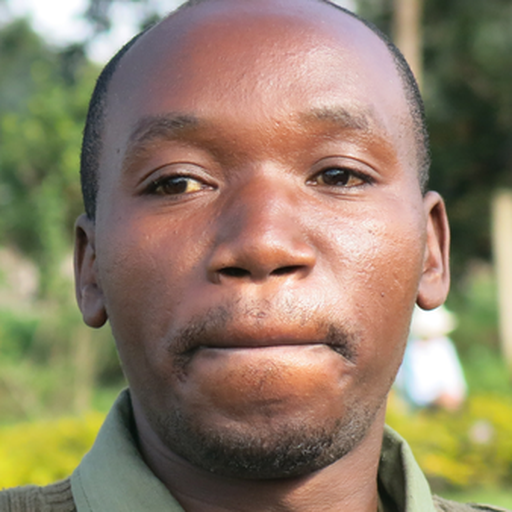} & 
        \hspace{\mrg}
        \includegraphics[width=\wid]{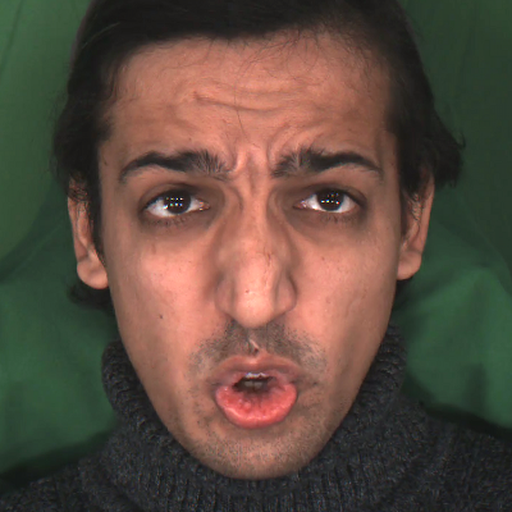} & 
        \hspace{\mrg}
        \includegraphics[width=\wid]{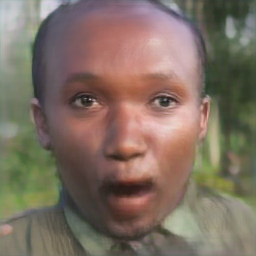} & 
        \hspace{\mrg}
        \includegraphics[width=\wid]{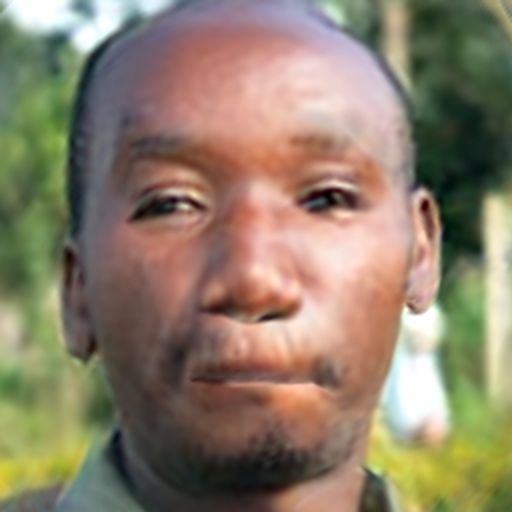} & 
        \hspace{\mrg}
        \includegraphics[width=\wid]{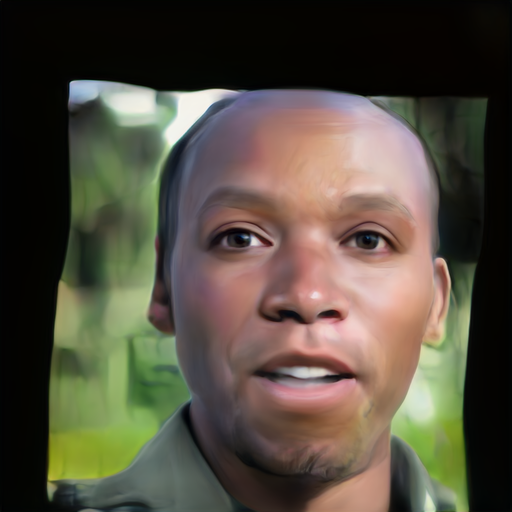} & 
        \hspace{\mrg}
        \includegraphics[width=\wid]{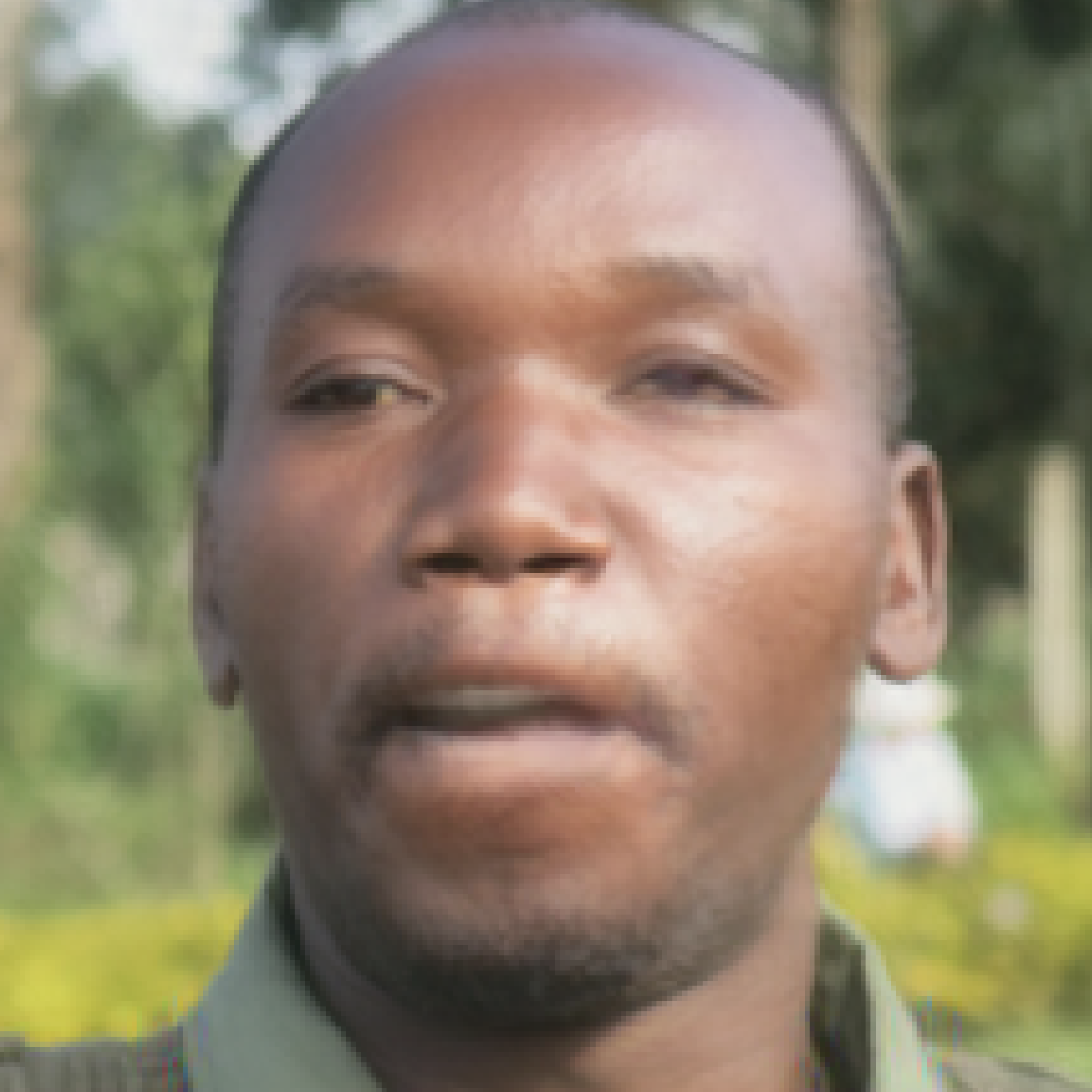} & 
        \hspace{\mrg}
        \includegraphics[width=\wid]{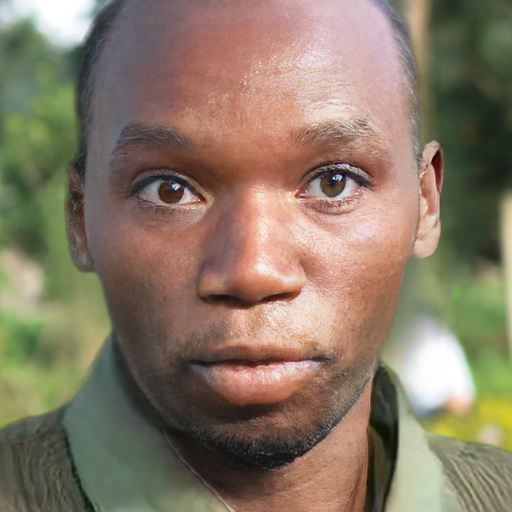} &
        \hspace{\mrg}
        \includegraphics[width=\wid]{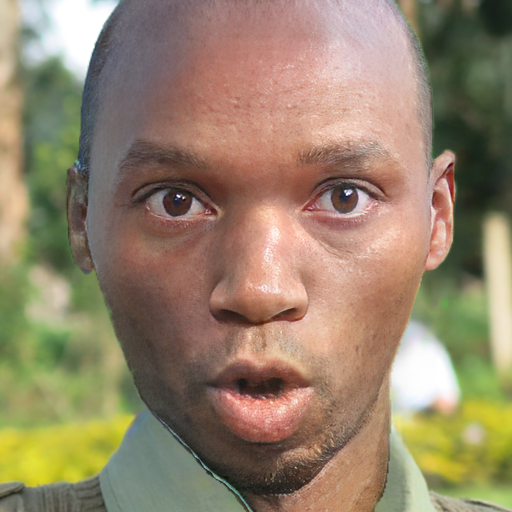} & 
        \\
        \includegraphics[width=\wid]{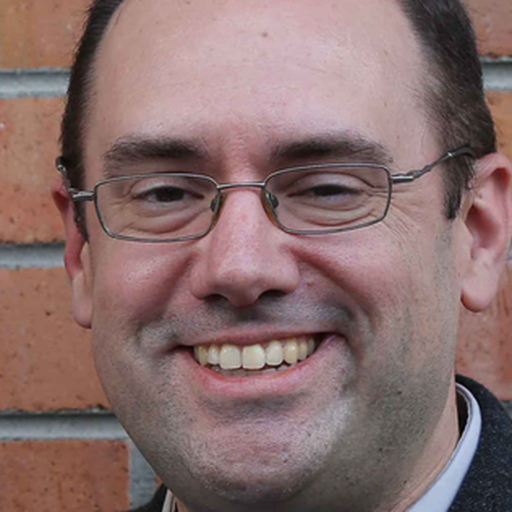} & 
        \hspace{\mrg}
        \includegraphics[width=\wid]{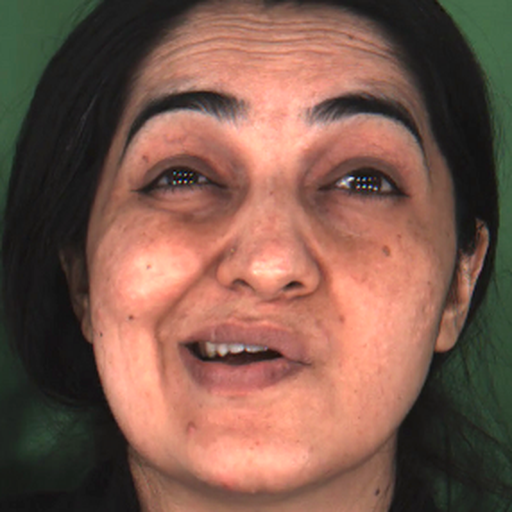} & 
        \hspace{\mrg}
        \includegraphics[width=\wid]{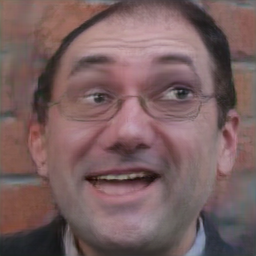} & 
        \hspace{\mrg}
        \includegraphics[width=\wid]{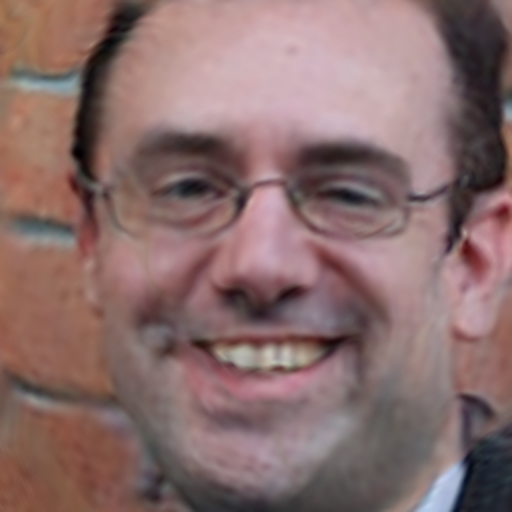} & 
        \hspace{\mrg}
        \includegraphics[width=\wid]{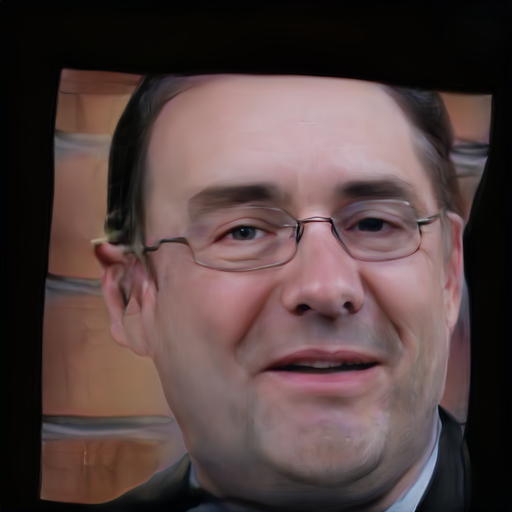} & 
        \hspace{\mrg}
        \includegraphics[width=\wid]{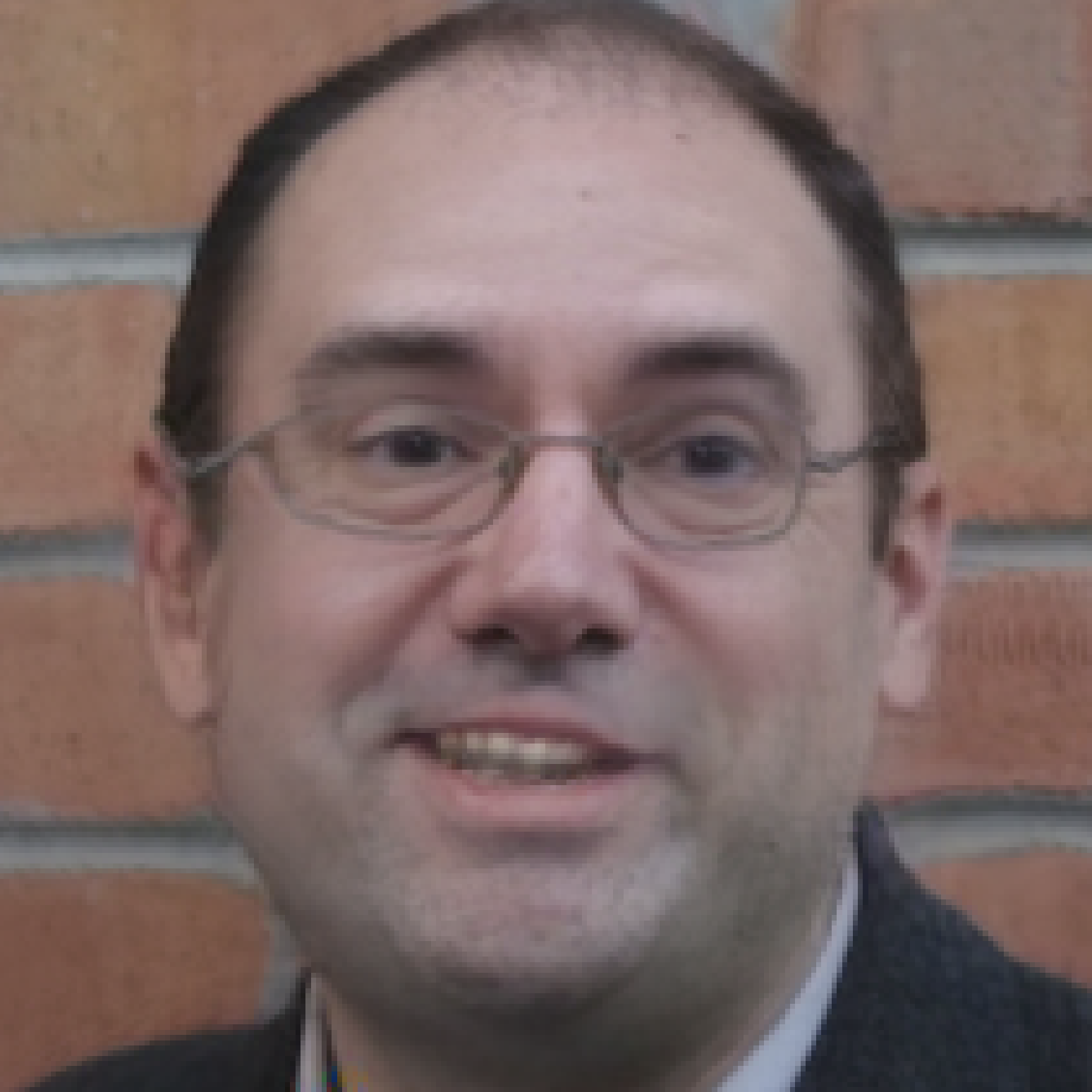} & 
        \hspace{\mrg}
        \includegraphics[width=\wid]{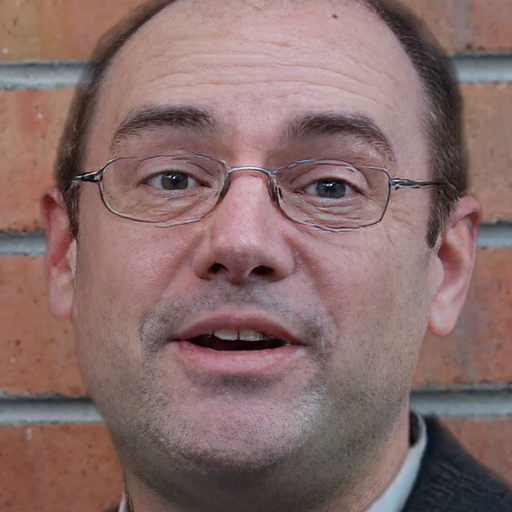} &
        \hspace{\mrg}
        \includegraphics[width=\wid]{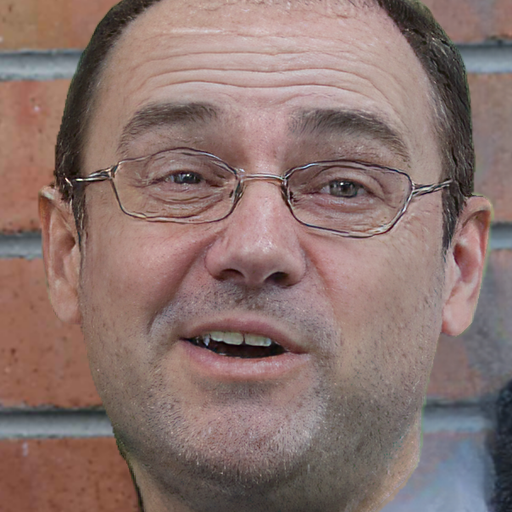} & 
        \\
        \includegraphics[width=\wid]{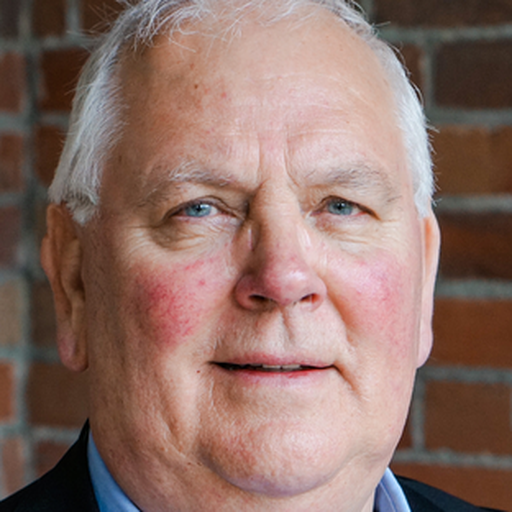} & 
        \hspace{\mrg}
        \includegraphics[width=\wid]{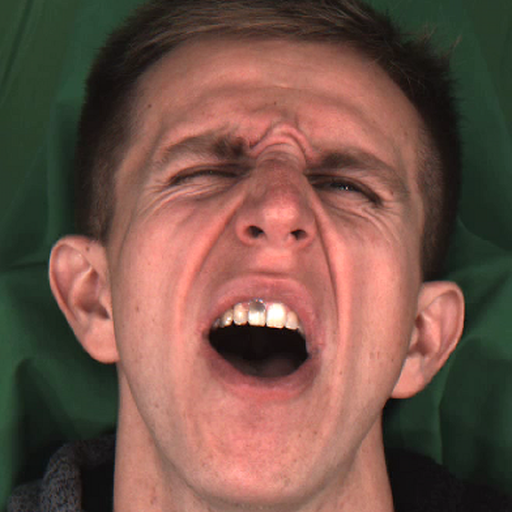} & 
        \hspace{\mrg}
        \includegraphics[width=\wid]{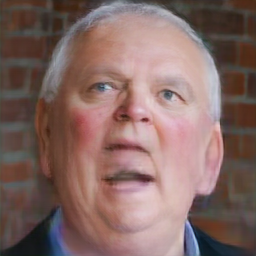} & 
        \hspace{\mrg}
        \includegraphics[width=\wid]{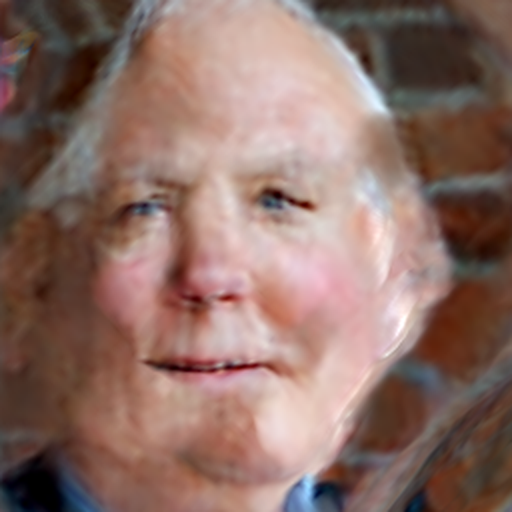} & 
        \hspace{\mrg}
        \includegraphics[width=\wid]{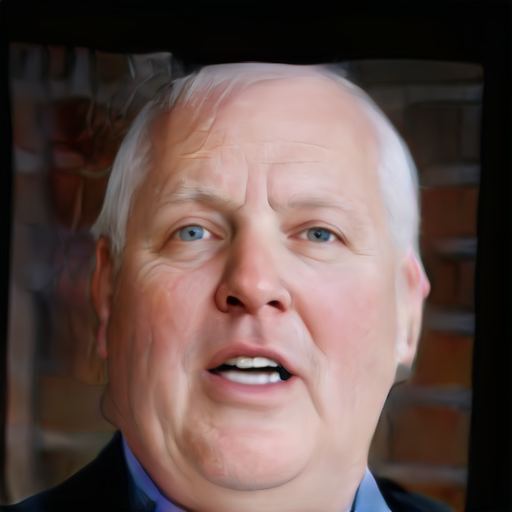} & 
        \hspace{\mrg}
        \includegraphics[width=\wid]{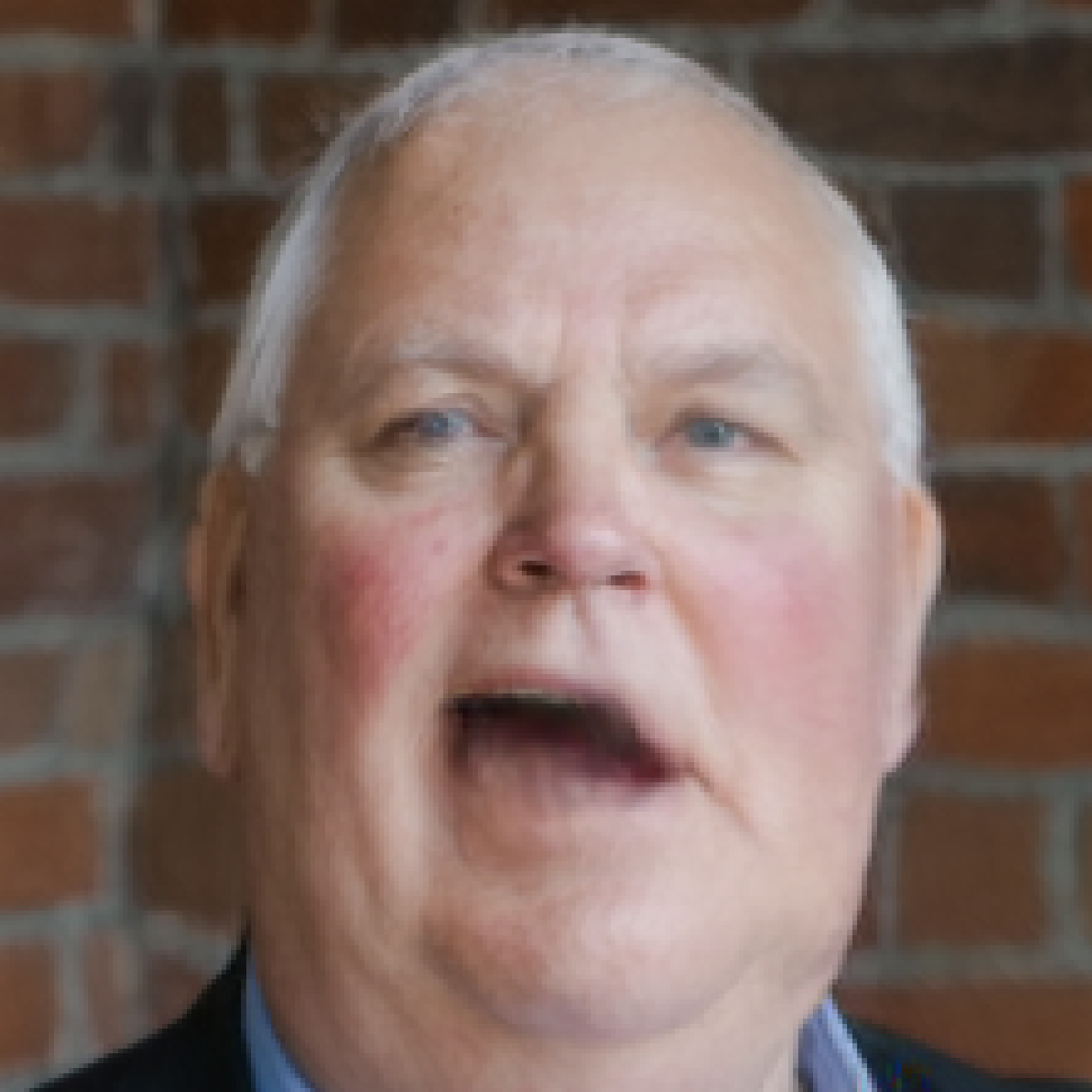} & 
        \hspace{\mrg}
        \includegraphics[width=\wid]{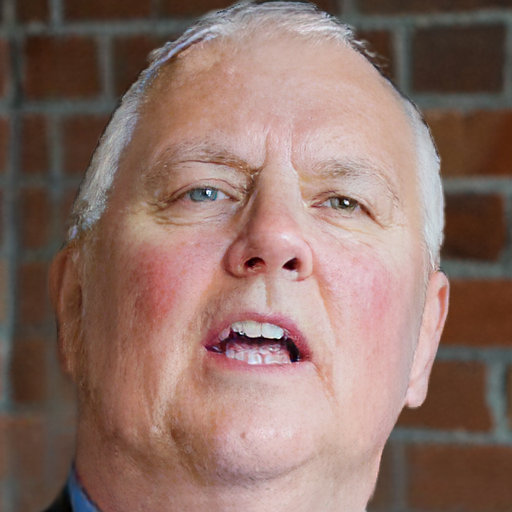} &
        \hspace{\mrg}
        \includegraphics[width=\wid]{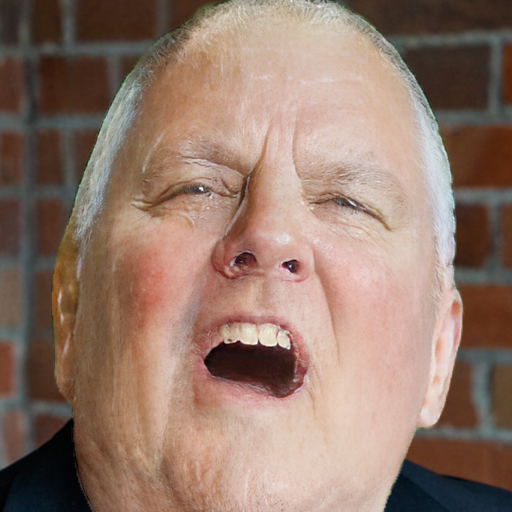} &  
        \\
        \includegraphics[width=\wid]{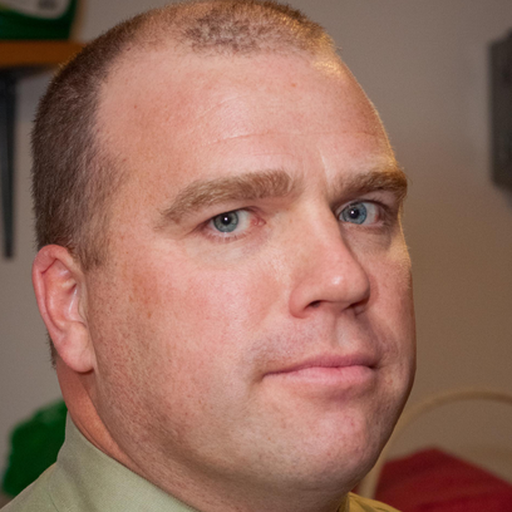} & 
        \hspace{\mrg}
        \includegraphics[width=\wid]{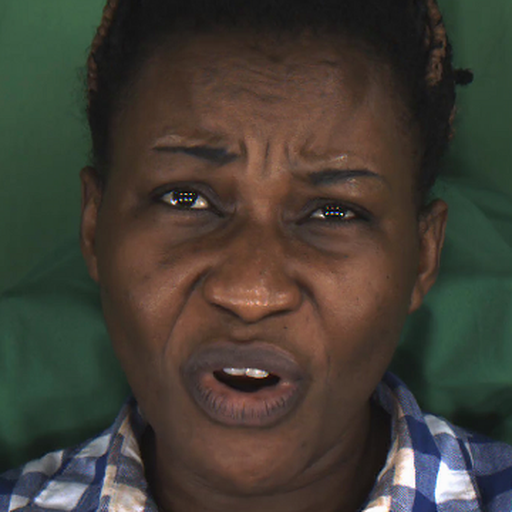} & 
        \hspace{\mrg}
        \includegraphics[width=\wid]{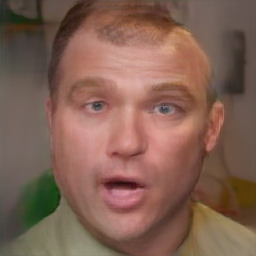} & 
        \hspace{\mrg}
        \includegraphics[width=\wid]{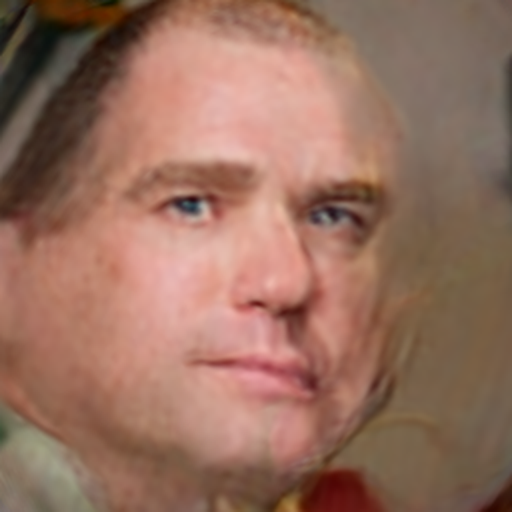} & 
        \hspace{\mrg}
        \includegraphics[width=\wid]{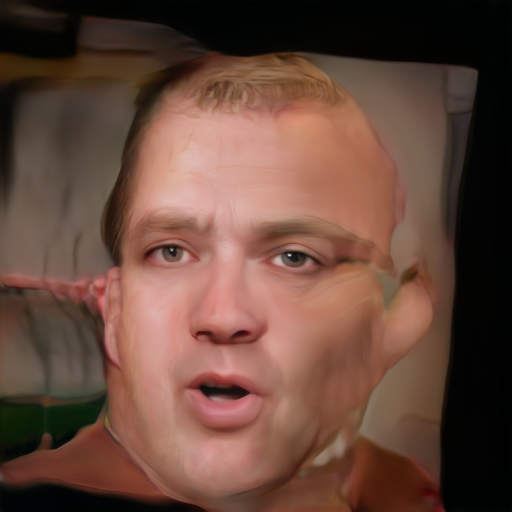} & 
        \hspace{\mrg}
        \includegraphics[width=\wid]{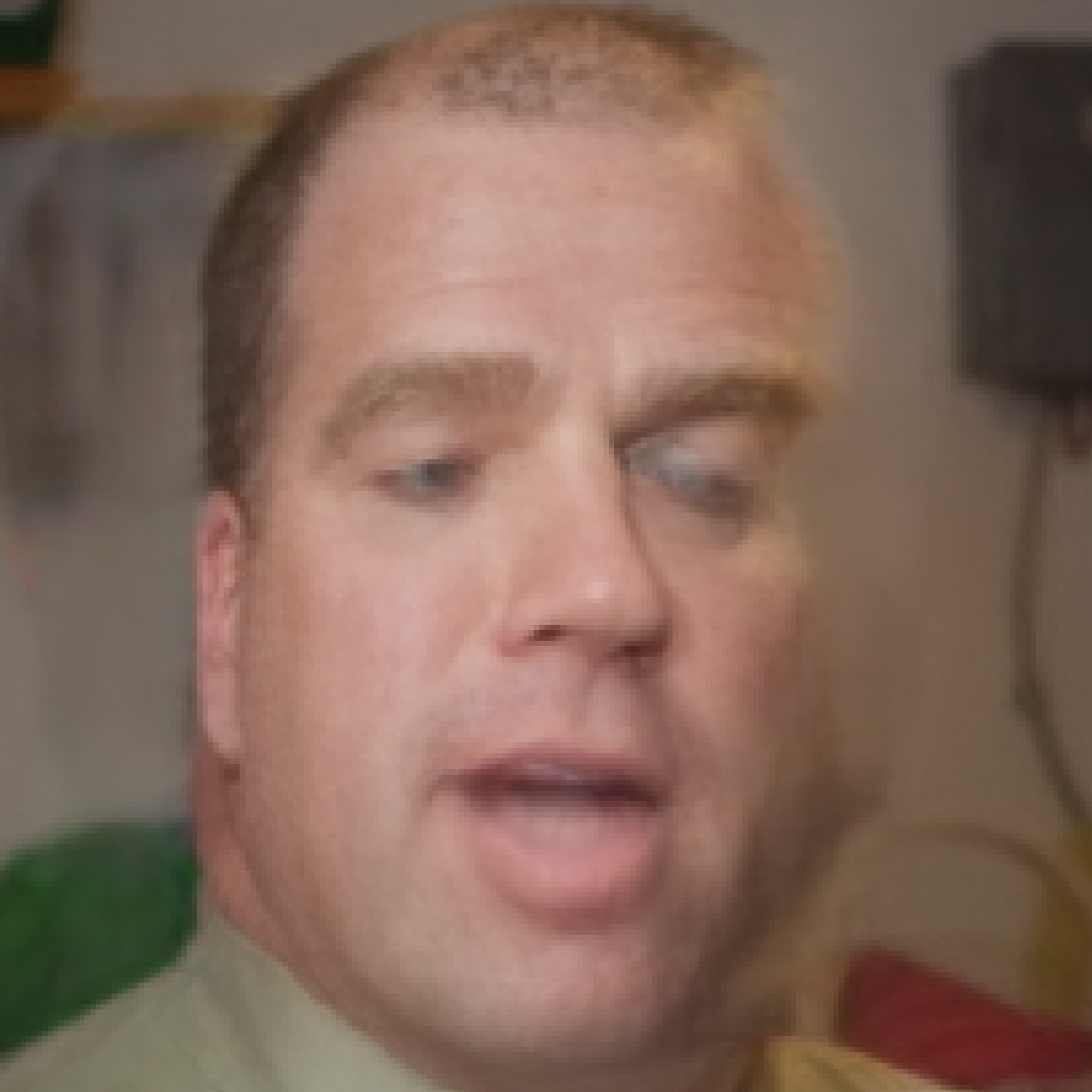} & 
        \hspace{\mrg}
        \includegraphics[width=\wid]{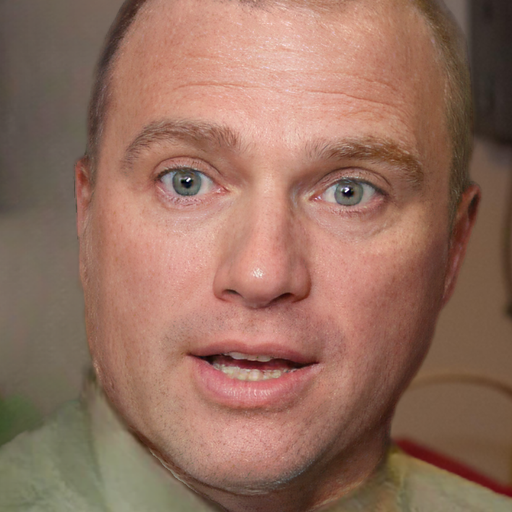} &
        \hspace{\mrg}
        \includegraphics[width=\wid]{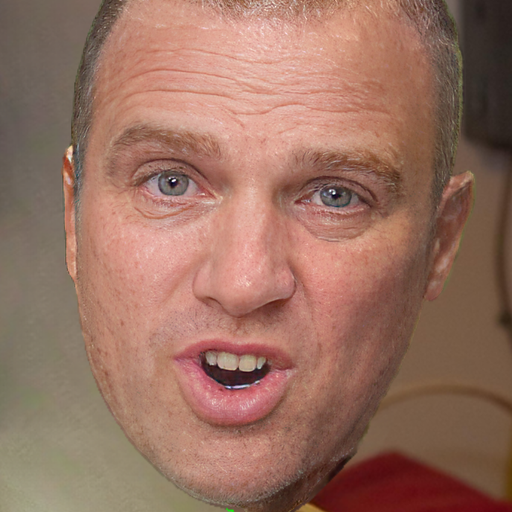} &  
        \\
        \includegraphics[width=\wid]{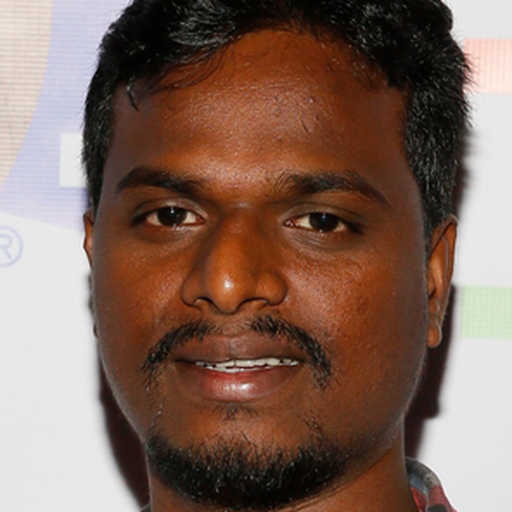} & 
        \hspace{\mrg}
        \includegraphics[width=\wid]{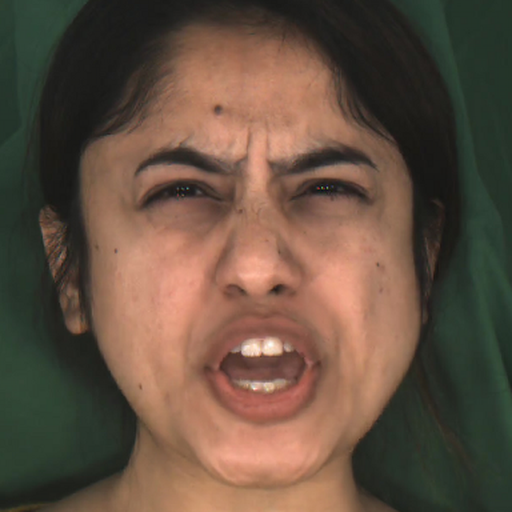} & 
        \hspace{\mrg}
        \includegraphics[width=\wid]{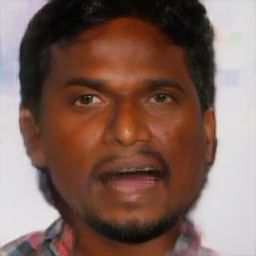} & 
        \hspace{\mrg}
        \includegraphics[width=\wid]{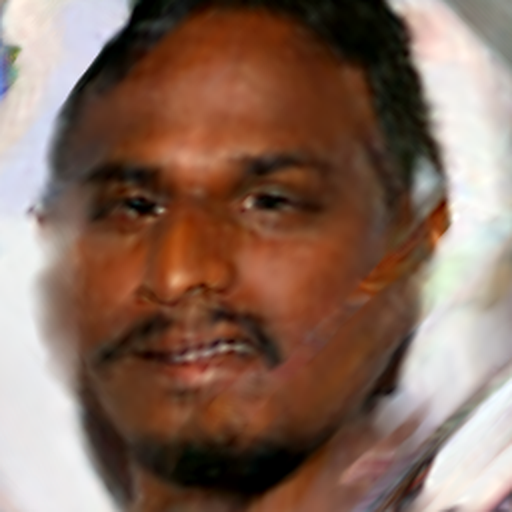} & 
        \hspace{\mrg}
        \includegraphics[width=\wid]{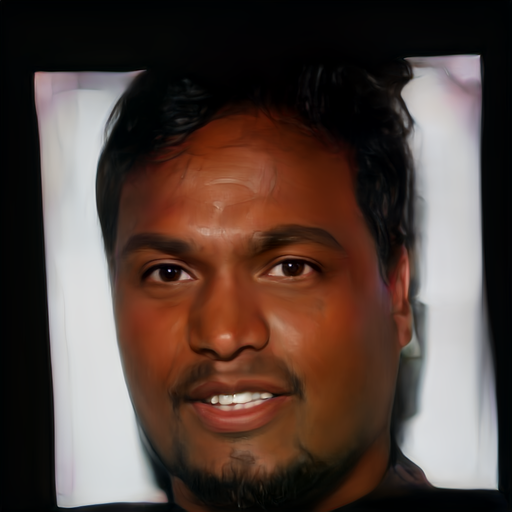} & 
        \hspace{\mrg}
        \includegraphics[width=\wid]{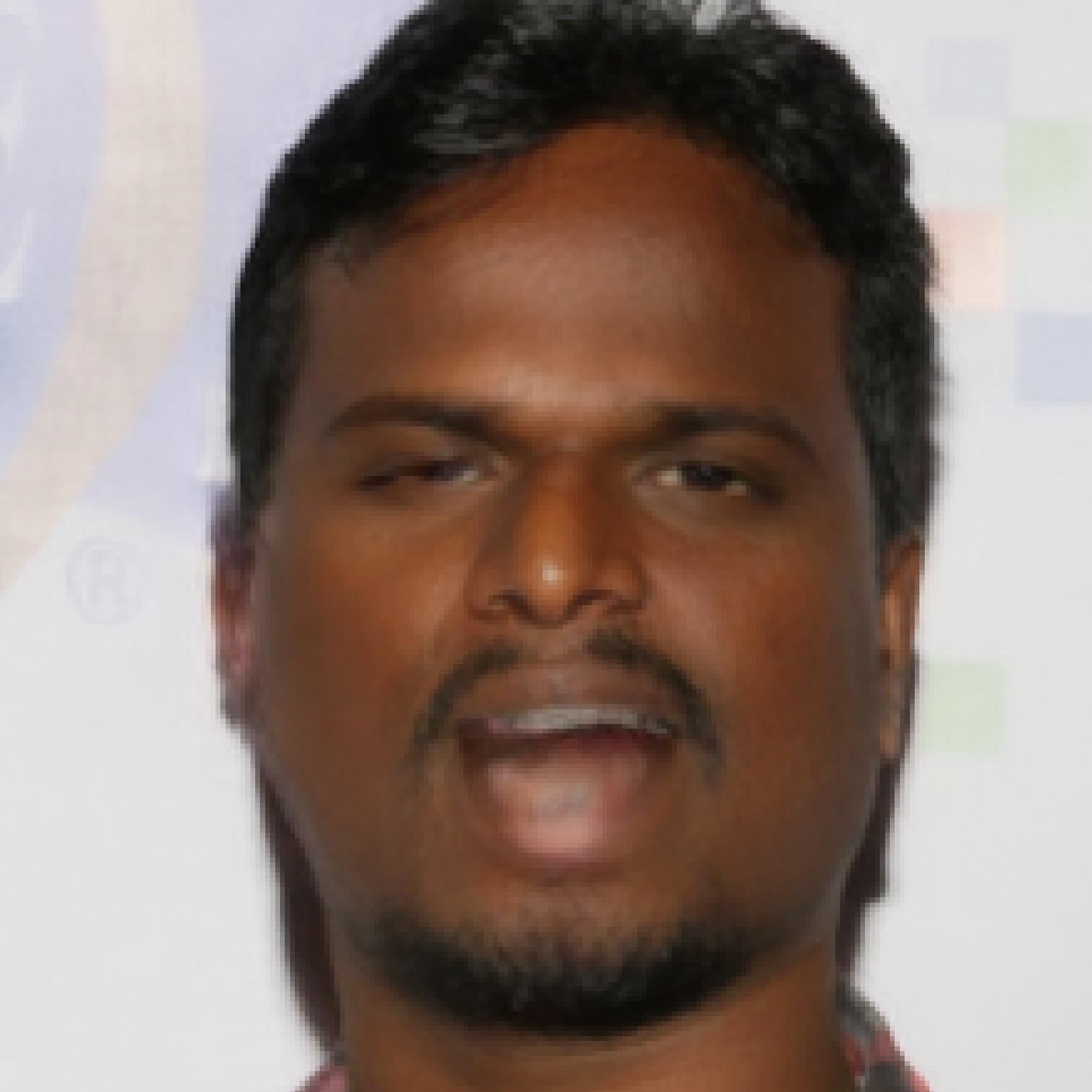} & 
        \hspace{\mrg}
        \includegraphics[width=\wid]{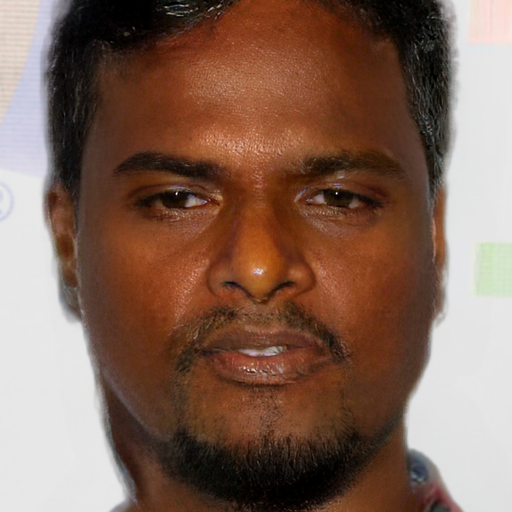} &
        \hspace{\mrg}
        \includegraphics[width=\wid]{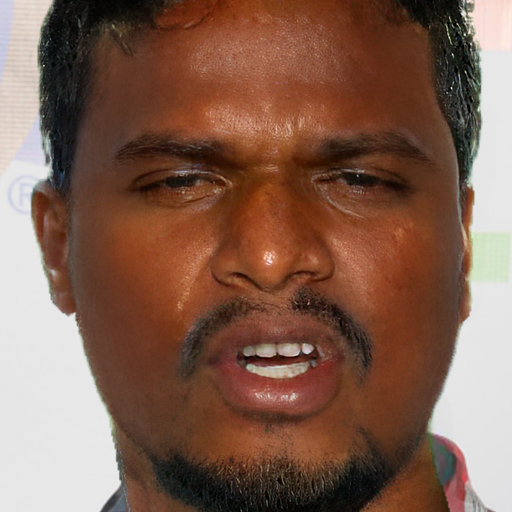} &  
        \\

        \textbf{Source} & 
        \hspace{\mrg} 
        \textbf{Driver} & 
        \hspace{\mrg} 
        \textbf{FOMM}~\cite{Siarohin2019FirstOM} & 
        \hspace{\mrg} 
        \textbf{UVA}~\cite{li2023generalizable} & 
        \hspace{\mrg} 
        \textbf{StyleHEAT}~\cite{yin2022styleheat} &
        \hspace{\mrg} 
        \textbf{MetaPortrait}~\cite{zhang2023metaportrait} &
        \hspace{\mrg} 
        \textbf{MegaPortraits}~\cite{drobyshev2023megaportraits} &
        \hspace{\mrg}
        \textbf{Ours}
    \end{tabular}
    }
    \vspace{-0.2cm}
    \caption{An additional qualitative comparison of head avatar systems in cross-reenactment scenario.}
    \label{fig:comparison_imgs_supp_2}
\end{figure*}

\begin{figure*}[]
    \centering    
    \setlength{\wid}{0.18\textwidth}
    \setlength{\mrg}{-0.3cm}
    \resizebox{\linewidth}{!}{
    \begin{tabular}{cccccccc}
        \includegraphics[width=\wid]{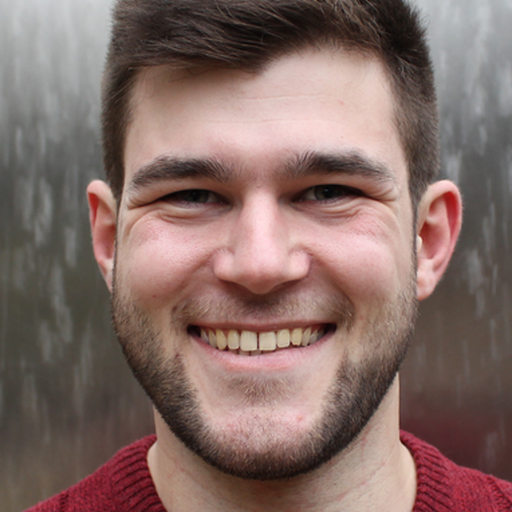} & 
        \hspace{\mrg}
        \includegraphics[width=\wid]{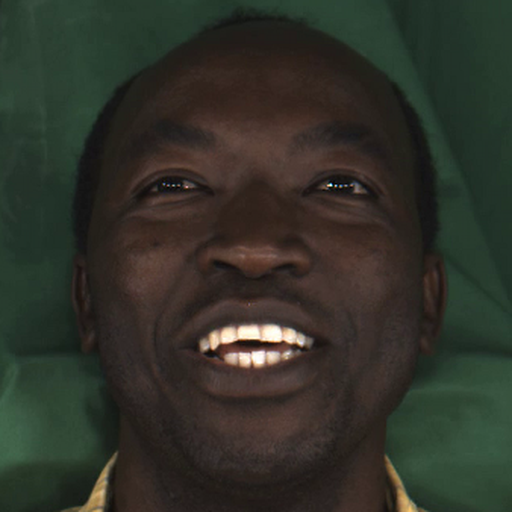} & 
        \hspace{\mrg}
        \includegraphics[width=\wid]{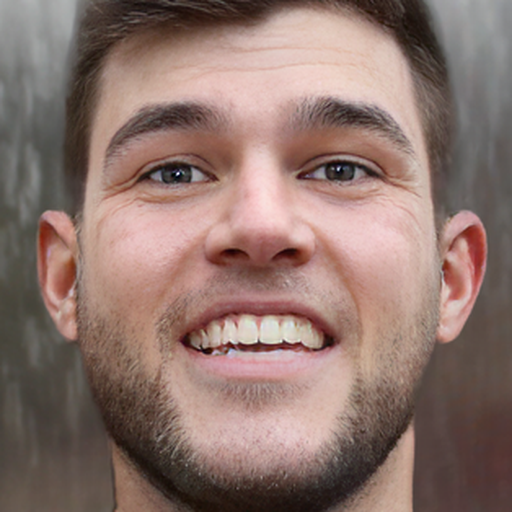} & 
        \hspace{\mrg}
        \includegraphics[width=\wid]{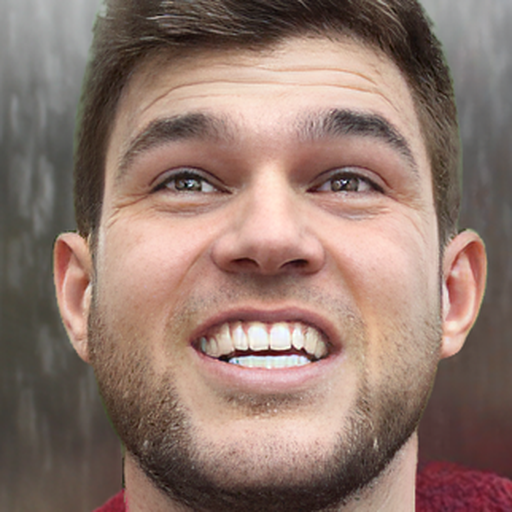} & 
        \hspace{\mrg}
        \includegraphics[width=\wid]{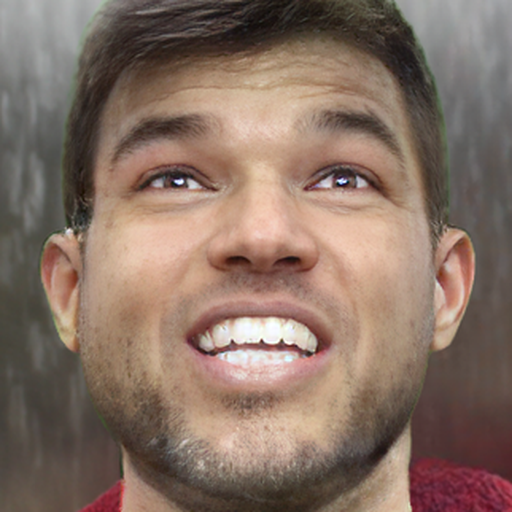} & 
        \hspace{\mrg}
        \includegraphics[width=\wid]{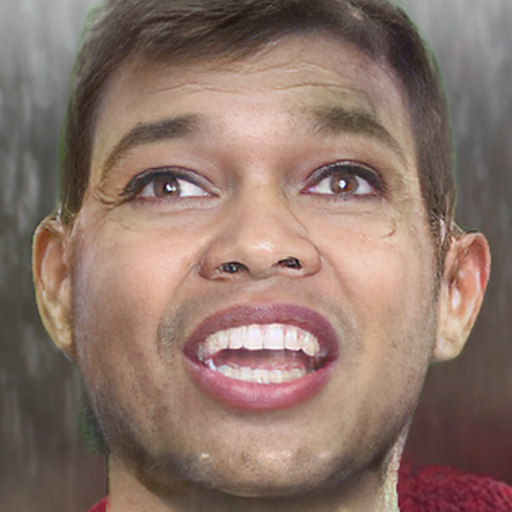} & 
        \hspace{\mrg}
        \includegraphics[width=\wid]{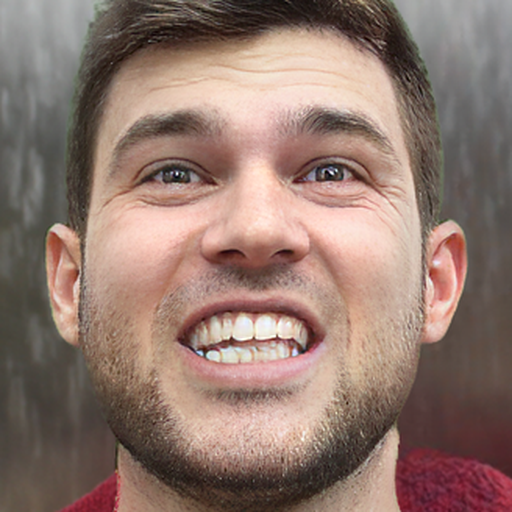} & 
        \\ %
        \includegraphics[width=\wid]{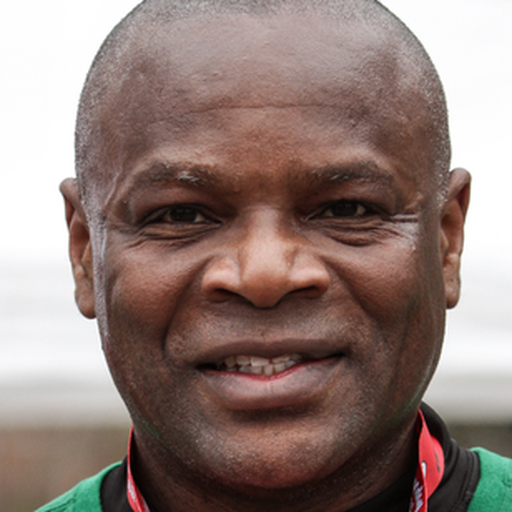} & 
        \hspace{\mrg}
        \includegraphics[width=\wid]{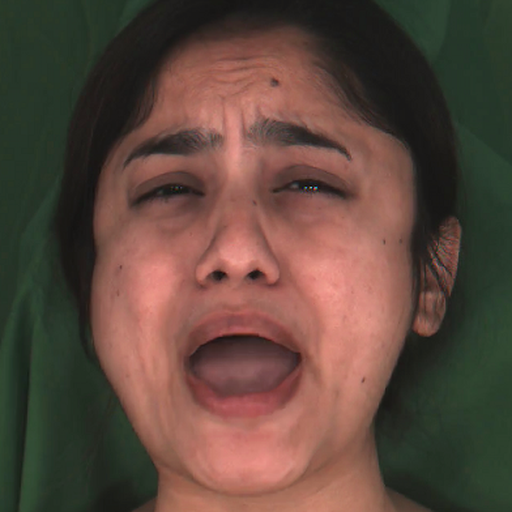} & 
        \hspace{\mrg}
        \includegraphics[width=\wid]{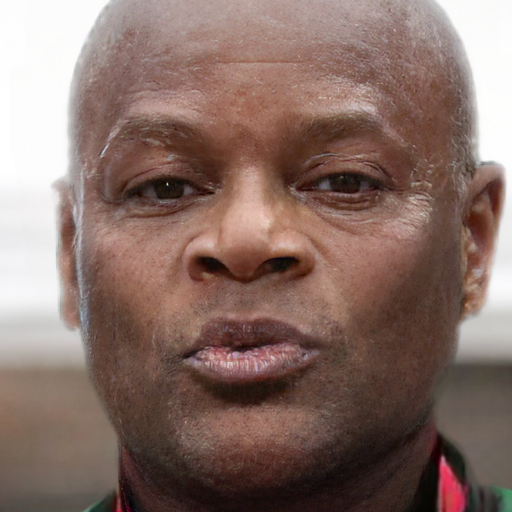} & 
        \hspace{\mrg}
        \includegraphics[width=\wid]{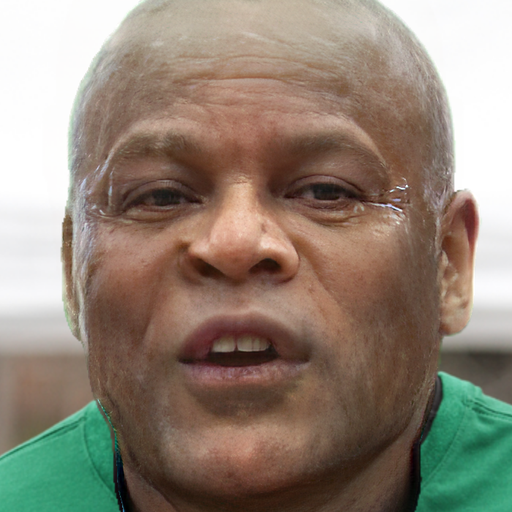} & 
        \hspace{\mrg}
        \includegraphics[width=\wid]{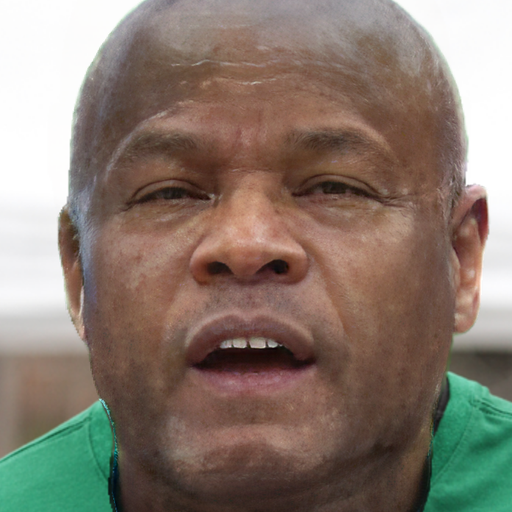} & 
        \hspace{\mrg}
        \includegraphics[width=\wid]{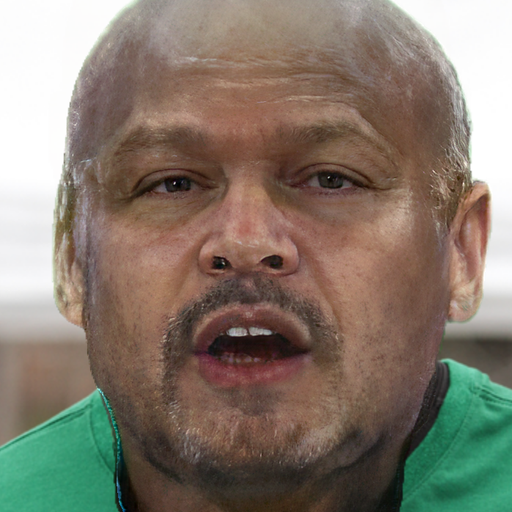} & 
        \hspace{\mrg}
        \includegraphics[width=\wid]{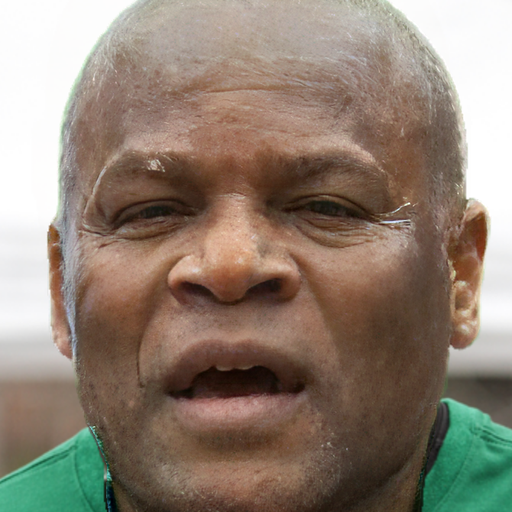} & 
        \\
        \includegraphics[width=\wid]{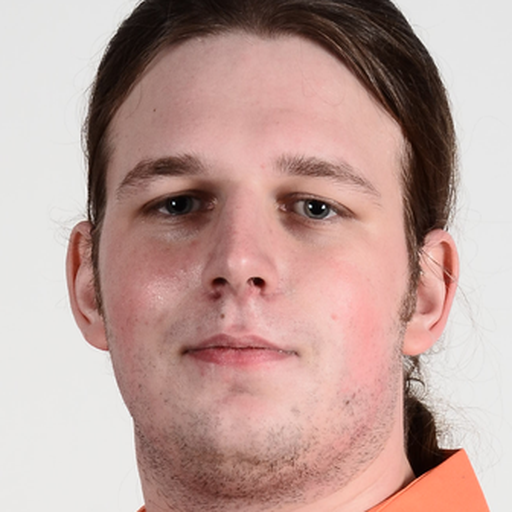} & 
        \hspace{\mrg}
        \includegraphics[width=\wid]{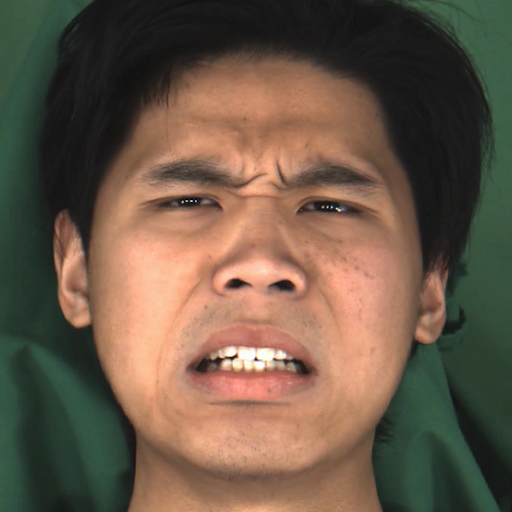} & 
        \hspace{\mrg}
        \includegraphics[width=\wid]{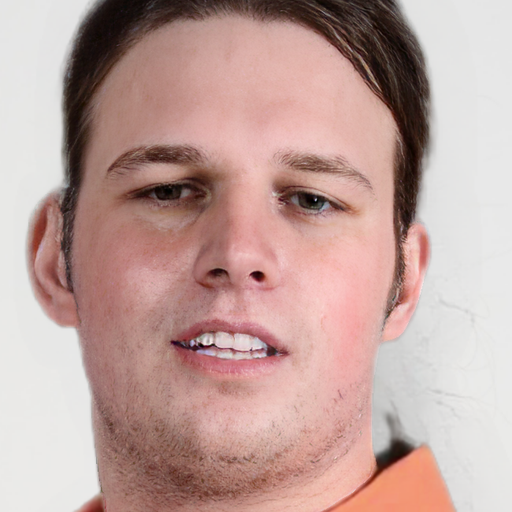} & 
        \hspace{\mrg}
        \includegraphics[width=\wid]{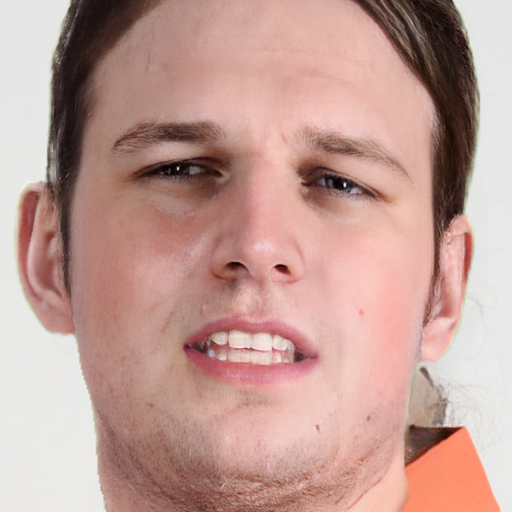} & 
        \hspace{\mrg}
        \includegraphics[width=\wid]{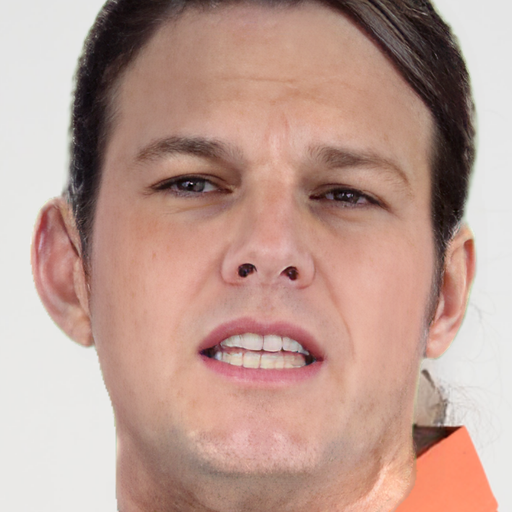} & 
        \hspace{\mrg}
        \includegraphics[width=\wid]{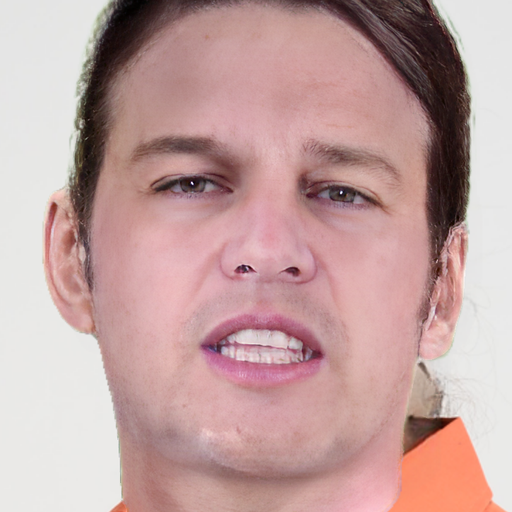} & 
        \hspace{\mrg}
        \includegraphics[width=\wid]{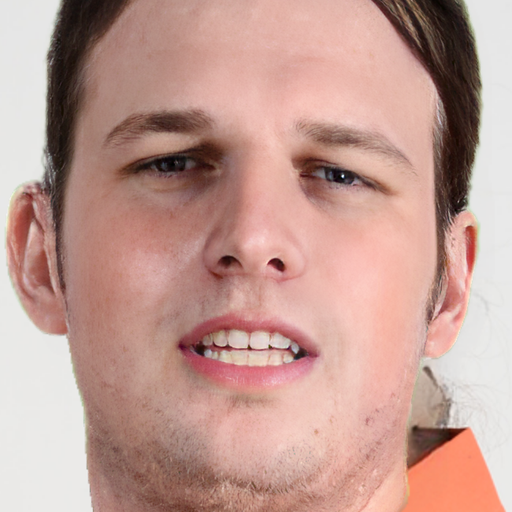} & 
        \\

        \textbf{Source} & 
        \hspace{\mrg} 
        \textbf{Driver} & 
        \hspace{\mrg} 
        MegaPortraits fine-tuned & 
        \hspace{\mrg} 
        w/o $\mathcal{L}_{CV}^n$ & 
        \hspace{\mrg} 
        w/o $\textit{dim}(\z) = 512$ &
        \hspace{\mrg} 
        $\mathcal{L}_{sdm}$ &
        \hspace{\mrg}
        \textbf{Ours}
    \end{tabular}
    }
    \vspace{-0.2cm}
    \caption{Visual comparison for our ablation study \cref{fig:ablation_img}.}
    \label{fig:ablation_img}
\end{figure*}

To supplement the comparisons described in the main paper, we provide additional examples for each method (refer to Figure~\ref{fig:comparison_imgs_supp_1}). As, for NOFA~\cite{yu2023nofa}, the range of examples is limited due to the restricted number of inferred identities provided by the authors, we provide a second set of additional examples excluding NOFA~\cite{yu2023nofa} (see Figure~\ref{fig:comparison_imgs_supp_2}). Furthermore, we include a visual comparison Figure~\ref{fig:ablation_img} for our ablation study (see Table 4 in the main text).

\section{Speech-driven mode}

\subsection{Implementation details}

\begin{figure}[t]
    \centering
    \includegraphics[width=\linewidth]{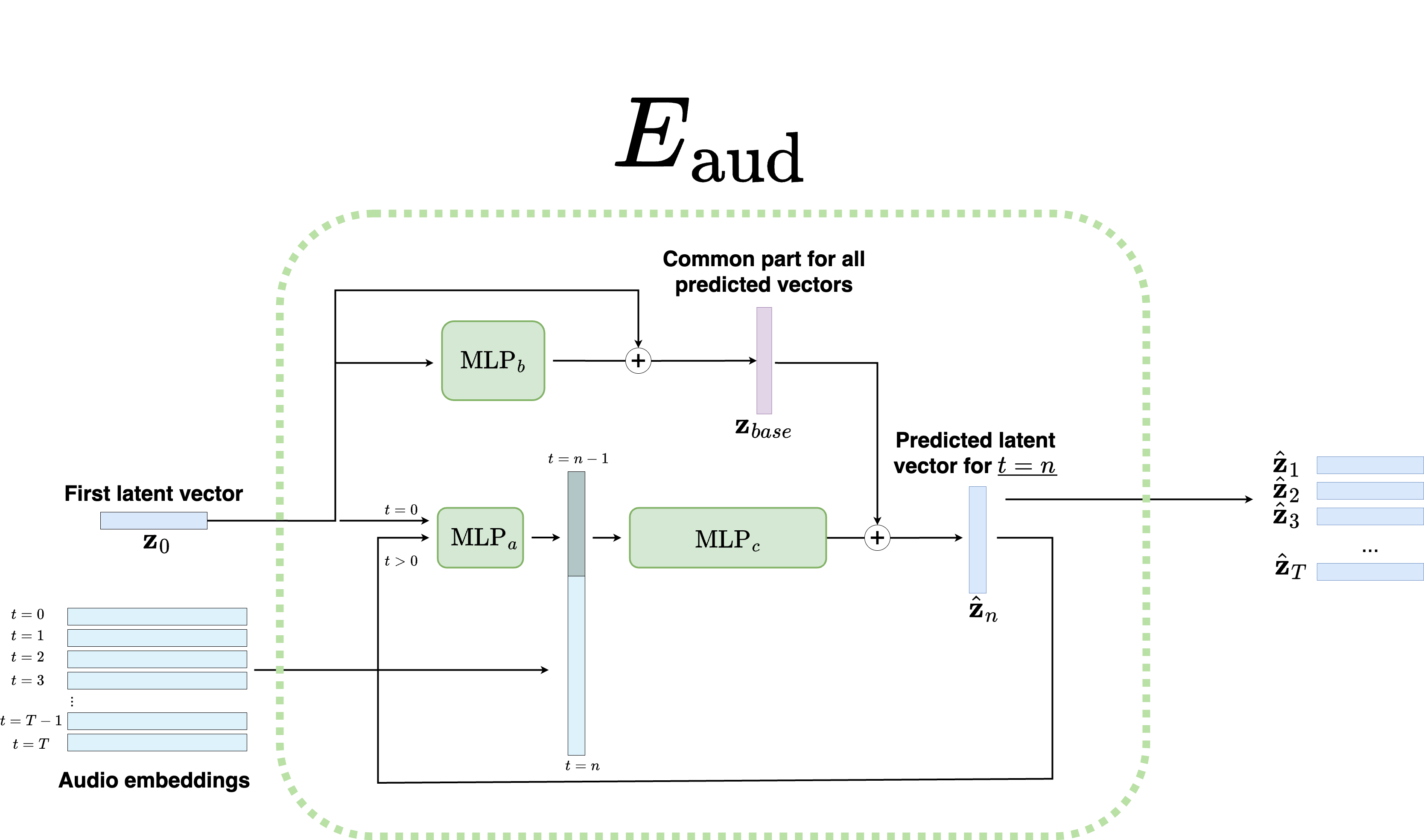}
    \caption{Comparison of our audio encoder used to predict latent expression vectors during speech-driven mode.}
    \label{fig:e_aud}
\end{figure}

In this section, we detail the workings of our audio encoder, $\E_{aud}$, as depicted in \cref{fig:main_scheme}, for its application in speech-driven scenarios. The encoder, $\E_{aud}$, is designed for generating latent expression vectors, denoted as $\z$, from speech inputs. The scheme of the encoder presented on \cref{fig:e_aud}.

First, we use Whisper \cite{radford2022robust} model to retrieve audio embeddings from a raw audio clip containing speech. This step yields a series of $T$ audio embedding vectors, where each vector is linked to a specific frame in the video clip. Next, we employ a multilayer perceptron (MLP), designated as $\text{MLP}_b$, to compute the base component of our latent expression vectors, $\z_{\text{base}}$. This base component encapsulates common facial features such as initial gaze direction and the pose of the upper facial region, as observed in the first frame of the input. Then, another MLP, $\text{MLP}_a$, is employed. It uses the previously calculated latent vector $\z$—initially which is $\z_0$ for $t=0$ and  $\hat\z_n$ for $t>0$—to align the latent features of $\z$ with those of the audio embeddings and derive features that useful for final prediction. In the final step, after merging the relevant audio embedding with the output from $\text{MLP}_a$, the next network, $\text{MLP}_c$ is utilized and responsible for computing a part of the vector which, when added to $\z_{\text{base}}$, forms the final latent expression vector $\hat\z_n$ for each frame.

\subsection{Data Preparation}
Similar to our primary model, we employ the VoxCeleb2 dataset \cite{Chung2018VoxCeleb2DS} for training our audio encoder, $\E_{aud}$. During each training iteration, we randomly select an audio clip, varying in length from 50 to 200 frames. The corresponding audio segment and the initial frame of this clip are fed into $\E_{aud}$ as inputs. All subsequent frames from the clip are utilized as reference frames (ground truth) for training purposes. Both the training and evaluation processes are conducted using a resolution of $512 \times 512$.

\subsection{Utilized Loss Functions}
\label{ssec:losses_aud}

Training of $\E_{aud}$ incorporates three distinct types of loss functions:

\textbf{Photometrical Identity Preservation Losses}: To ensure the identity of the individual in the video is preserved, we apply an L1 loss between the predicted image (using output expression vectors from $\E_{aud}$, denoted as $\hat{\x}_n^\text{aud}$) and the ground truth image $\x_n$. Additionally, we implement a perceptual loss comparing facial features in these images using the \textit{VGGFace} model \cite{Parkhi2015DeepFR}. The identity preservation loss is represented as:
    \begin{equation}
    \mathcal{L}_\text{idt} = w_\text{L1} \mathcal{L}_\text{L1} + w_\text{face} \mathcal{L}_\text{face}
    \end{equation}

\textbf{Latent Mouth Movement Losses}: For accurately translating mouth movements, we use an L1 loss focusing on the principal components related to mouth movements in $\hat{\z}_n$ and $\z_n$. This loss, detailed in \cref{ssec:mouth_move}, is denoted as $\mathcal{L}_\text{PCA}(\z_i, \z_j, n)$, with $n=8$ in our experiments. A secondary L1 loss, $\mathcal{L}_\text{vtr}$, with a reduced weight, ensures a closer match between $\hat{\z}_n$ and $\z_n$ vectors. 
    \begin{equation}
    \mathcal{L}_\text{latent} = w_\text{PCA} \mathcal{L}_\text{PCA} + w_\text{vtr} \mathcal{L}_\text{vtr}
    \end{equation}

\textbf{Photometrical Lip Movement Losses}: For enhanced translation of mouth movements, we employ the FaceParsing network \cite{yu2018bisenet} to generate masks for the upper lip, lower lip, and inner mouth regions in both the predicted ($\hat{\x}_n^\text{aud}$) and ground truth ($\x_n$) images. These masks are compared using the Binary Cross Entropy loss, $\mathcal{L}_\text{BCE}$. Additionally, we extract pixels corresponding to the mouth in both predicted and actual images, comparing them using an L1 loss, $\mathcal{L}_\text{lips}$:

\begin{equation}
    \mathcal{L}_\text{mouth} = w_\text{BCE} \mathcal{L}_\text{BCE} + w_\text{lips} \mathcal{L}_\text{lips}
\end{equation}

The overall loss function used to train our model is the cumulative sum of these individual losses:

\begin{equation}
    \mathcal{L}_\text{speech} = \mathcal{L}_\text{idt} + \mathcal{L}_\text{latent}  + \mathcal{L}_\text{mouth}
\end{equation}

We utilized the AdamW optimizer~\cite{Loshchilov2019DecoupledWD} with a cosine learning rate schedule. The initial learning rate was gradually reduced from \(1 \times 10^{-4}\) to zero over the training iterations. The hyperparameters for the losses were set as follows: \(w_\text{L1} = 10\), \(w_\text{face} = 100\), \(w_\text{PCA} = 200\), \(w_\text{vtr} = 5\), \(w_\text{BCE} = 5 * 10^3\), \(w_\text{lips} = 5*10^5\).

\section{Generating head rotations}

\begin{figure}[t]
    \centering
    \includegraphics[width=\linewidth]{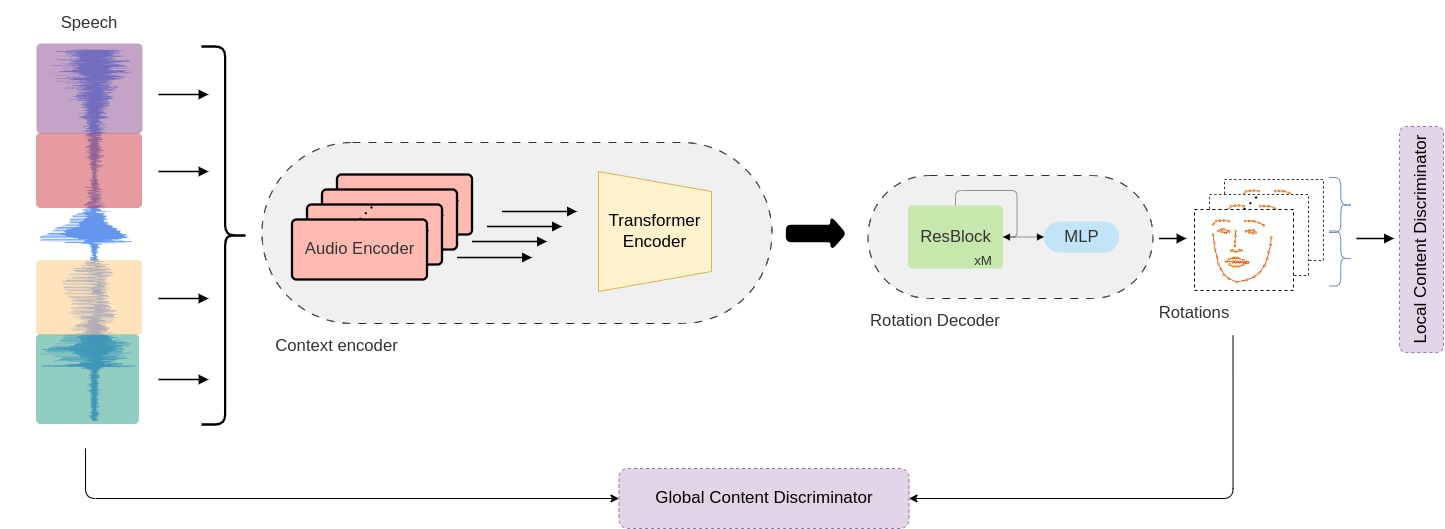}
  \caption{\textbf{Audio-to-rotations model}. The input signal is split into overlapping segments for the audio encoder, processed through a Transformer and a ResNet decoder. The resulting rotation sequence and audio input are then given to the Global Content Discriminator, while the Local Temporal Discriminator receives smaller segments of these rotations.}
    \label{fig:aud2rot}
\end{figure}

To control the main model using head rotations generated from speech, we have developed a generative adversarial model. This model takes a speech recording as input and outputs a series of rotations, as depicted in Fig~\ref{fig:aud2rot}.

\subsection{Rotation Representation}

We represent the 3D rotations with six dimensions as described in \cite{zhou2019rot}. This ensures the continuity of the representation, which is more suitable for learning. We also add an extra three parameters to predict the translation. Therefore, we can formulate a head transformation sequence $X$ as a sequence of rotations and translation across $T$ consecutive frames, $X \in R^{T \times 9}$ where each $X_t \in R^{9}$ is a vector representing the transformation from the reference frame.
To map the 6D representation again to the 3D rotation group, we can use the following formula:
\begin{equation}
    f_{GS}\left( \begin{bmatrix} 
            \mid & \mid \\
            a_{1} & a_{2} \\
            \mid & \mid
            \end{bmatrix} \right) = \begin{bmatrix} 
            \mid & \mid & \mid \\
            b_{1} & b_{2} & b_{3} \\
            \mid & \mid & \mid
            \end{bmatrix}
\label{eq:gm1}
\end{equation}
\begin{equation}
    b_{i}\left[ \left\{
                \begin{array}{ll}
                  N(a_{1})                         & \text{if} \: i = 1\\
                  N(a_{2}-(b_{1}\cdot a_{2})b_{1}) & \text{if} \: i = 2\\
                  b_{1} \times b_{2} & \text{if} \: i = 3
                \end{array}
              \right. \right]^{T}
\label{eq:gm2}
\end{equation}

Here $N(\cdot)$ denotes a normalization function and $f_{GS}$ a Gram-Schmidt process. The model produces a 3x2 matrix with $a_1$, $a_2$ being its columns. The Gram-Schmidt process in Equation (\ref{eq:gm2}) produces the third column $b_3$ by taking the cross product of the two first columns $b_1$ and $b_2$, making it normal to the plane containing them. This process ensures that the resulting 3x3 matrix is orthogonal. The remaining three dimensions map directly to the translations by construction.

\subsection{Generator}

Our generator is split into two components: a context encoder and a head pose decoder. The context encoder merges an audio encoder, Whisper, with a transformer encoder for temporal analysis. This setup efficiently leverages existing audio embeddings from other parts of our system. The head pose is decoded using the encoder's hidden states, processed through ResNet layers and an MLP. This converts $R^{T \times H}$ to $R^{T \times 9}$, where $H$ is the Transformer's hidden size.

\subsection{Discriminators}

To evaluate our generated rotations, we use two discriminators assessing local and global coherence.

\textbf{Local Content Discriminator.} Inspired by Isola et al. \cite{isola2018imagetoimage}, we use a 1D temporal PatchGAN variant. This discriminator targets patch-level structures, classifying patches of N frames as real or fake. With the discriminator convolution spanning the entire sequence, averaging all responses yields the final judgment. We found N=8 optimizes frame-to-frame coherence.

\textbf{Global Content Discriminator.} This discriminator assesses the full sequence's coherence with the audio input. We encode the rotation sequence using 1D ResNet blocks and a global pooling layer, then concatenate it with the audio embeddings from Whisper. A final MLP layer determines if the sequence is authentic or generated.

\subsection{Losses}

To train the model, we use a weighted combination of different losses. The resulting loss is described as follows:
\begin{equation}
\mathcal{L}_{tot} = \lambda_{recons}\mathcal{L}_{recons} +  \lambda_{adv}\mathcal{L}_{adv}  + \lambda_{smooth}\mathcal{L}_{smooth} 
\label{eq:totloss}
\end{equation}

\textbf{Reconstruction loss} $\mathcal{L}_{recons}$ \textbf{.} Is a $L_{1}$ loss between predicted $X$ and ground-truth $Y$ head poses. The head poses can be separated into rotation $r$ and translation $t$.
\begin{equation}
    \mathcal{L}_{recons} = \sum_{i=1}^{T} \: \abs*{r_{i} - \hat{r}_{i}} + \sum_{i=1}^{T} \: \abs*{t_{i} - \hat{t}_{i}}
\label{eq:transloss}
\end{equation}

We then have two different coefficient factors $\lambda_{rot}$ and $\lambda_{trans}$ for the reconstruction of rotations and translations.

\textbf{Smoothing loss} $\mathcal{L}_{smooth}$ \textbf{.} Acts as a regularization loss and ensures smoothness over consecutive frames.
\begin{equation}
    \mathcal{L}_{smooth} = \sum_{i=2}^{T} \: \abs*{X_{i} - X_{i-1}}
    \label{eq:smoothloss}
\end{equation}

\textbf{Adversarial loss} $\mathcal{L}_{adv}$ \textbf{.} We adopt WGAN-GP \cite{gulrajani2017improved} for improved stability of the training and for avoiding mode collapse. The discriminator and generator losses are as follows:
\begin{equation}
\begin{split}
    \mathcal{L}_{D} & = \mathbb{E}_{\Tilde{x} \sim \mathbb{P}_{g}}[D(\Tilde{x})] - 	\mathbb{E}_{x \sim \mathbb{P}_{r}}[D(x)] \\
    & + \lambda \mathbb{E}_{\hat{x} \sim \mathbb{P}_{\hat{x}}}[(\norm{\nabla_{\hat{x}} D(\hat{x})}_2 - 1 )^2]
\end{split}
    \label{eq:advlossdis}
\end{equation}

\begin{equation}
    \mathcal{L}_{G} = - \mathbb{E}_{\Tilde{x} \sim \mathbb{P}_{g}}[D(\Tilde{x})]
    \label{eq:advlossgen}
\end{equation}

$\mathbb{P}_{\hat{x}}$ is defined as a uniform sampling along straight lines between pairs of points sampled from the data distribution $\mathbb{P}_{r}$ and the generator distribution $\mathbb{P}_{g}$. We set $\lambda$ to 10 and update five times the discriminator for every single update of the generator as suggested in the original paper. 



\begin{figure*}[]
    \centering
    \includegraphics[width=\linewidth]{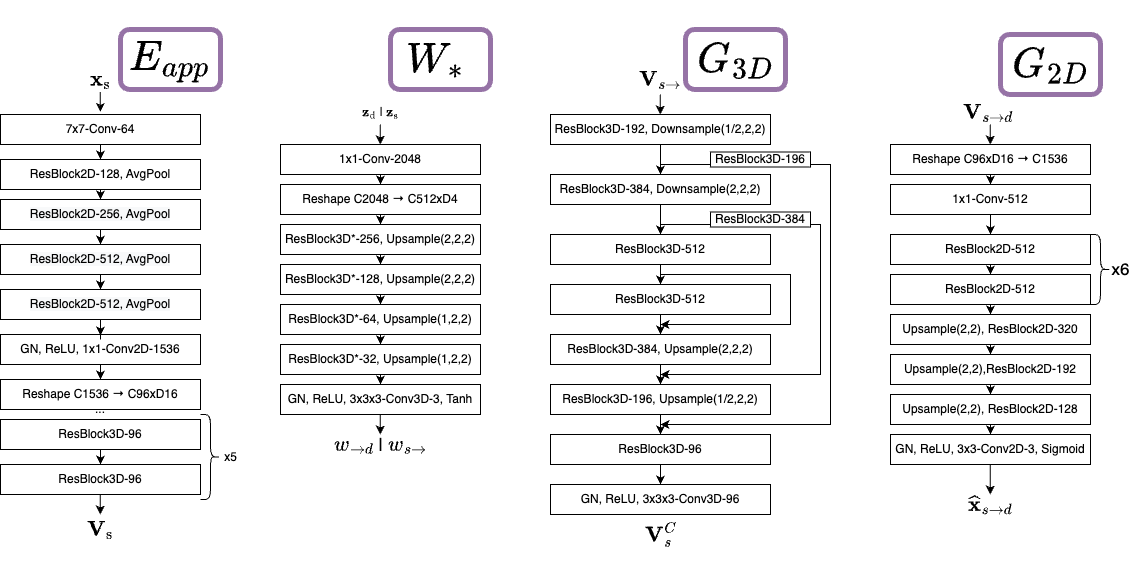}
    \caption{Architectures of main components of our main model}
    \label{fig:networks_arch}
\end{figure*}



\end{document}